\documentclass[11pt, a4paper]{thuc3i}
\usepackage[sort&compress]{natbib}
\setheadertext{A Survey of Reinforcement Learning for Large Reasoning Models}

\usepackage[utf8]{inputenc} %
\usepackage[T1]{fontenc}    %
\usepackage{hyperref}       %
\usepackage{url}            %
\usepackage{tabularx}
\usepackage{makecell}
\usepackage{array,ragged2e,booktabs}
\newcolumntype{P}[1]{>{\RaggedRight\arraybackslash}p{#1}}
\usepackage{longtable}
\usepackage{amsfonts}       %
\usepackage{amsthm}         %

\usepackage{nicefrac}       %
\usepackage{microtype}      %
\usepackage{xcolor}         %

\usepackage{algorithm}
\usepackage{algorithmicx}
\usepackage{algpseudocode}

\definecolor{darkblue}{rgb}{0, 0, 0.5}
\hypersetup{colorlinks=true, citecolor=darkblue, linkcolor=darkblue, urlcolor=darkblue}

\usepackage{xurl}

\usepackage{soul}
\usepackage{amsmath}
\usepackage{subcaption}
\usepackage{graphicx}
\usepackage{changes}
\usepackage{amsmath,mathtools}
\usepackage{changes}
\usepackage{pifont}
\usepackage{tocloft}

\usepackage{lipsum}
\usepackage{colortbl}
\usepackage{wrapfig}
\usepackage{xspace}
\usepackage{multirow}
\usepackage{enumitem}

\usepackage{pifont}
\usepackage{listings}
\usepackage{fontawesome5} 

\usepackage{wrapfig}
\usepackage{caption}

\usepackage{makecell}

\usepackage{tabularx}
\usepackage{afterpage}

\usepackage[edges]{forest}
\usepackage[normalem]{ulem}
\usepackage{caption}
\usepackage{CJKutf8}
\usepackage{bbding}
\usepackage[most,skins,theorems]{tcolorbox}
\usepackage{longtable}
\usepackage{booktabs}
\usepackage{multirow}
\usepackage{hyperref}
\usepackage{array}

\usepackage[tikz]{bclogo}
\usepackage[framemethod=tikz]{mdframed}
\definecolor{bgblue}{RGB}{245,243,253}
\definecolor{ttblue}{RGB}{91,194,224}

\mdfdefinestyle{mystyle}{%
  rightline=true,
  innerleftmargin=10,
  innerrightmargin=10,
  outerlinewidth=3pt,
  topline=false,
  rightline=true,
  bottomline=false,
  skipabove=\topsep,
  skipbelow=\topsep
}

\newtcolorbox{myboxi}[1][]{
  breakable,
  title=#1,
  colback=red!5,
  colbacktitle=red!5,
  coltitle=black,
  fonttitle=\bfseries,
  bottomrule=0pt,
  toprule=0pt,
  leftrule=2pt,
  rightrule=2pt,
  titlerule=0pt,
  arc=0pt,
  outer arc=0pt,
  colframe=red,
}

\newtcolorbox{myboxnote}[1][]{
  breakable,
  title=#1,
  colback=orange!0,
  colbacktitle=orange!0,
  coltitle=black,
  fonttitle=\bfseries,
  bottomrule=0pt,
  toprule=0pt,
  leftrule=2pt,
  rightrule=2pt,
  titlerule=0pt,
  arc=0pt,
  outer arc=0pt,
  colframe=orange,
}

\usepackage{epigraph}

\newtcolorbox{myboxii}[1][]{
  breakable,
  freelance,
  title=#1,
  colback=white,
  colbacktitle=white,
  coltitle=black,
  fonttitle=\bfseries,
  bottomrule=0pt,
  boxrule=0pt,
  colframe=white,
  overlay unbroken and first={
  \draw[red!75!black,line width=3pt]
    ([xshift=5pt]frame.north west) -- 
    (frame.north west) -- 
    (frame.south west);
  \draw[red!75!black,line width=3pt]
    ([xshift=-5pt]frame.north east) -- 
    (frame.north east) -- 
    (frame.south east);
  },
  overlay unbroken app={
  \draw[red!75!black,line width=3pt,line cap=rect]
    (frame.south west) -- 
    ([xshift=5pt]frame.south west);
  \draw[red!75!black,line width=3pt,line cap=rect]
    (frame.south east) -- 
    ([xshift=-5pt]frame.south east);
  },
  overlay middle and last={
  \draw[red!75!black,line width=3pt]
    (frame.north west) -- 
    (frame.south west);
  \draw[red!75!black,line width=3pt]
    (frame.north east) -- 
    (frame.south east);
  },
  overlay last app={
  \draw[red!75!black,line width=3pt,line cap=rect]
    (frame.south west) --
    ([xshift=5pt]frame.south west);
  \draw[red!75!black,line width=3pt,line cap=rect]
    (frame.south east) --
    ([xshift=-5pt]frame.south east);
  },
}

\definecolor{myblue}{rgb}{0.9, 0.1, 0.94}
\definecolor{mygreen}{rgb}{0.64, 0.56, 0.88}
\definecolor{myyellow}{rgb}{0.68, 0.6, 0.1}
\definecolor{fancygreen}{rgb}{0.33, 0.68, 0.20}
\definecolor{salmon}{rgb}{0.94, 0.52, 0.49}
\definecolor{tablegreen}{rgb}{0.82, 0.94, 0.75}
\definecolor{tableblue}{rgb}{0.81, 0.90, 0.94}
\definecolor{tablered}{rgb}{0.97, 0.85, 0.85}
\definecolor{tableorange}{rgb}{0.96, 0.85, 0.81}

\definecolor{myorange}{rgb}{1.0, 0.49, 0.0}	
\definecolor{tlgreen}{rgb}{0.33, 0.68, 0.20}

\newenvironment{itemize*}%
 {\leftmargini=10pt\begin{itemize}%
  \setlength{\itemsep}{0pt}%
  \setlength{\parskip}{0pt}%
  }%
 {\end{itemize}}
\newenvironment{enumerate*}%
 {\begin{enumerate}%
  \setlength{\itemsep}{0pt}%
  \setlength{\parskip}{0pt}}%
 {\end{enumerate}}

\tikzset{%
    every node/.style={font=\tiny},
    parent/.style =          {align=center,text width=2cm,rounded corners=3pt, line width=0.3mm, fill=gray!10,draw=gray!80},
    child/.style =           {align=center,text width=2.0cm,rounded corners=3pt, fill=blue!10,draw=blue!80,line width=0.3mm},
    grandchild/.style =      {align=center,text width=2cm,rounded corners=3pt},
    greatgrandchild/.style = {align=center,text width=1.5cm,rounded corners=3pt},
    greatgrandchild2/.style = {align=center,text width=1.5cm,rounded corners=3pt},    
    referenceblock/.style =  {align=center,text width=1.5cm,rounded corners=2pt},
    pretrain/.style =           {align=center,text width=2.0cm,rounded corners=3pt, fill=blue!10,draw=blue!80,line width=0.3mm},   
    pretrain_work/.style =           {align=center, text width=8.5cm,rounded corners=3pt, fill=blue!10,draw=blue!0,line width=0.3mm},  
    template/.style =           {align=center,text width=2.0cm,rounded corners=3pt, fill=red!10,draw=red!80,line width=0.3mm},   
    template_work/.style =           {align=center,text width=8.5cm,rounded corners=3pt, fill=red!10,draw=red!0,line width=0.3mm},    
    answer/.style =           {align=center,text width=2.0cm,rounded corners=3pt, fill= cyan!10,draw= cyan!80,line width=0.3mm},   
    answer_work/.style =           {align=center,text width=8.5cm,rounded corners=3pt, fill= cyan!10,draw= cyan!0,line width=0.3mm},      
    multiple/.style =           {align=center,text width=2.0cm,rounded corners=3pt, fill= orange!10,draw= orange!80,line width=0.3mm},   
    multiple_work/.style =           {align=center,text width=8.5cm,rounded corners=3pt, fill= orange!10,draw= orange!0,line width=0.3mm},        
    tuning/.style =           {align=center,text width=2.0cm,rounded corners=3pt, fill= magenta!10,draw= magenta!80,line width=0.3mm},   
    tuning_work/.style =           {align=center,text width=8.5cm,rounded corners=3pt, fill= magenta!10,draw= magenta!0,line width=0.3mm},          
}

\usepackage{tikz}

\newcommand{\lstbg}[3][0pt]{{\fboxsep#1\colorbox{#2}{\strut #3}}}

\lstdefinelanguage{diff}{
  basicstyle=\ttfamily\small,
  morecomment=[f][\lstbg{red!20}]-,
  morecomment=[f][\lstbg{green!20}]+,
}
\lstdefinelanguage{diffpython}{
  language=diff,
  morekeywords={def, if, else, for, while, return, import, from, as, class, with, try, except, finally, raise, lambda, and, or, not, in, is, None, True, False},
  morecomment=[l]{\#},
  morestring=[b]",
  morestring=[b]',
}

\definecolor{darkgreen}{RGB}{50,100,0}
\definecolor{darkred}{RGB}{200, 0, 0}
\definecolor{lightblue}{RGB}{220,235,250}

\tcbset{
  takeawaysbox/.style={
    title=Takeaways,
    colback=lightblue!80,
    colframe=black,
    fonttitle=\bfseries\small,
    coltitle=white,
    colbacktitle=black,
    enhanced,
    attach boxed title to top left={xshift=2.5mm,yshift=-2.5mm},
    boxed title style={rounded corners, size=small, colframe=black, colback=black},
    width=\linewidth,
    arc=3.5mm
  }
}

\definecolor{darkgreen}{RGB}{50,100,0}
\definecolor{darkred}{RGB}{200, 0, 0}

\NewDocumentCommand{\kaiyan}
{ mO{} }{\textcolor{purple}{\textsuperscript{\textit{kaiyan}}\textsf{\textbf{\small[#1]}}}}
\NewDocumentCommand{\yuxin}
{ mO{} }{\textcolor{cyan}{\textsuperscript{\textit{yuxin}}\textsf{\textbf{\small[#1]}}}}
\NewDocumentCommand{\bx}
{ mO{} }{\textcolor{green}{\textsuperscript{\textit{bx}}\textsf{\textbf{\small[#1]}}}}
\NewDocumentCommand{\at}
{ mO{} }{\textcolor{red}{\textsuperscript{\textit{AT}}\textsf{\textbf{\small[#1]}}}}
\NewDocumentCommand{\re}
{ mO{} }{\textcolor{blue}{\textsuperscript{\textit{RE}}\textsf{\textbf{\small[#1]}}}}
\NewDocumentCommand{\ybsun}
{ mO{} }{\textcolor{magenta}{\textsuperscript{\textit{youbang}}\textsf{\textbf{\small[#1]}}}}
\NewDocumentCommand{\runze}
{ mO{} }{\textcolor{orange}{\textsuperscript{\textit{runze}}\textsf{\textbf{\small[#1]}}}}

\definecolor{darkgreen}{RGB}{0,100,0} 
\NewDocumentCommand{\add}
{ mO{} }{\textcolor{darkgreen}{\textsuperscript{\textit{Maybe Consider Discuss}}\textsf{\textbf{[#1]}}}}

\setlist[itemize]{leftmargin=20pt}

\usepackage[edges]{forest}
\definecolor{hidden-blue}{RGB}{194,232,247}
\definecolor{hidden-black}{RGB}{20,68,106}

\usepackage{tabularx, booktabs, xcolor, colortbl}
\usepackage{pifont}   %
\usepackage{xstring}  %

\usepackage{multirow}
\usepackage{hyperref}
\usepackage{fontawesome5}

\usepackage[export]{adjustbox}

\definecolor{yes}{HTML}{C6EFCE}      %
\definecolor{no}{HTML}{FFC7CE}       %
\definecolor{partial}{HTML}{FFEB9C}  %
\definecolor{external}{HTML}{D9E1F2} %
\definecolor{hdr}{HTML}{F2F2F2}

\usepackage{amssymb}
\newcommand{\cmark}{\textcolor{darkgreen}{\boldmath$\checkmark$}}
\newcommand{\xmark}{\textcolor{darkred}{\boldmath$\times$}}

\newcommand{\cellstatus}[1]{%
  \begingroup
  \StrTrim{#1}[\statusval]%
  \IfStrEq{\statusval}{Yes}{\cellcolor{yes}\cmark}{}%
  \IfStrEq{\statusval}{No}{\cellcolor{no}\xmark}{}%
  \IfBeginWith{\statusval}{Yes (}{\cellcolor{yes}\cmark~\textit{\statusval\unskip}}{}%
  \IfStrEq{\statusval}{Partial}{\cellcolor{partial}\textbf{Partial}}{}%
  \IfStrEq{\statusval}{External}{\cellcolor{external}\textbf{External}}{}%
  \endgroup
}

\newcommand{\faHuggingFace}{%
  \raisebox{-0.13em}{%
    \includegraphics[height=1em]{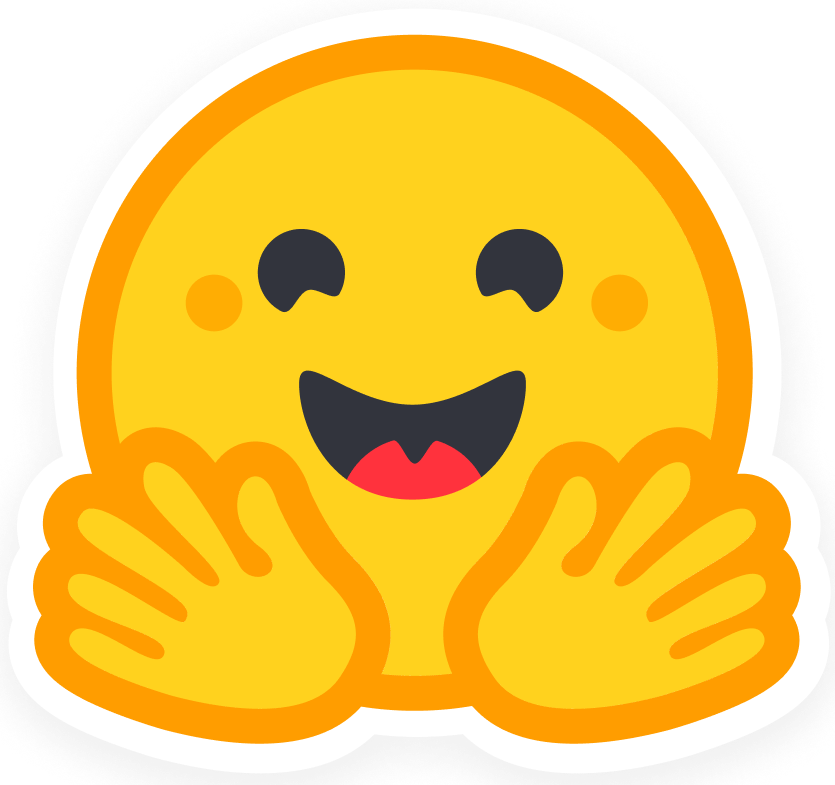}%
  }%
}

\newcommand{\passk}{\emph{Pass@K}}
\newcommand{\passo}{\emph{Pass@1}}
\newcommand{\tstyle}[1]{\underline{\textit{#1}}}

\title{A Survey of Reinforcement Learning for Large Reasoning Models}

\author{%
    Kaiyan Zhang$^{1*\dagger}$, Yuxin Zuo$^{1*\dagger}$, Bingxiang He$^{1*}$, Youbang Sun$^{1*}$, Runze Liu$^{1*}$, Che Jiang$^{1*}$, Yuchen Fan$^{2,3*}$, \quad Kai Tian$^{1*}$, Guoli Jia$^{1*}$, Pengfei Li$^{2,6*}$, Yu Fu$^{9*}$, Xingtai Lv$^{1*}$, Yuchen Zhang$^{2,4*}$, Sihang Zeng$^{7*}$, Shang Qu$^{1,2*}$, Haozhan Li$^{1*}$, Shijie Wang$^{2*}$, Yuru Wang$^{1*}$, Xinwei Long$^{1}$, Fangfu Liu$^{1}$, Xiang Xu$^{5}$, Jiaze Ma$^{1}$, Xuekai Zhu$^{3}$, Ermo Hua$^{1,2}$, Yihao Liu$^{1,2}$, Zonglin Li$^{2}$, Huayu Chen$^{1}$, Xiaoye Qu$^{2}$, Yafu Li$^{2}$, Weize Chen$^{1}$, Zhenzhao Yuan$^{1}$, Junqi Gao$^{6}$, Dong Li$^{6}$, Zhiyuan Ma$^{8}$, Ganqu Cui$^{2}$, Zhiyuan Liu$^{1}$, Biqing Qi$^{2\ddagger}$, Ning Ding$^{1,2\ddagger}$, Bowen Zhou$^{1,2\ddagger}$
    \vspace{1mm} \\
    $^1$ Tsinghua University \quad
    $^2$ Shanghai AI Laboratory \quad
    $^3$ Shanghai Jiao Tong University \quad
    $^4$ Peking University \\
    $^5$ University of Science and Technology of China \quad
    $^6$ Harbin Institute of Technology \quad
    $^7$ University of Washington \\
    $^8$ Huazhong University of Science and Technology \quad
    $^9$ University College London  \quad
    \vspace{1mm} \\
    \textbf{$^\dagger$ Project Lead.}~~ $^*$ \textbf{Core Contributors.}~~  \textbf{$^\ddagger$ Corresponding Authors.}
    \vspace{1mm} \\
    \faEnvelope[regular]~\texttt{zhang-ky22@mails.tsinghua.edu.cn}  \quad
    \faGithub~\href{https://github.com/TsinghuaC3I/Awesome-RL-for-LRMs}{TsinghuaC3I/Awesome-RL-for-LRMs}
}

\begin{abstract}
In this paper, we survey recent advances in Reinforcement Learning (RL) for reasoning with Large Language Models (LLMs). RL has achieved remarkable success in advancing the frontier of LLM capabilities, particularly in addressing complex logical tasks such as mathematics and coding. As a result, RL has emerged as a foundational methodology for transforming LLMs into LRMs. With the rapid progress of the field, further scaling of RL for LRMs now faces foundational challenges not only in computational resources but also in algorithm design, training data, and infrastructure. To this end, it is timely to revisit the development of this domain, reassess its trajectory, and explore strategies to enhance the scalability of RL toward Artificial SuperIntelligence (ASI). In particular, we examine research applying RL to LLMs and LRMs for reasoning abilities, especially since the release of DeepSeek-R1, including foundational components, core problems, training resources, and downstream applications, to identify future opportunities and directions for this rapidly evolving area. We hope this review will promote future research on RL for broader reasoning models.
\end{abstract}

\begin{document}

\maketitle

\begin{figure}[h]
\centering
\includegraphics[width=\textwidth]{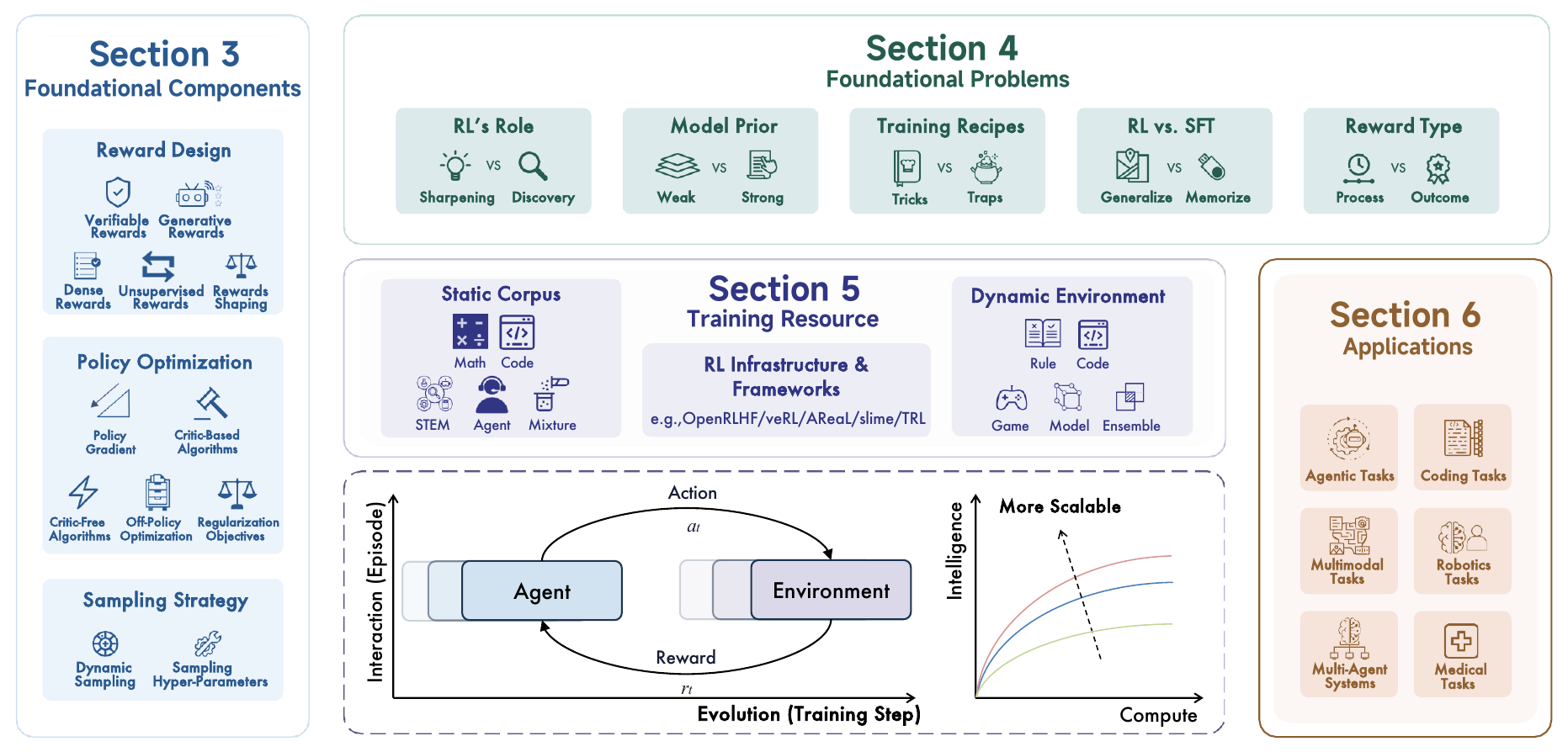}
\caption{
Overview of the survey.
We introduce the foundational components of RL for LRMs, along with open problems, training resources, and applications. Central to this survey is a focus on large-scale interactions between language agents and environments throughout long-term evolution.
}
\label{fig:teaser}
\end{figure}

\newpage
\begingroup
\setlength{\baselineskip}{1.25\baselineskip}
\tableofcontents
\endgroup

\newpage

\section{Introduction}

Reinforcement Learning (RL)~\citep{sutton1998introduction} has repeatedly demonstrated that narrow, well-specified reward signals can drive artificial agents to superhuman competence on complex tasks. Landmark systems such as AlphaGo~\citep{silver2016mastering} and AlphaZero~\citep{silver2017mastering}, which learned exclusively through self-play and reward feedback, surpassed world champions in Go, chess, shogi and Stratego~\citep{silver2018general,schrittwieser2020mastering,perolat2022mastering}, establishing RL as a practical and promising technology for high-level problem solving.
In the era of Large Language Models (LLMs)~\citep{zhao2023survey}, RL initially rose to prominence as a post-training strategy for human alignment~\citep{ouyang2022training}. Widely adopted methods such as Reinforcement Learning from Human Feedback (RLHF)~\citep{christiano2017deep} and Direct Preference Optimization (DPO)~\citep{rafailov2023direct} finetune pre-trained models to follow instructions and reflect human preferences, markedly improving helpfulness, honesty, and harmlessness (3H)~\citep{bai2022constitutional}.

More recently, a new trend has emerged: RL for Large Reasoning Models (LRMs)~\citep{xu2025towards}, which aims not merely to align behavior but to incentivize reasoning itself. 
Two recent milestones (i.e., OpenAI o1~\citep{jaech2024openai} and DeepSeek-R1~\citep{guo2025deepseek}) demonstrate that training LLMs using reinforcement learning with verifiable rewards (RLVR), such as answer correctness for mathematics or unit-test pass rates for code, can enable models to perform long-form reasoning, including planning, reflection, and self-correction.
OpenAI reports~\citep{jaech2024openai} that o1's performance improves smoothly with both additional RL (increased train-time compute) and more time spent ``thinking'' at inference (test-time compute)~\citep{snell2024scaling,brown2024large,liu2025can1b}, revealing a new scaling axis beyond pre-training alone~\citep{kaplan2020scaling,aghajanyan2023scaling}.
DeepSeek-R1~\citep{guo2025deepseek} employs explicit, rule-based accuracy rewards for mathematics, as well as compiler- or test-based rewards for coding tasks. This approach demonstrates that large-scale reinforcement learning, specifically, Group Relative Policy Optimization (GRPO), can induce sophisticated reasoning behaviors even in base models prior to subsequent alignment stages.

This shift reframes reasoning as a capability that can be explicitly trained and scaled~\citep{openai-o3,openai-gpt5}: LRMs allocate significant test-time compute to generate, evaluate, and revise intermediate chain-of-thought~\citep{wei2022chain}, and their performance rises as this compute budget increases. This dynamic introduces a complementary path to capability gains, orthogonal to data and parameter scaling during pre-training~\citep{kaplan2020scaling,aghajanyan2023scaling}, while leveraging a reward maximization objective~\citep{silver2021reward}, automatically checkable rewards wherever reliable verifiers exist (e.g., competition mathematics~\citep{jaech2024openai,guo2025deepseek}, competitive programming~\citep{el2025competitive}, and selected scientific domains~\citep{bai2025intern}). Furthermore, RL can overcome data limitations~\citep{villalobos2022will,shumailov2024ai} by enabling self-generated training data~\citep{silver2018general,zhao2025absolute}. As a result, RL is increasingly regarded as a promising technology for achieving Artificial SuperIntelligence (ASI) on a broader range of tasks through continual scaling.

At the same time, further scaling of RL for LRMs introduces new constraints, not only in computational resources, but also in algorithm design, training data, and infrastructure. How and where RL for LRMs can be scaled to achieve high-level intelligence and generate real-world value remain unresolved issues. Therefore, we argue that it is timely to revisit the development of this domain and explore strategies to enhance the scalability of RL toward artificial superintelligence.
In summary, this survey reviews recent work on RL for LRMs as follows:
\begin{itemize}
    \item We introduce the preliminary definitions of RL modeling in the context of LRMs ($\S$~\ref{sec:pre_background}) and outline the development of frontier reasoning models since the release of OpenAI o1 ($\S~\ref{sec:pre_frontier}$).
    \item We review recent literature on the foundational components of RL for LRMs, including reward design ($\S$~\ref{sec:reward}), policy optimization ($\S$~\ref{sec:policy}), and sampling strategies ($\S$~\ref{sec:sampling}), comparing the different research directions and technical approaches for each component.
    \item We discuss foundational and still controversial problems in RL for LRMs ($\S$~\ref{sec:problems}), such as the role of RL ($\S$~\ref{sec:problems_role}), RL versus Supervised Fine-Tuning (SFT) ($\S$~\ref{sec:problems_rl_vs_sft}), model priors ($\S$~\ref{sec:problems_model_prior}), training recipes ($\S$~\ref{sec:problems_training_recipes}), and reward definitions ($\S$~\ref{sec:problems_reward_type}). We argue that these issues warrant further exploration to enable continued scaling of RL.
    \item We examine training resources for RL ($\S$~\ref{sec:resource}), including static corpora ($\S$~\ref{sec:resource_static_corpus}), dynamic environments ($\S$~\ref{sec:resource_dynamic_environment}), and training infrastructure ($\S$~\ref{sec:resource_rl_infra}). While these resources are reusable in both research and production, further standardization and development are needed.
    \item We review applications of RL to a wide range of tasks ($\S$~\ref{sec:application}), such as coding tasks ($\S$~\ref{sec:application_coding}), agentic tasks ($\S$~\ref{sec:application_agentic}), multimodal tasks ($\S$~\ref{sec:application_multimodal}), multi-agent systems ($\S$~\ref{sec:application_mas}), robotics tasks ($\S$~\ref{sec:application_robotics}), and medical applications ($\S$~\ref{sec:application_medical}).
    \item Finally, we discuss future directions in RL for language models ($\S$~\ref{sec:future}), covering novel algorithms, mechanisms, features, and additional research avenues.
\end{itemize}

\begin{figure}[!t]
\centering
\includegraphics[width=\textwidth]{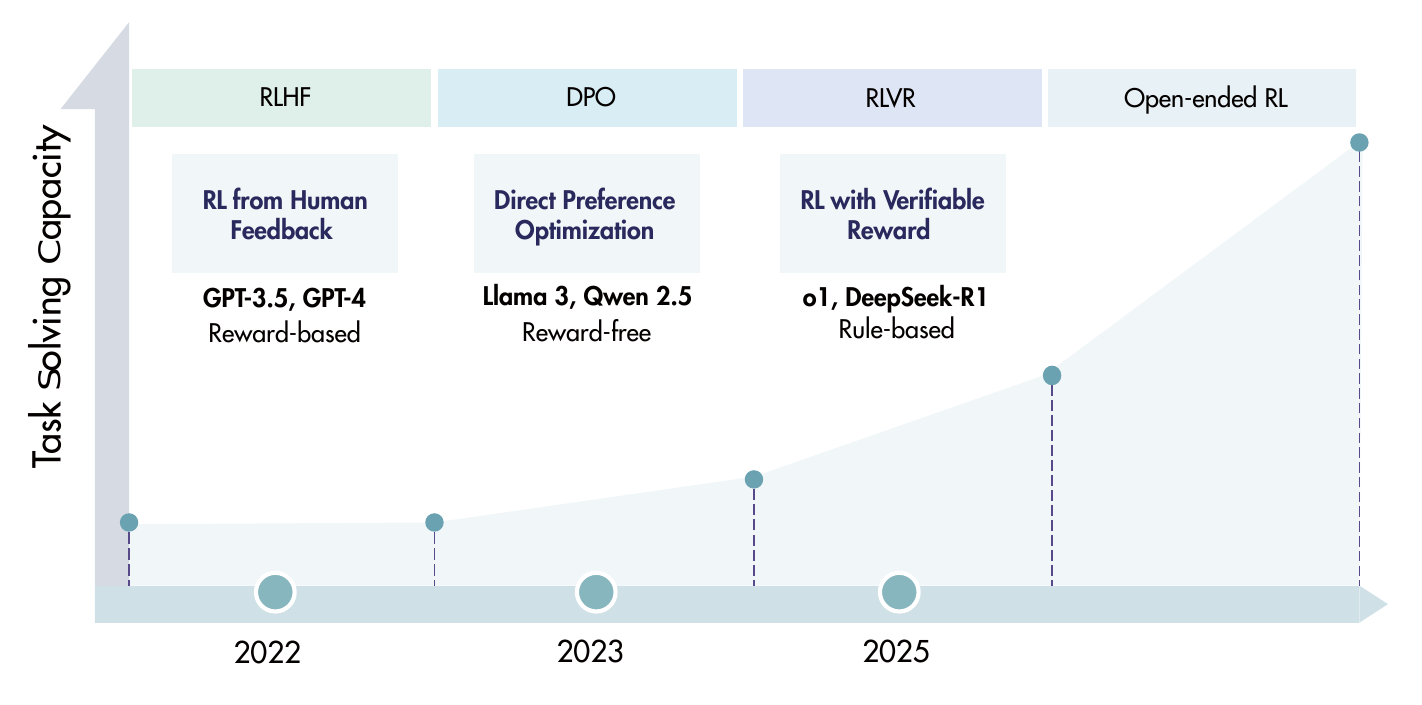}
\caption{
RLHF and DPO have been the two predominant RL methodologies for human alignment in recent years. In contrast, RLVR represents an emerging trend in RL for LRMs, significantly enhancing their capacity for complex task solving. The next stage of scaling RL for LLMs remains an open question, with open-ended RL presenting a particularly challenging and promising direction.
}
\label{fig:rl_evol}
\end{figure}

\section{Preliminaries}
\label{sec:pre}

\subsection{Background}
\label{sec:pre_background}

\begin{figure}[!t]
\centering
\includegraphics[width=\textwidth]{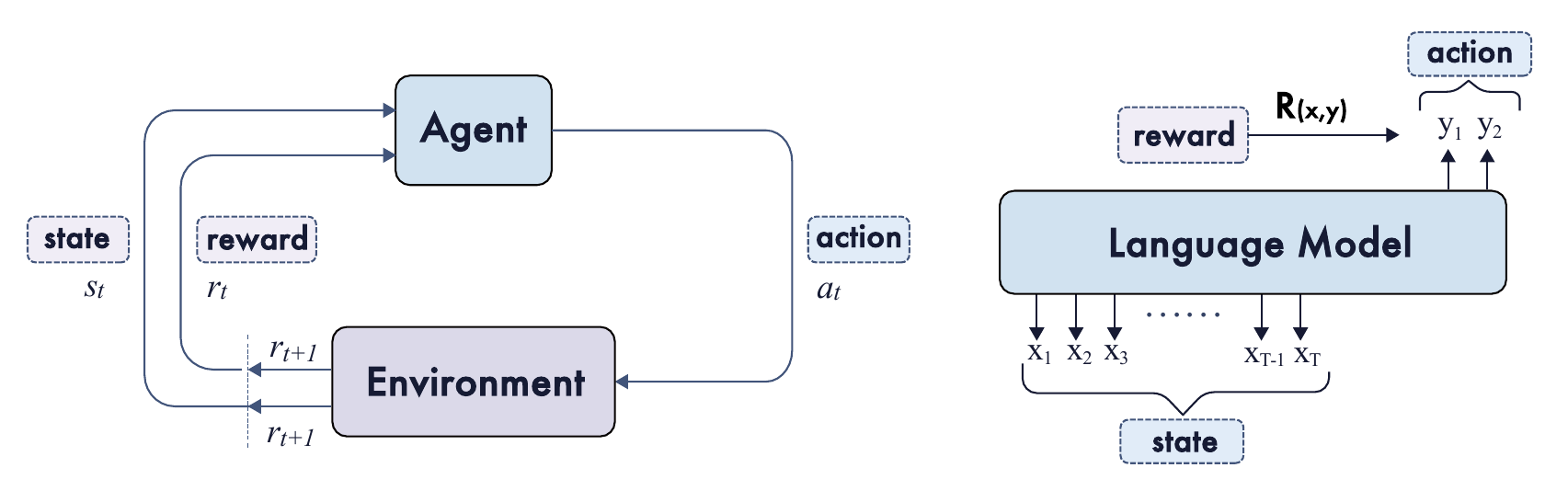}
\caption{Basic components of RL and language models (LMs) as agents. The agent selects actions, while the environment provides states and rewards at each turn. In the context of LMs, completion tokens are treated as actions, which are concatenated with the context to form the state. Rewards are typically assigned at the level of the entire response.}
\label{fig:rl_lm_agent}
\end{figure}

In this subsection, we introduce the basic components of RL and describe how language models can be configured as agents within RL frameworks.
As shown in Figure~\ref{fig:rl_lm_agent}, RL provides a general framework for sequential decision making, in which an agent interacts with an environment by taking actions to maximize cumulative reward.
In classical RL, the problem is typically formulated as a  Markov Decision Process (MDP)~\citep{sutton1998introduction}, which is defined by a tuple $(\mathcal{S}, \mathcal{A}, \mathcal{P}, R, \gamma)$. 
The main components include a state space $\mathcal{S}$, an action space $\mathcal{A}$, transition dynamics $\mathcal{P}: \mathcal{S} \times \mathcal{A} \mapsto \mathcal{S}$, a reward function $R: \mathcal{S} \times \mathcal{A} \mapsto \mathbb{R}$, and a discount factor $\gamma \in [0, 1]$. 
At each step, the agent observes a state $s_t$, selects an action $a_t$ according to its policy $\pi_\theta$ parameterized by $\theta$, receives a reward $r_t$, and transits to the next state $s_{t+1}$.
When applying RL to language models, these concepts can be naturally mapped to the language domain with minimal adaptation. The mapping is summarized as follows:
\begin{itemize}
	\item \textbf{Prompt/Task ($x$)}:
    Corresponds to the initial state or environment context, drawn from a data distribution and corresponding to the dataset $\mathcal{D}$.
    \item \textbf{Policy ($\pi_\theta$)}:
    Represents the language model, which generates a sequence of length $T$ denoting as $y = (y_1, \ldots, y_T)$ in response to the prompt.
    \item \textbf{State ($s_t$)}:
    Defined as the prompt together with the tokens generated so far, i.e., $s_{t} = (x, a_{1:t-1}$).
    \item \textbf{Action ($a_t$)}:
    The unit chosen at step $t$ from the action space $\mathcal{A}$. Depending on the granularity, the action may be an entire sequence $y$ (sequence-level), a token $a_t \in \mathcal{V}$ (token-level), or a segment $y^{(k)} = (y_1^{(k)}, \ldots, y_{T_k}^{(k)})$ (step-level), with a detailed comparison in Table~\ref{tab:comparison_action_reward_definition}.
    \item \textbf{Transition Dynamics ($\mathcal{P}$)}:
    The state transition is usually deterministic in the context of LLMs since $s_{t+1} = [s_t, a_t]$, where $[\cdot, \cdot]$ denotes string concatenation. When the state contains an \texttt{EOS} token, the policy transits to a terminal state, meaning the trajectory ends.
    \item \textbf{Reward ($R(x, y)$ or $r_t$)}:
    Assigned based on the action granularity, e.g., sequence-level $R(x, y)$ at trajectory end, token-level $r_t = R(x, a_{1:t})$ per token, or step-level $r_k = R(x, y^{(1:k)})$ per segment.
    \item \textbf{Return ($G$)}:
    The cumulative reward of the whole trajectory $y$ for prompt $x$ (typically with $\gamma=1$ for finite horizons). It reduces to the single scalar $R(x,y)$ with sequence-level reward, or aggregates per-token/step rewards otherwise, as detailed in Table~\ref{tab:comparison_action_reward_definition}.
\end{itemize}

In this setting, the learning objective~\citep{sutton1998introduction} is to maximize the expected cumulative reward over the data distribution $\mathcal{D}$, that is,
\begin{equation}
    \max_{\theta} \mathcal{J} (\theta) := \mathbb{E}_{x \sim \mathcal{D}, y \sim \pi_{\theta} (x)} [G].
\end{equation}

In practice, it is common to regularize the learned policy towards a reference policy $\pi_\text{ref}$, often implemented as KL-divergence constraints to stabilize training and maintain language quality.
In the following sections, we present various algorithms that build upon this fundamental formulation.

\subsection{Frontier Models}
\label{sec:pre_frontier}

In this subsection, we provide an overview of state-of-the-art large reasoning models trained with RL-like methods, organized roughly chronologically along three major directions: LRMs, agentic LRMs, and multimodal LRMs.

Over the past year, RL has progressively expanded the frontier of reasoning models and their applications. The first large reasoning models, OpenAI's o1~\citep{jaech2024openai} series, established the effectiveness of scaling both train-time RL and test-time compute towards more powerful reasoning abilities, achieving leading results on mathematics, coding, and science benchmarks. DeepSeek's flagship model R1~\citep{guo2025deepseek} followed as the first open-source model to match o1's performance across benchmarks. It employs a multi-stage training pipeline to ensure well-rounded model abilities, and explores the route of pure RL without supervised finetuning (i.e., Zero RL). Other proprietary model releases promptly followed: Claude-3.7-Sonnet~\citep{anthropic2025claude} featured hybrid reasoning, Gemini 2.0 and 2.5~\citep{comanici2025gemini25} introduced longer context lengths, Seed-Thinking 1.5~\citep{seed_seed15-thinking_2025} featured generalization across domains, and the o3~\citep{openai-o3} series showcased increasingly advanced reasoning abilities. Recently, OpenAI introduced their first open-source reasoning model gpt-oss-120b~\citep{openai2025gpt-oss-120b}, and subsequently GPT5~\citep{openai-gpt5}, their most capable AI system to date, which flexibly switches between an efficient model and a deeper reasoning model GPT-5 thinking.
Parallel open-source efforts continued to expand the landscape. Within the Qwen family, QwQ-32B~\citep{qwq32b} matched R1's performance, and was followed by the Qwen3~\citep{yang2025qwen3} series, with the representative model Qwen3-235B further improving benchmark scores. The Skywork-OR1~\citep{he2025skywork} suite of models were based on R1-distilled models, and achieved scalable RL training through effective data mixtures and algorithmic innovations. Minimax-M1~\citep{chen2025minimax} was the first model to introduce hybrid attention to scale RL efficiently. Other works include Llama-Nemotron-Ultra~\citep{bercovich2025llama-nemotron}, which aimed to balance accuracy and efficiency; Magistral 24B~\citep{mistralai2025magistral}, trained through RL from scratch without distillation from prior models; and Seed-OSS~\citep{seed2025seed-oss}, emphasizing long-context reasoning abilities.

Model reasoning improvements have in turn extended their use cases in coding and agentic scenarios. The Claude series has been known for their leading performance on agentic coding tasks, and this was exemplified by Claude-4.1-Opus~\citep{anthropic2025claude41}, which further pushed forward the state-of-the-art results on SWE-bench~\citep{jimenez2023swe}. Kimi K2~\citep{team2025kimi} is a recent representative agentic model which was specifically optimized for agentic tasks, forging large-scale agentic training data synthesis and a general RL procedure that accommodates non-verifiable rewards. Shortly after, both the GLM4.5~\citep{zeng2025glm} and DeepSeek-V3.1 releases emphasized tool-use and agentic tasks, showing substantial improvements on relevant benchmarks.

\begin{figure}[!t]
\centering

\includegraphics[width=1.0\textwidth]{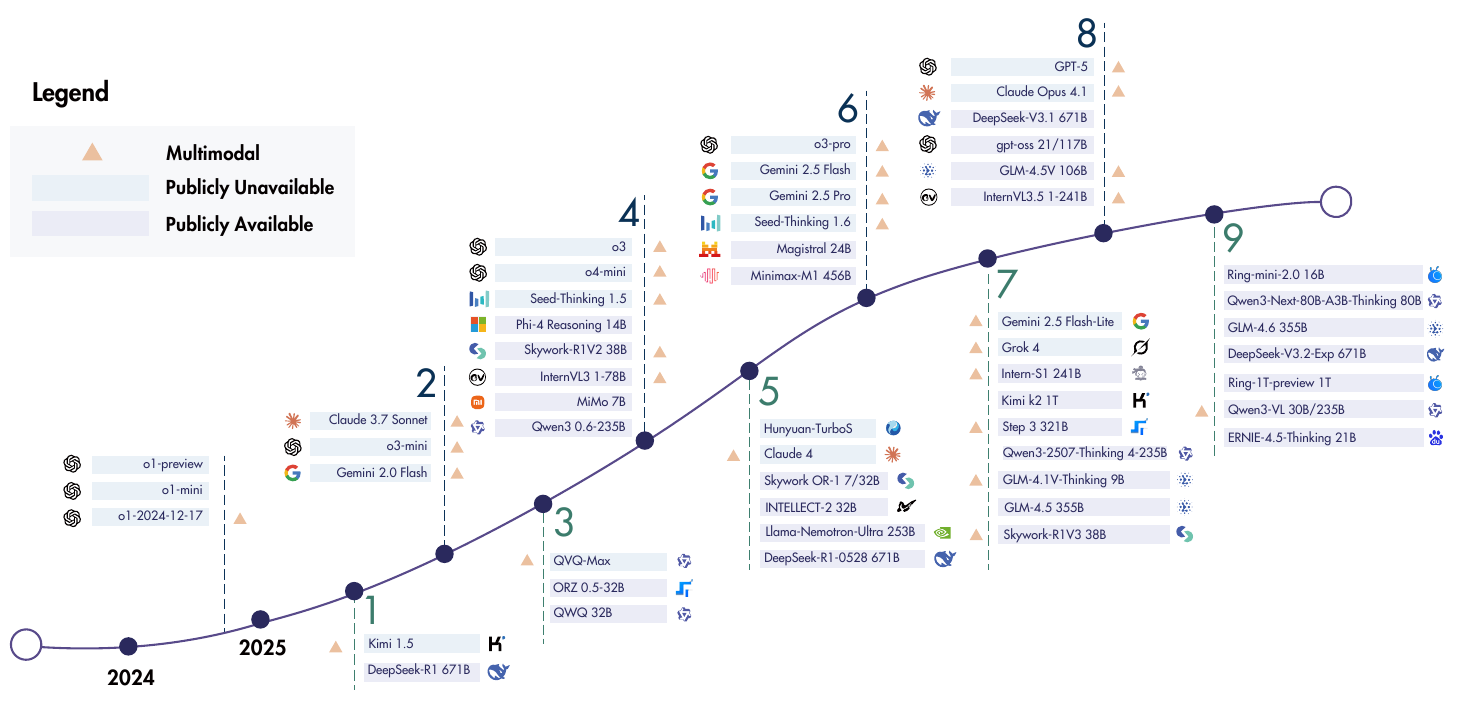}
\caption{
Timeline of representative open-source and closed-source reasoning models trained with RL, including language models, multimodal models, and agentic models.
}
\label{fig:models}
\end{figure}

Multimodality is a key component behind the widespread adoption of reasoning models. Most frontier proprietary models, including GPT-5, o3, Claude, and Gemini families, are natively multimodal. Gemini-2.5~\citep{comanici2025gemini25} notably emphasized strong performance across text, images, video, and audio. On the open-source side, Kimi 1.5~\citep{team2025kimi} represents an early effort towards multimodal reasoning, highlighting long context scaling as well as joint reasoning over text and vision domains. QVQ~\citep{qvq} excels in visual reasoning and analytical thinking, while Skywork R1V2~\citep{wang_skywork_2025} balances reasoning and general abilities through hybrid RL, using both MPO and GRPO. As notable additions to the InternVL series, InternVL3~\citep{zhu_internvl3_2025} adopted a unified native multimodal pretraining phase, and later InternVL3.5~\citep{wang_internvl35_2025} used a two-stage cascade RL framework, achieving improved efficiency and versatility. More recently, the Intern-S1~\citep{bai2025intern} model focused on multimodal scientific reasoning across diverse domains, benefiting from a mixture-of-rewards design during online RL to facilitate simultaneous training on a wide range of tasks. Other recent models include Step3~\citep{wang2025step}, designed for efficient training and minimizing decoding costs, and GLM-4.5V~\citep{team_glm-45v_2025}, with state-of-the-art performance across most visual multimodal benchmarks.
MiniCPM-V 4.5~\citep{yu2025minicpm} is an 8B model that achieves high efficiency and strong performance through optimized architecture, data strategy, and RL-based training methods.

In addition to the aforementioned models, we provide a comprehensive list of reasoning models in Figure~\ref{fig:models} and detailed information on open-source models in Table~\ref{tab:models}.

\begingroup
\footnotesize
\setlength{\tabcolsep}{3pt} %
\begin{longtable}{
  @{}>{\raggedright\arraybackslash}p{0.09\textwidth}   %
  @{\hspace{2pt}}>{\raggedright\arraybackslash}p{0.20\textwidth} %
  @{\hspace{2pt}}>{\raggedright\arraybackslash}p{0.14\textwidth} %
  @{\hspace{2pt}}>{\raggedright\arraybackslash}p{0.13\textwidth} %
  @{\hspace{2pt}}>{\raggedright\arraybackslash}p{0.14\textwidth} %
  @{\hspace{2pt}}>{\raggedright\arraybackslash}p{0.10\textwidth} %
  @{\hspace{2pt}}>{\centering\arraybackslash}p{0.10\textwidth}    %
  @{\hspace{2pt}}>{\raggedright\arraybackslash}p{0.07\textwidth}@{} %
}
\caption{
Comparison of representative open-source models trained with RL. OPMD denotes Online Policy Mirror Descent; MPO denotes Mixed Preference Optimization; CISPO denotes Clipped IS-weight Policy Optimization. T, I, and V indicate Text, Image, and Video modalities, respectively.
}
\label{tab:models} \\
\toprule
\textbf{Date} & \textbf{Model} & \textbf{Organization} & \textbf{Architecture} & \textbf{Parameters} & \textbf{Algorithm} & \textbf{Modal} & \textbf{Link} \\
\midrule
\endfirsthead

\multicolumn{8}{l}{\tablename\ \thetable\ -- \textit{Continued from previous page}}\\
\toprule
\textbf{Date} & \textbf{Model} & \textbf{Organization} & \textbf{Architecture} & \textbf{Parameters} & \textbf{Algorithm} & \textbf{Modal} & \textbf{Link} \\
\midrule
\endhead

\midrule \multicolumn{8}{r}{\textit{Continued on next page}}\\
\endfoot
\bottomrule
\endlastfoot

\multirow{2}{*}{2025.01} & DeepSeek-R1 & \multirow{2}{*}{DeepSeek} & \multirow{2}{*}{MoE/MLA} & \multirow{2}{*}{671B} & \multirow{2}{*}{GRPO} & \multirow{2}{*}{Text} & \multirow{2}{*}{\href{https://github.com/deepseek-ai/DeepSeek-R1}{\faGithub}\;
\href{https://huggingface.co/deepseek-ai/DeepSeek-R1}{\faHuggingFace}} \\
& \citep{guo2025deepseek} & & & & & & \\
\midrule
\multirow{2}{*}{2025.03} & ORZ & \multirow{2}{*}{StepAI} & \multirow{2}{*}{Dense} & \multirow{2}{*}{0.5-32B} & \multirow{2}{*}{PPO} & \multirow{2}{*}{Text} & \multirow{2}{*}{\href{https://github.com/Open-Reasoner-Zero/Open-Reasoner-Zero}{\faGithub}\;
\href{https://huggingface.co/Open-Reasoner-Zero}{\faHuggingFace}} \\
& \citep{hu2025open} & & & & & & \\
\midrule
\multirow{2}{*}{2025.03} & QwQ & \multirow{2}{*}{Alibaba Qwen} & \multirow{2}{*}{Dense} & \multirow{2}{*}{32B} & \multirow{2}{*}{-} & \multirow{2}{*}{Text} & \multirow{2}{*}{\href{https://github.com/QwenLM/QwQ}{\faGithub}\;
\href{https://huggingface.co/Qwen/QwQ-32B}{\faHuggingFace}} \\
& \citep{qwq32b} & & & & & & \\
\midrule
\multirow{2}{*}{2025.04} & Phi-4 Reasoning & \multirow{2}{*}{Microsoft} & \multirow{2}{*}{Dense} & \multirow{2}{*}{14B} & \multirow{2}{*}{GRPO} & \multirow{2}{*}{Text} & \multirow{2}{*}{\href{https://github.com/marketplace/models/azureml/Phi-4-reasoning}{\faGithub}\;
\href{https://huggingface.co/microsoft/Phi-4-reasoning}{\faHuggingFace}} \\
& \citep{abdin2025phi} & & & & & & \\
\midrule
\multirow{2}{*}{2025.04} & Skywork-R1V2 & \multirow{2}{*}{Skywork} & \multirow{2}{*}{Dense} & \multirow{2}{*}{38B} & \multirow{2}{*}{MPO/GRPO} & \multirow{2}{*}{T/I} & \multirow{2}{*}{\href{https://github.com/SkyworkAI/Skywork-R1V}{\faGithub}\;
\href{https://huggingface.co/Skywork/Skywork-R1V2-38B}{\faHuggingFace}} \\
& \citep{wang_skywork_2025} & & & & & & \\
\midrule
\multirow{2}{*}{2025.04} & InternVL3 & \multirow{2}{*}{Shanghai AI Lab} & \multirow{2}{*}{Dense} & \multirow{2}{*}{1-78B} & \multirow{2}{*}{MPO} & \multirow{2}{*}{T/I/V} & \multirow{2}{*}{\href{https://github.com/OpenGVLab/InternVL}{\faGithub}\;
\href{https://huggingface.co/spaces/OpenGVLab/InternVL}{\faHuggingFace}} \\
& \citep{zhu_internvl3_2025} & & & & & & \\
\midrule
\multirow{2}{*}{2025.04} & MiMo & \multirow{2}{*}{Xiaomi} & \multirow{2}{*}{Dense} & \multirow{2}{*}{7B} & \multirow{2}{*}{GRPO} & \multirow{2}{*}{Text} & \multirow{2}{*}{\href{https://github.com/XiaomiMiMo/MiMo}{\faGithub}\;
\href{https://huggingface.co/XiaomiMiMo}{\faHuggingFace}} \\
& \citep{xiaomi2025mimo} & & & & & & \\
\midrule
\multirow{2}{*}{2025.04} & Qwen3 & \multirow{2}{*}{Alibaba Qwen} & \multirow{2}{*}{MoE/Dense} & \multirow{2}{*}{0.6-235B} & \multirow{2}{*}{GRPO} & \multirow{2}{*}{Text} & \multirow{2}{*}{\href{https://github.com/QwenLM/Qwen3}{\faGithub}\;
\href{https://huggingface.co/collections/Qwen/qwen3-67dd247413f0e2e4f653967f}{\faHuggingFace}} \\
& \citep{yang2025qwen3} & & & & & & \\
\midrule
\multirow{2}{*}{2025.05} & Llama-Nemotron & \multirow{2}{*}{NVIDIA} & \multirow{2}{*}{Dense} & \multirow{2}{*}{253B} & \multirow{2}{*}{GRPO} & \multirow{2}{*}{Text} & \multirow{2}{*}{\href{https://github.com/NVIDIA/Megatron-LM}{\faGithub}\;
\href{https://huggingface.co/nvidia/Llama-3_1-Nemotron-Ultra-253B-v1}{\faHuggingFace}} \\
& \citep{bercovich2025llama-nemotron} & & & & & & \\
\midrule
\multirow{2}{*}{2025.05} & INTELLECT-2 & \multirow{2}{*}{Intellect AI} & \multirow{2}{*}{Dense} & \multirow{2}{*}{32B} & \multirow{2}{*}{GRPO} & \multirow{2}{*}{Text} & \multirow{2}{*}{\href{https://huggingface.co/PrimeIntellect/INTELLECT-2}{\faHuggingFace}} \\
& \citep{team2025intellect} & & & & & & \\
\midrule
\multirow{2}{*}{2025.05} & Hunyuan-TurboS & \multirow{2}{*}{Tencent} & \multirow{2}{*}{Hybrid MoE} & \multirow{2}{*}{560B} & \multirow{2}{*}{GRPO} & \multirow{2}{*}{Text} & \multirow{2}{*}{\href{https://github.com/Tencent/Hunyuan-TurboS}{\faGithub}\;
\href{https://huggingface.co/spaces/tencent/hunyuan-turbos}{\faHuggingFace}} \\
& \citep{team2025hunyuan} & & & & & & \\
\midrule
\multirow{2}{*}{2025.05} & Skywork OR-1 & \multirow{2}{*}{Skywork} & \multirow{2}{*}{Dense} & \multirow{2}{*}{7B/32B} & \multirow{2}{*}{GRPO} & \multirow{2}{*}{Text} & \multirow{2}{*}{\href{https://github.com/SkyworkAI/Skywork-OR1}{\faGithub}\;
\href{https://huggingface.co/collections/Skywork/skywork-or1-67fa1bcb41b436ef2def76b9}{\faHuggingFace}} \\
& \citep{he2025skywork} & & & & & & \\
\midrule
\multirow{2}{*}{2025.05} & DeepSeek-R1-0528 & \multirow{2}{*}{DeepSeek} & \multirow{2}{*}{MoE/MLA} & \multirow{2}{*}{671B} & \multirow{2}{*}{GRPO} & \multirow{2}{*}{Text} & \multirow{2}{*}{\href{https://github.com/deepseek-ai/DeepSeek-R1}{\faGithub}\;
\href{https://huggingface.co/deepseek-ai/DeepSeek-R1-0528}{\faHuggingFace}} \\
& \citep{guo2025deepseek} & & & & & & \\
\midrule
\multirow{2}{*}{2025.06} & Magistral & \multirow{2}{*}{Mistral AI} & \multirow{2}{*}{Dense} & \multirow{2}{*}{24B} & \multirow{2}{*}{GRPO} & \multirow{2}{*}{Text} & \multirow{2}{*}{\href{https://huggingface.co/mistralai/Magistral-Small-2506}{\faHuggingFace}} \\
& \citep{mistralai2025magistral} & & & & & & \\
\midrule
\multirow{2}{*}{2025.06} & Minimax-M1 & \multirow{2}{*}{Minimax} & \multirow{2}{*}{Hybrid MoE} & \multirow{2}{*}{456B} & \multirow{2}{*}{CISPO} & \multirow{2}{*}{Text} & \multirow{2}{*}{\href{https://github.com/MiniMax-AI/MiniMax-M1}{\faGithub}\;
\href{https://huggingface.co/collections/MiniMaxAI/minimax-m1-68502ad9634ec0eeac8cf094}{\faHuggingFace}} \\
& \citep{chen2025minimax} & & & & & & \\
\midrule
\multirow{2}{*}{2025.07} & Intern-S1 & \multirow{2}{*}{Shanghai AI Lab} & \multirow{2}{*}{MoE} & \multirow{2}{*}{241B} & \multirow{2}{*}{GRPO} & \multirow{2}{*}{T/I/V} & \multirow{2}{*}{\href{https://github.com/InternLM/Intern-S1}{\faGithub}\;
\href{https://huggingface.co/collections/internlm/intern-s1-6882e325e8ac1c58ba108aa5}{\faHuggingFace}} \\
& \citep{bai2025intern} & & & & & & \\
\midrule
\multirow{2}{*}{2025.07} & Kimi K2 & \multirow{2}{*}{Kimi} & \multirow{2}{*}{MoE} & \multirow{2}{*}{1T} & \multirow{2}{*}{OPMD} & \multirow{2}{*}{Text} & \multirow{2}{*}{\href{https://github.com/MoonshotAI/Kimi-K2}{\faGithub}\;
\href{https://huggingface.co/collections/moonshotai/kimi-k2-6871243b990f2af5ba60617d}{\faHuggingFace}} \\
& \citep{kimiteam2025kimik2openagentic} & & & & & & \\
\midrule
\multirow{2}{*}{2025.07} & Step 3 & \multirow{2}{*}{Step AI} & \multirow{2}{*}{MoE} & \multirow{2}{*}{321B} & \multirow{2}{*}{-} & \multirow{2}{*}{T/I/V} & \multirow{2}{*}{\href{https://github.com/stepfun-ai/Step3}{\faGithub}\;
\href{https://huggingface.co/collections/stepfun-ai/step3-688a3d652dbb45d868f9d42d}{\faHuggingFace}} \\
& \citep{wang2025step} & & & & & & \\
\midrule
\multirow{2}{*}{2025.07} & Qwen3-2507 & \multirow{2}{*}{Alibaba Qwen} & \multirow{2}{*}{MoE/Dense} & \multirow{2}{*}{4-235B} & \multirow{2}{*}{GSPO} & \multirow{2}{*}{Text} & \multirow{2}{*}{\href{https://github.com/QwenLM/Qwen3}{\faGithub}\;
\href{https://huggingface.co/collections/Qwen/qwen3-67dd247413f0e2e4f653967f}{\faHuggingFace}} \\
& \citep{yang2025qwen3} & & & & & & \\
\midrule
\multirow{2}{*}{2025.07} & GLM-4.1V-Thinking & \multirow{2}{*}{Zhipu AI} & \multirow{2}{*}{Dense} & \multirow{2}{*}{9B} & \multirow{2}{*}{GRPO} & \multirow{2}{*}{T/I/V} & \multirow{2}{*}{\href{https://github.com/zai-org/GLM-V}{\faGithub}\;
\href{https://huggingface.co/collections/zai-org/glm-41v-thinking-6862bbfc44593a8601c2578d}{\faHuggingFace}} \\
& \citep{team_glm-45v_2025} & & & & & & \\
\midrule
\multirow{2}{*}{2025.07} & GLM-4.5 & \multirow{2}{*}{Zhipu AI} & \multirow{2}{*}{MoE} & \multirow{2}{*}{355B} & \multirow{2}{*}{GRPO} & \multirow{2}{*}{Text} & \multirow{2}{*}{\href{https://github.com/zai-org/GLM-4.5}{\faGithub}\;
\href{https://huggingface.co/collections/zai-org/glm-45-687c621d34bda8c9e4bf503b}{\faHuggingFace}} \\
& \citep{zeng2025glm} & & & & & & \\
\midrule
\multirow{2}{*}{2025.07} &  Skywork-R1V3 & \multirow{2}{*}{Skywork} & \multirow{2}{*}{Dense} & \multirow{2}{*}{38B} & \multirow{2}{*}{GRPO} & \multirow{2}{*}{T/I} & \multirow{2}{*}{\href{https://github.com/SkyworkAI/Skywork-R1V}{\faGithub}\;
\href{https://huggingface.co/Skywork/Skywork-R1V3-38B}{\faHuggingFace}} \\
& \citep{shen2025skywork} & & & & & & \\
\midrule
\multirow{2}{*}{2025.08} & gpt-oss & \multirow{2}{*}{OpenAI} & \multirow{2}{*}{MoE} & \multirow{2}{*}{117B/21B} & \multirow{2}{*}{-} & \multirow{2}{*}{Text} & \multirow{2}{*}{\href{https://github.com/openai/gpt-oss}{\faGithub}\;
\href{https://huggingface.co/collections/openai/gpt-oss-68911959590a1634ba11c7a4}{\faHuggingFace}} \\
& \citep{openai2025gpt-oss-120b} & & & & & & \\
\midrule
\multirow{2}{*}{2025.08} & Seed-OSS & \multirow{2}{*}{Bytedance Seed} & \multirow{2}{*}{Dense} & \multirow{2}{*}{36B} & \multirow{2}{*}{-} & \multirow{2}{*}{Text} & \multirow{2}{*}{\href{https://github.com/ByteDance-Seed/seed-oss}{\faGithub}\;
\href{https://huggingface.co/collections/ByteDance-Seed/seed-oss-68a609f4201e788db05b5dcd}{\faHuggingFace}} \\
& \citep{seed2025seed-oss} & & & & & & \\
\midrule
\multirow{2}{*}{2025.08} & GLM-4.5V & \multirow{2}{*}{Zhipu AI} & \multirow{2}{*}{MoE} & \multirow{2}{*}{106B} & \multirow{2}{*}{GRPO} & \multirow{2}{*}{T/I/V} & \multirow{2}{*}{\href{https://github.com/zai-org/GLM-V}{\faGithub}\;
\href{https://huggingface.co/collections/zai-org/glm-45v-68999032ddf8ecf7dcdbc102}{\faHuggingFace}} \\
& \citep{team_glm-45v_2025} & & & & & & \\
\midrule
\multirow{2}{*}{2025.08} & InternVL3.5 & \multirow{2}{*}{Shanghai AI Lab} & \multirow{2}{*}{MoE/Dense} & \multirow{2}{*}{1-241B} & \multirow{2}{*}{MPO/GSPO} & \multirow{2}{*}{T/I/V} & \multirow{2}{*}{\href{https://github.com/OpenGVLab/InternVL}{\faGithub}\;
\href{https://huggingface.co/spaces/OpenGVLab/InternVL}{\faHuggingFace}} \\
& \citep{wang_internvl35_2025} & & & & & & \\
\midrule
\multirow{2}{*}{2025.09} & ERNIE-4.5-Thinking & \multirow{2}{*}{Baidu} & \multirow{2}{*}{MoE} & \multirow{2}{*}{21B-A3B} & \multirow{2}{*}{-} & \multirow{2}{*}{Text} & \multirow{2}{*}{
\href{https://huggingface.co/baidu/ERNIE-4.5-21B-A3B-Thinking}{\faHuggingFace}} \\
& \citep{ernie2025technicalreport} & & & & & & \\
\midrule
\pagebreak
\multirow{2}{*}{2025.09} & Ring-mini-2.0 & \multirow{2}{*}{inclusionAI} & \multirow{2}{*}{MoE} & \multirow{2}{*}{16B} & \multirow{2}{*}{-} & \multirow{2}{*}{Text} & \multirow{2}{*}{
\href{https://huggingface.co/inclusionAI/Ring-mini-2.0}{\faHuggingFace}} \\
& \citep{Ringmini2.0} & & & & & & \\
\midrule
\multirow{2}{*}{2025.09} & Qwen3-Next-80B-A3B-Thinking & \multirow{2}{*}{Alibaba Qwen} & \multirow{2}{*}{MoE} & \multirow{2}{*}{80B} & \multirow{2}{*}{GSPO} & \multirow{2}{*}{Text} & \multirow{2}{*}{
\href{https://huggingface.co/Qwen/Qwen3-Next-80B-A3B-Thinking}{\faHuggingFace}} \\
& \citep{Ringmini2.0} & & & & & & \\
\midrule
\multirow{2}{*}{2025.09} & GLM-4.6 & \multirow{2}{*}{Zhipu AI} & \multirow{2}{*}{MoE} & \multirow{2}{*}{355B-A32B} & \multirow{2}{*}{-} & \multirow{2}{*}{Text} & \multirow{2}{*}{\href{https://github.com/zai-org/GLM-4.5}{\faGithub}\;
\href{https://huggingface.co/zai-org/GLM-4.6}{\faHuggingFace}} \\
& \citep{GLM4.6} & & & & & & \\
\midrule
\multirow{2}{*}{2025.09} & DeepSeek-V3.2-Exp & \multirow{2}{*}{DeepSeek} & \multirow{2}{*}{MoE/DSA} & \multirow{2}{*}{671B} & \multirow{2}{*}{GRPO} & \multirow{2}{*}{Text} & \multirow{2}{*}{\href{https://github.com/deepseek-ai/DeepSeek-V3.2-Exp}{\faGithub}\;
\href{https://huggingface.co/deepseek-ai/DeepSeek-V3.2-Exp}{\faHuggingFace}} \\
& \citep{deepseekai2024deepseekv32} & & & & & & \\
\midrule
\multirow{2}{*}{2025.09} & Ring-1T-preview & \multirow{2}{*}{inclusionAI} & \multirow{2}{*}{MoE} & \multirow{2}{*}{1T} & \multirow{2}{*}{-} & \multirow{2}{*}{Text} & \multirow{2}{*}{
\href{https://huggingface.co/inclusionAI/Ring-1T-preview}{\faHuggingFace}} \\
& \citep{Ring1T} & & & & & & \\
\midrule
\multirow{2}{*}{2025.09} & Qwen3-VL & \multirow{2}{*}{Alibaba Qwen} & \multirow{2}{*}{MoE/Dense} & \multirow{2}{*}{30B/235B} & \multirow{2}{*}{-} & \multirow{2}{*}{T/I/V} & \multirow{2}{*}{\href{https://github.com/QwenLM/Qwen3-VL}{\faGithub}\;
\href{https://huggingface.co/Qwen/Qwen3-VL-235B-A22B-Thinking}{\faHuggingFace}} \\
& \citep{Qwen3VL} & & & & & & \\
\end{longtable}

\endgroup

\subsection{Related Surveys}
\label{sec:pre_scope}

In this subsection, we compare recent surveys related to RL and LLMs.
Several surveys focus primarily on RL itself, covering both classical RL and its recent extensions.
\citet{ghasemi2024comprehensive} present a general RL survey covering algorithms and real-world challenges, \citet{huh2023multi} focuse on multi-agent RL, \citet{zhang2024survey} review self-play techniques, and \citet{wu2025reinforcement} survey RL in computer vision tasks.
While these works offer broad perspectives on RL, they do not explicitly address its application to LLMs.
In contrast, other surveys center on LLMs and their emerging capabilities, such as long chain-of-thought reasoning~\citep{xia2024beyond,chen2025towards,li2025system} and adaptive behaviors~\citep{sui2025stop,feng2025efficient}, where RL is often introduced as a key method to support these advances.
\citet{zhao2023survey} provide a broad overview of LLM architectures and applications, while more recent works concentrate specifically on reasoning abilities.
\citet{zhang2025100} survey replication studies on reasoning LLMs in the wake of DeepSeek-R1, \citet{chen2025towards} examine long chain-of-thought reasoning, and \citet{li2025system} analyze the transition from System~1 to System~2 reasoning.
These studies highlight RL-based methods such as RLHF and RLVR as useful tools, but treat them as only one element among a wide range of reasoning strategies.
\citet{sun2025survey} offer a broader, structured take on reasoning via foundation models. It highlights key foundation models that are either proposed or adapted specifically for reasoning, as well as recent progress across diverse reasoning tasks, methodologies, and benchmarks.
\citet{zhang2025landscapeagenticreinforcementlearning} examine how RL can endow LLMs with autonomous decision-making and adaptive agentic capabilities.
\citet{xu2025towards} move closer to our focus by discussing reinforced reasoning for LLMs, emphasizing how trial-and-error optimization can improve complex reasoning.
\citet{wu2025sailing} complement this view by surveying reward models and strategies for learning from feedback. Nevertheless, these works remain oriented towards reasoning performance or reward design, rather than offering a systematic treatment of RL methods as a whole for LLMs.
\citet{srivastava2025technical} represent a more recent attempt to bridge the two fields by reviewing RL algorithms for LLM alignment and enhancement, primarily through methods such as RLHF~\citep{christiano2017deep}, RLAIF~\citep{lee2023rlaif}, and DPO~\citep{rafailov2023direct}. It remains primarily focused on alignment rather than reasoning capabilities.

Unlike previous surveys that cover either general RL or reasoning in LLMs, we place RL at the center and provide a systematic synthesis of its role throughout the LLM training lifecycle, including reward design, policy optimization, and sampling strategies.
Our aim is to identify new directions for scaling reinforcement learning in LRMs toward ASI, focusing on long-term interactions and evolution.

\section{Foundational Components}
\label{sec:components}

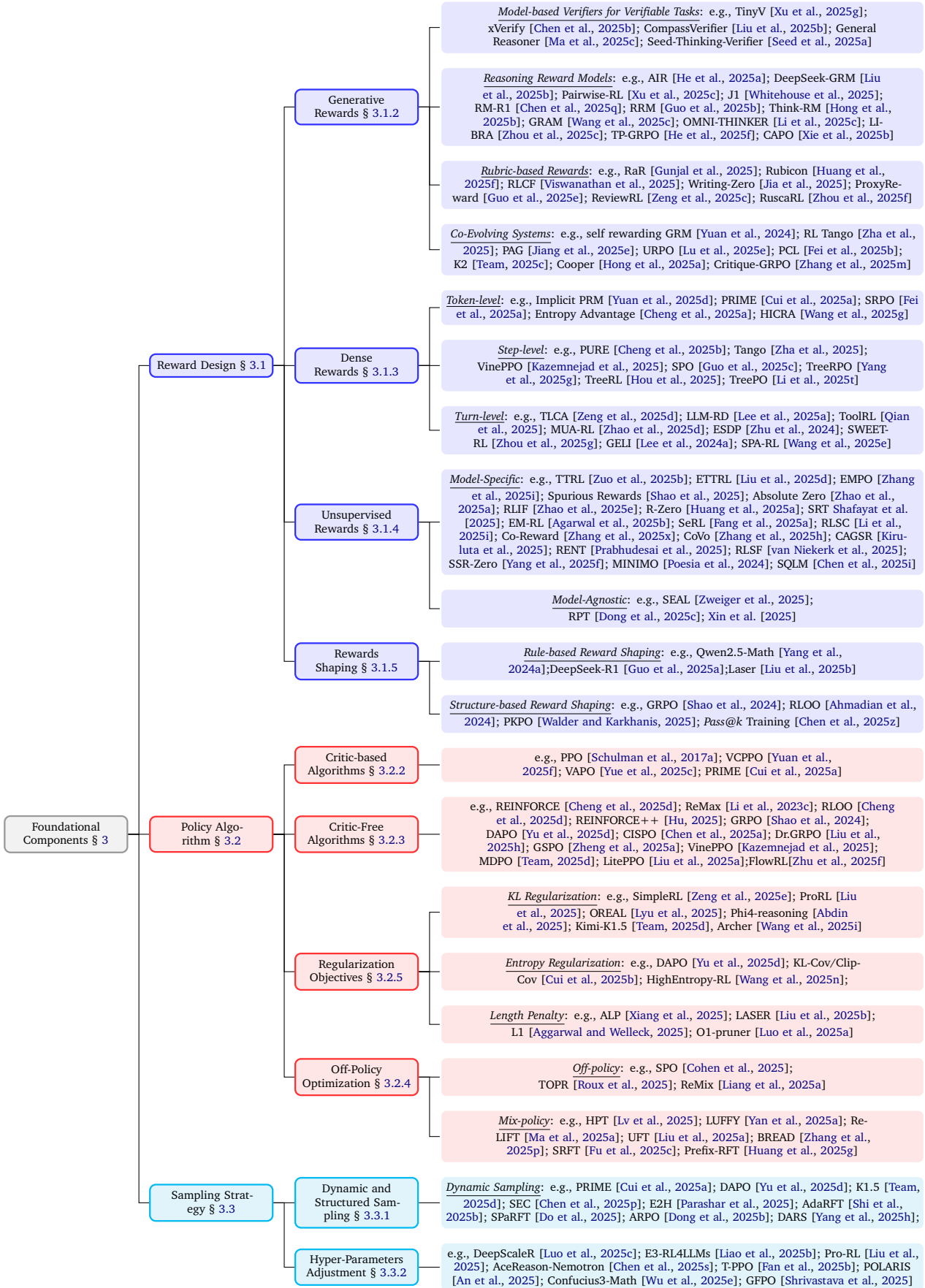
\begin{figure}
\footnotesize
\begin{forest}
    for tree={
        forked edges,
        grow'=0,
        draw,
        rounded corners,
        node options={align=center,},
        text width=2.7cm,
        s sep=6pt,
        calign=child edge, calign child=(n_children()+1)/2,
    },
    [Foundational Components~\S~\ref{sec:components}, fill=gray!45, parent
        [Reward Design~\S~\ref{sec:reward}, for tree={ pretrain}
            [Generative \\Rewards~\S~\ref{sec:reward_generative}, for tree={ pretrain}
                [
                    {\tstyle{Model-based Verifiers for Verifiable Tasks}: e.g., TinyV~\citep{xu2025tinyv}; xVerify~\citep{chen2025xverify}; CompassVerifier~\citep{liu2025compassverifier}; General Reasoner~\citep{ma2025general}; Seed-Thinking-Verifier~\citep{seed2025seed1}}, pretrain_work
                ]
                [
                    {\tstyle{Reasoning Reward Models}: e.g.,
                    AIR~\citep{he2025air}; DeepSeek-GRM~\citep{liu2025inference};  Pairwise-RL~\citep{xu2025unified}; J1~\citep{whitehouse2025j1}; RM-R1~\citep{chen2025rm}; RRM~\citep{guo2025reward}; Think-RM~\citep{hong2025think}; GRAM~\citep{wang2025gram}; 
                    OMNI-THINKER~\citep{li2025omni};
                    LIBRA~\citep{zhou2025libra}; TP-GRPO~\citep{he2025good}; CAPO~\citep{xie2025capo}}, pretrain_work
                ]
                [
                    {\tstyle{Rubric-based Rewards}: e.g., RaR~\citep{gunjal2025rubrics}; Rubicon~\citep{huang2025reinforcement}; RLCF~\citep{viswanathan2025checklists}; Writing-Zero~\citep{jia2025writing}; ProxyReward~\citep{guo2025general}; ReviewRL~\citep{zeng2025reviewrl}; RuscaRL~\citep{zhou2025breaking}}, pretrain_work
                ]
                [
                    {\tstyle{Co-Evolving Systems}: e.g., self rewarding GRM~\citep{yuan2024self}; RL Tango~\citep{zha2025rl}; PAG~\citep{jiang2025pag};  URPO~\citep{lu2025urpo}; PCL~\citep{fei2025post}; K2~\citep{kimiteam2025kimik2openagentic}; Cooper~\citep{hong2025cooper}; Critique-GRPO~\citep{zhang2025critique}}, pretrain_work
                ]
            ]
            [Dense \\Rewards~\S~\ref{sec:reward_dense}, for tree={ pretrain}
                [{\tstyle{Token-level}: e.g., Implicit PRM~\citep{yuan2024free}; PRIME~\citep{cui2025process}; SRPO~\citep{fei2025self}};
                Entropy Advantage~\citep{cheng2025reasoning};
                HICRA~\citep{wang2025emergent}
                , pretrain_work]
                [{\tstyle{Step-level}: e.g., 
                PURE~\citep{cheng2025stop}; Tango~\citep{zha2025rl}; VinePPO~\citep{kazemnejad2024vineppo}; SPO~\citep{guo2025segment}; TreeRPO~\citep{yang2025treerpo}; TreeRL~\citep{hou2025treerl}; TreePO~\citep{li2025treepo}}
                , pretrain_work]
                [{\tstyle{Turn-level}: e.g., TLCA~\citep{zeng2025reinforcing};
                LLM-RD~\citep{lee2025aligning};
                ToolRL~\citep{qian2025toolrl};
                MUA-RL~\citep{zhao2025mua};
                ESDP~\citep{zhu2024emotion};
                SWEET-RL~\citep{zhou2025sweet};
                GELI~\citep{lee2024improving};
                SPA-RL~\citep{wang2025spa}}
                , pretrain_work]
            ]
            [Unsupervised Rewards~\S~\ref{sec:reward_unsupervised}, for tree={ pretrain}
                [{\tstyle{Model-Specific}: e.g., TTRL~\citep{zuo2025ttrl}; ETTRL~\citep{liu2025ettrl}; EMPO~\citep{zhang2025right}; Spurious Rewards~\citep{shao2025spurious}; Absolute Zero~\citep{zhao2025absolute}; RLIF~\citep{zhao2025learning}; 
                R-Zero~\citep{huang2025r}; SRT~\cite{shafayat2025can}; EM-RL~\citep{agarwal2025unreasonable}; SeRL~\citep{fang2025serl}; RLSC~\citep{li2025confidence}; Co-Reward~\citep{zhang2025co}; CoVo~\citep{zhang2025consistent}; CAGSR~\citep{kiruluta2025self}; RENT~\citep{prabhudesai2025maximizing}; RLSF~\citep{van2025post}; SSR-Zero~\citep{yang2025ssr}; MINIMO~\citep{poesia2024learning}; SQLM~\citep{chen2025selfques}}
                , pretrain_work]
                [{\tstyle{Model-Agnostic}: e.g., SEAL~\citep{zweiger2025self}; RPT~\citep{dong2025reinforcement};~\citet{xin2025surrogate}}
                , pretrain_work]
            ]
            [Rewards \\Shaping~\S~\ref{sec:reward_shaping}, for tree={ pretrain}
                [{\tstyle{Rule-based Reward Shaping}: e.g., Qwen2.5-Math~\citep{yang2024qwen2};DeepSeek-R1~\citep{guo2025deepseek};Laser~\citep{liu2025learn}}
                , pretrain_work]
                [{\tstyle{Structure-based Reward Shaping}: e.g., GRPO~\citep{shao2024deepseekmath}; RLOO~\citep{ahmadian2024back}; PKPO~\citep{walder2025pass}; \emph{Pass@k} Training~\citep{chen2025pass}}
                , pretrain_work]
            ]
        ]
        [Policy Algorithm~\S~\ref{sec:policy}, for tree={fill=red!45,template}
            [Critic-based \\Algorithms~\S~\ref{sec:policy_critic_based},  template
                [{e.g., PPO~\citep{schulman2017equivalence}; 
                VCPPO~\citep{yuan2025s}; VAPO~\citep{yue2025vapo};
                PRIME~\citep{cui2025process}}
                , template_work]
            ]
            [Critic-Free \\Algorithms~\S~\ref{sec:policy_critic_free},  template
                [{e.g., REINFORCE~\citep{cheng2025revisiting}; ReMax~\citep{li2023remax}; RLOO~\citep{cheng2025revisiting}; REINFORCE++~\citep{hu2025reinforce++}; GRPO~\citep{shao2024deepseekmath}; DAPO~\citep{yu2025dapo}; CISPO~\citep{chen2025minimax}; Dr.GRPO~\citep{liu2025understanding}; GSPO~\citep{zheng2025group}; VinePPO~\citep{kazemnejad2024vineppo}; MDPO~\citep{team2025kimi}; LitePPO~\citep{liu2025part}};FlowRL\citep{zhu2025flowrl}
                , template_work]
            ]
            [Regularization Objectives~\S~\ref{sec:policy_regular},  template
                [{\tstyle{KL Regularization}: e.g., SimpleRL~\citep{zeng2025simplerl}; ProRL~\citep{liu2025prorl}; OREAL~\citep{lyu2025exploring}; Phi4-reasoning~\citep{abdin2025phi}; Kimi-K1.5~\citep{team2025kimi}, Archer~\citep{wang2025stabilizing}}
                , template_work]
                [{\tstyle{Entropy Regularization}: e.g., DAPO~\citep{yu2025dapo}; KL-Cov/Clip-Cov~\citep{cui2025entropy}; HighEntropy-RL~\citep{wang2025beyond}; 
                }
                , template_work]
                [{\tstyle{Length Penalty}: e.g., 
                ALP~\citep{xiang2025just};
                LASER~\citep{liu2025learn};
                L1~\citep{aggarwal2025l1};
                O1-pruner~\citep{luo2025o1}}
                , template_work]
            ]
            [Off-Policy \\Optimization~\S~\ref{sec:policy_off_policy},  template
                [{\tstyle{Off-policy}: e.g., SPO~\citep{cohen2025soft}; TOPR~\citep{roux2025tapered}; ReMix~\citep{liang2025squeeze}}
                , template_work]
                [{\tstyle{Mix-policy}: e.g., HPT~\citep{lv2025towards}; LUFFY~\citep{yan2025learning}; ReLIFT~\citep{ma2025learning}; UFT~\citep{liu2025uft}; BREAD~\citep{zhang2025bread}; SRFT~\citep{fu2025srft}; Prefix-RFT~\citep{huang2025blending}}
                , template_work]
            ]
        ]
        [Sampling Strategy~\S~\ref{sec:sampling}, for tree={fill=blue!45, answer}
            [Dynamic and Structured Sampling~\S~\ref{sec:sampling_dynamic}, answer
                [{\tstyle{Dynamic Sampling}: e.g., PRIME \citep{cui2025process}; DAPO \citep{yu2025dapo}; K1.5 \citep{team2025kimi}; SEC \citep{chen2025self}; E2H \citep{parashar2025curriculum}; 
                AdaRFT \citep{shi2025efficient}; SPaRFT \citep{do2025sparft}; ARPO \citep{dong2025agentic}; DARS \citep{yang2025depth};
                }
                , answer_work]
            ]
            [Hyper-Parameters Adjustment~\S~\ref{sec:sampling_hyper_params}, answer
                [{e.g., 
                DeepScaleR \citep{luo2025deepscaler}; E3-RL4LLMs \citep{liao2025enhancing}; Pro-RL \citep{liu2025prorl}; AceReason-Nemotron \citep{chen2025acereason}; 
                T-PPO \citep{fan2025truncated}; POLARIS \citep{an2025polaris}; Confucius3-Math \citep{wu2025confucius3}; GFPO \citep{shrivastava2025sample}}, answer_work];
            ]
        ]
    ]
\end{forest}
\caption{Taxonomy of foundational components and representative works for each direction.}
\label{tree:components}
\end{figure}

In this section, we review the foundational components of RL for LRMs, including reward design ($\S$~\ref{sec:reward}), policy optimization algorithms ($\S$~\ref{sec:policy}), and sampling strategies ($\S$~\ref{sec:sampling}). The taxonomy of the foundational components are shown in Figure~\ref{tree:components}.

\subsection{Reward Design}
\label{sec:reward}

In this subsection, we provide a comprehensive examination of reward design in RL for LRMs.
We begin in $\S$~\ref{sec:reward_verifiable} with verifiable rewards, which offer a natural starting point. There are substantial advances in this direction, exemplified by the success of DeepSeek-R1, which demonstrated the scalability of RL through verifiable reward mechanisms.
In contrast, $\S$~\ref{sec:reward_generative} examines generative rewards, wherein the model is engaged to either verify or directly generate reward signals.
However, both verifiable and generative rewards are typically expressed as sparse numerical feedback. An important complementary dimension lies in the density of the reward signal. $\S$~\ref{sec:reward_dense} accordingly examines approaches that incorporate dense rewards.
A further axis of categorization pertains to whether rewards are computed from external ground truth or instead estimated directly by the model. This distinction motivates our discussion of unsupervised rewards in $\S$~\ref{sec:reward_unsupervised}.
Building upon these four categories, we then turn in $\S$~\ref{sec:reward_shaping} to reward shaping, where we analyze strategies for combining or transforming diverse reward signals to facilitate learning.

\subsubsection{Verifiable Rewards}
\label{sec:reward_verifiable}
\begin{myboxi}[Takeaways]
\begin{itemize}
    \item Rule-based rewards provide scalable and reliable training signals for RL, especially in math and code tasks, by leveraging accuracy and format checks.
    \item Verifier's law highlights that tasks with clear and automatic verification enable efficient RL optimization, while subjective tasks remain challenging.
\end{itemize}
\end{myboxi}

\paragraph{Rule-based Rewards.}
The reward serves as the training signal of RL, determining the optimization direction~\citep{guo2025deepseek}.
Recently, rule-based verifiable rewards have been predominantly employed to train LRMs in large-scale RL.
Such rewards enable the reliable enhancement of mathematical and coding reasoning abilities by encouraging longer and more reflective chain-of-thought~\citep{guo2025deepseek,yu2025dapo,kimiteam2025kimik2openagentic}.
This paradigm was formalized as RLVR in the Tülu 3~\citep{lambert2024tulu}, which replaces a learned reward model with a programmatic verifier (e.g., answer checkers or unit tests).
Such verifiers provide binary, checkable signals in domains with objectively verifiable outcomes.
Similar rule-based approaches to verifiable reward design were subsequently integrated into DeepSeek’s training pipeline.
For instance, DeepSeek-V3~\citep{liu2024deepseek} explicitly incorporated a rule-based reward system tailored to deterministic tasks, while DeepSeek-R1~\citep{guo2025deepseek} further employed accuracy-based and format-based rewards.
Rule-based rewards stand in contrast to outcome-based or process-based Reward Models (RMs), such as standard RLHF with a learned reward model trained on human preference rankings~\citep{ouyang2022training} and Process Reward Models (PRMs) trained on step-level annotations~\citep{yuan2024free,setlur2024rewarding,sun2025freeprm}.
DeepSeek-V3 and DeepSeek R1 demonstrate that RMs may suffer from reward hacking when scaled to large-scale RL settings, but by leveraging rule-based rewards wherever possible, we ensure greater reliability by making the system resistant to manipulation and exploitation~\citep{liu2024deepseek,guo2025deepseek}.
In practice, two kinds of rule-based verifiable rewards are widely used:
\begin{itemize}
    \item \textbf{Accuracy rewards:} For tasks with deterministic outcomes (e.g., math), the policy must produce the final solution within a prescribed delimiter (commonly \verb|\boxed{...}|). An automatic checker then compares this output to the ground truth. For coding tasks, unit tests, or compilers provide the pass/fail signal~\citep{guo2025deepseek,chen2025r1,albalak2025big}.
    \item \textbf{Format rewards:} These impose a structural constraint requiring the model to place its private chain-of-thought between \texttt{<think>} and \texttt{</think>}, and to output the final answer in a separate field (e.g., \texttt{<answer>\dots</answer>}). This improves reliable parsing and verification in large-scale RL~\citep{guo2025deepseek,lambert2024tulu}.
\end{itemize}

\paragraph{Rule-based Verifier.}
Rule-based rewards are typically derived from rule-based verifiers. These rely on a large collection of manually written equivalence rules to determine whether a predicted answer matches the ground truth.
Currently, widely used mathematical verifiers are primarily built on the Python libraries Math-Verify\footnote{\url{https://github.com/huggingface/Math-Verify}} and SymPy\footnote{\url{https://www.sympy.org/}}.
In addition, some works such as DAPO~\citep{yu2025dapo} and DeepScaleR~\citep{luo2025deepscaler}, also provide open-source and well-established verifiers.
Recently, \citet{huang2025pitfalls} highlight the distinctive limitations associated with both rule-based and model-based verifiers, to inform the design of more reliable reward systems.

In practice, tasks such as mathematical problem solving and code generation are difficult to solve yet comparatively easy to verify, thereby satisfying the main criteria for efficient RL optimization~\citep{guo2025deepseek,he2025skywork}: the existence of clear ground truth, the availability of rapid automated verification, the scalability of evaluating many candidate solutions, and a reward signal that is closely aligned with correctness.
By contrast, tasks lacking fast or objective verification (e.g., open-ended question answering or free-form writing) remain challenging for outcome-based RL, as they rely on noisy learned reward models or subjective human feedback~\citep{zhou2025reinforcing,yu2025rlpr}. Verifier’s Law posits that the ease of training AI systems to perform a task is proportional to the degree to which the task is verifiable\footnote{\url{https://www.jasonwei.net/blog/asymmetry-of-verification-and-verifiers-law}}. It emphasizes that once a task can be equipped with robust automated feedback, it becomes amenable to rapid improvement via RL.
The successful applications discussed in $\S$\ref{sec:application} substantiate this principle, as their central challenge lies in the design of reliable verifiable feedback. Conversely, many of the open problems highlighted in $\S$\ref{sec:future} arise precisely from the absence of dependable automated rewards.

\subsubsection{Generative Rewards}
\label{sec:reward_generative}
\begin{myboxi}[Takeaways]
\begin{itemize}
    \item Generative Reward Models (GenRMs) extend RL to subjective, non-verifiable domains by providing nuanced, text-based feedback, overcoming the limitations of rule-based systems.
    \item A dominant trend is training RMs to reason before judging, often using structured rubrics to guide evaluation or co-evolving them with the policy model in a unified RL loop.
\end{itemize}
\end{myboxi}

While rule-based rewards provide reliable signals for verifiable tasks, as discussed previously ($\S$~\ref{sec:reward_verifiable}), their applicability is limited. Many complex reasoning tasks, particularly in open-ended or creative domains, lack objective ground truth, making them intractable for simple verifiers. To bridge this gap, GenRMs have emerged as a powerful alternative. Instead of outputting a simple scalar score, GenRMs leverages the generative capabilities of LRMs to produce structured critiques, rationales, and preferences, providing a more interpretable and nuanced reward signal~\citep{zhang2024generative, mahan2024generative}. This approach addresses two key challenges: first, it improves the robustness of verification for verifiable tasks that are difficult to parse; second, and more importantly, it enables the application of RL to subjective, non-verifiable domains.

\paragraph{Model-based Verifiers for Verifiable Tasks.}
A primary challenge with rule-based systems is their brittleness; they often produce false negatives when a model generates a correct answer in an unexpected format. To mitigate this, one line of research uses \textit{Specification-Based GenRMs} as flexible, model-based verifiers. These models are trained to semantically assess the equivalence between a model's free-form output and a reference answer. This approach has been used to develop lightweight verifiers that augment existing rule-based systems~\citep{xu2025tinyv}, as well as more comprehensive, multi-domain verifiers capable of handling diverse data types and reasoning tasks~\citep{chen2025xverify, liu2025compassverifier, ma2025general, seed2025seed1}. By replacing or supplementing rigid string matching with learned semantic judgment, these verifiers provide more accurate reward signals for RL in verifiable domains.

\paragraph{Generative Rewards for Non-Verifiable Tasks.}
Another core application of GenRMs is \textit{Assessment-Based GenRMs}, which enable RL for tasks where Verifier's Law does not hold. This paradigm has evolved from using powerful LLMs as zero-shot evaluators to sophisticated, co-evolving systems. We can categorize these approaches based on their core design principles.
\begin{itemize}
    \item \textbf{Reasoning Reward Models (Learning to Think)}: A major advancement beyond simple preference prediction is to train RMs to explicitly reason before rendering a judgment. This approach, foundational to the LLM-as-a-Judge concept~\citep{zheng2023judging, li2023generative}, involves prompting the RM to generate a CoT critique or rationale. For instance, CLoud RMs first generate a natural language critique and then use it to predict a scalar reward~\citep{ankner2024critique}. This principle of formulating reward modeling as a reasoning task is now central to state-of-the-art RMs, which are trained to produce detailed rationales before assigning a score or preference~\citep{chen2025rm, guo2025reward, liu2025inference, hong2025think, wang2025gram, zhou2025libra}. To further improve their judgment capabilities, these reasoning RMs are often trained with RL themselves, using simple, verifiable meta-rewards based on the correctness of their final verdict~\citep{chen2025judgelrm, whitehouse2025j1}. This line of work also explores different reward formats, such as deriving soft rewards from token probabilities~\citep{zhang2024generative, mahan2024generative, su2025crossing} and weighing the trade-offs between pointwise and pairwise scoring schemes~\citep{he2025air, xu2025unified}.

    \item \textbf{Rubric-based Rewards (Structuring Subjectivity)}: To anchor the evaluation of subjective tasks in more consistent criteria, many frameworks employ structured rubrics. Unlike rule-based approaches that rely on hard-coded logic for objective, verifiable tasks, rubric-based methods leverage natural language descriptions to capture nuanced evaluation criteria for subjective, non-verifiable domains where traditional binary rules would be insufficient. This involves using an LLM to either generate or follow a checklist of principles to guide its assessment. Frameworks like RaR~\citep{gunjal2025rubrics}, QA-LIGN~\citep{dineen2025qa}, Rubicon~\citep{huang2025reinforcement}, and RLCF~\citep{viswanathan2025checklists} use such rubrics to produce fine-grained, multi-faceted rewards. This concept extends to decomposing high-level tasks into a set of verifiable proxy questions~\citep{guo2025general} or generating domain-specific principles, such as for creative writing~\citep{jia2025writing} or scientific reviews~\citep{zeng2025reviewrl}. Furthermore, rubrics can serve a dual purpose as both instructional scaffolds to guide policy exploration and as criteria for the final reward~\citep{zhou2025breaking}.
    \cite{bhaskar2025language} introduce RLMT, a RL paradigm that uses model-rewarded thinking to improve reasoning and chat capabilities in language models, outperforming standard RLHF pipelines and achieving benchmarks in chat and creative writing tasks.

    \item \textbf{Co-Evolving Systems (Unifying Policy and Reward):} The most advanced paradigm moves beyond a static policy-reward relationship and toward dynamic systems where the generator and verifier improve together. This can occur through:
    \begin{itemize}
        \item \textbf{Self-Rewarding}, where a single model generates its own training signals. This was notably demonstrated in Self-Rewarding Language Models~\citep{yuan2024self} and has been operationalized in frameworks where a model alternates between policy and verifier roles~\citep{jiang2025pag}, performs self-correction based on its own critique~\citep{xiong2025self, zhang2025critique, kimiteam2025kimik2openagentic}, or internalizes the reward function via post-completion learning~\citep{fei2025post}.
        \item \textbf{Co-Optimization}, where the policy and a separate reward model are trained concurrently. For example, RL Tango jointly trains the generator and a process-level GenRM using a shared outcome-level reward~\citep{zha2025rl}. Similarly, Cooper co-optimizes both models to enhance robustness and mitigate reward hacking~\citep{hong2025cooper}. Other works unify the policy (``player'') and reward (``referee'') functions within a single model trained via a unified RL loop~\citep{lu2025urpo}.
    \end{itemize}
\end{itemize}

This evolution from static judges to dynamic, co-evolving systems is often supported by hybrid reward schemes that combine rule-based and generative signals~\citep{li2025omni, seed2025seed1}. Additionally, GenRMs are being adapted to provide more granular, process-level feedback to address the credit assignment problem in complex reasoning chains~\citep{zhao2025genprm, he2025good, xie2025capo, khalifa2025process}. In essence, generative rewards are proving indispensable for scaling RL to the full spectrum of tasks targeted by general-purpose LRMs.

\subsubsection{Dense Rewards}
\label{sec:reward_dense}

\begin{myboxi}[Takeaways]
\begin{itemize}
    \item Dense rewards (e.g., process reward models) provide fine-grained credit assignment and improve training efficiency and optimization stability in RL.
    \item Scaling remains challenging for tasks like open-domain text generation due to the difficulty of defining dense rewards or using verifiers.
\end{itemize}
\end{myboxi}

In classical RL such as gaming and robotic manipulation tasks~\citep{schrittwieser2020mastering, liu2022meta, sun2025large}, dense rewards provide frequent feedback at (nearly) every decision step. Such shaping shortens the credit assignment horizon and often improves sample efficiency and optimization stability, but it also risks mis-specification and reward hacking if the signal is poorly designed~\citep{hadfield2017inverse}.
As for LLM reasoning, dense rewards are typically process-based signals that supervise intermediate steps rather than only outcomes, and they have been found effective, often outperforming outcome-based rewards~\citep{uesato2022solving, lightman2023let}.
Based on the definitions in \S~\ref{sec:pre_background}, we further formalize sparse/outcome and dense rewards in the context of LLM RL, according to the action and reward granularity, as shown in Table~\ref{tab:comparison_action_reward_definition}.

\begin{table}[!t]
\centering
\caption{Definitions of action and reward granularity in RL for language models ($z^{(u)}$ is the environment feedback at turn $u$).}
\label{tab:comparison_action_reward_definition}
\begingroup
\renewcommand{\arraystretch}{1.2}
\begin{tabular}{@{}p{2.4cm} p{5.5cm} p{3.8cm} p{2.4cm}@{}}
\toprule
\textbf{Granularity} & \textbf{Action} & \textbf{Reward} & \textbf{Return ($G$)} \\
\midrule
Trajectory & Entire sequence $y=(a_1,\dots,a_T)$ & Scalar $R(x, y)$ & $R(x, y)$ \\
Token & Each token $a_t \in \mathcal{V}$ & $r_t = R(x, a_{1:t})$ & $\sum_{t=1}^T \gamma^{t-1} r_t$ \\
Step & Segment $y^{(k)}$ (e.g., sentence) & $r_k = R(x, y^{(1:k)})$ & $\sum_{k=1}^K \gamma^{k-1} r_k$ \\
Turn (Agent) & Agent response $y^{(u)}$ per turn & $r_u = R(x, y^{(1:u)}, z^{(1:u)})$ & $\sum_{u=1}^U \gamma^{u-1} r_u$ \\
\bottomrule
\end{tabular}
\endgroup
\end{table}

\paragraph{Token-Level Rewards.}
DPO~\citep{rafailov2023direct} and its subsequent work~\citep{rafailov2024r} show that token-level rewards can be computed as log-likelihood ratios between the policy and the reference model. Implicit PRM~\citep{yuan2024free} further shows that token-level rewards can be obtained by training an ORM and using the parameterization of~\citet{rafailov2024r}. PRIME~\citep{cui2025process} integrates ORM learning into RL training and uses implicit token-level rewards to train the policy. SRPO~\citep{fei2025self} removes the ORM in PRIME and improves advantage estimation.
Another line of works focus on using internal feedback as token-level rewards, such as token entropy~\citep{cheng2025reasoning, tan2025gtpo} and strategic grams~\citep{wang2025emergent}.

\paragraph{Step-Level Rewards.}
Approaches to step-level rewards fall into two classes: \textit{model-based} and \textit{sampling-based}. Early works rely on human experts to annotate step-level dense rewards~\citep{uesato2022solving, lightman2023let}, which is costly and difficult to scale. 
\begin{itemize}
    \item \textbf{Model-based}: To reduce annotation cost, Math-Shepherd~\citep{wang2023math} uses Monte Carlo estimation to obtain step-level labels and demonstrates that process verification with trained PRMs is effective in RL. PAV~\citep{setlur2024rewarding} further improves process rewards via advantage modeling. To mitigate reward hacking with model-based step-level rewards, PURE~\citep{cheng2025stop} adopts min-form credit assignment rather than sum-form, while Tango~\citep{zha2025rl} and AIRL-S~\citep{jin2025your} jointly train the policy and PRMs. With the strong verification capabilities of generative PRMs~\citep{zhao2025genprm} (discussed in \S~\ref{sec:reward_generative}), ReasonFlux-PRM~\citep{zou2025reasonflux}, TP-GRPO~\citep{he2025good}, and CAPO~\citep{xie2025capo} leverage them to provide step-level rewards for RL training.
    SGPO~\citep{chen2025stepwise} leverages a strong judge model to identify the first incorrect step and computes advantage values based on the index of that step.
    Nevertheless, model-based dense rewards are vulnerable to reward hacking, and training PRMs online is expensive.
    \item \textbf{Sampling-based}: Another line of works use Monte Carlo sampling for online process reward estimation~\citep{kazemnejad2024vineppo, guo2025segment, yang2025treerpo, hou2025treerl, zheng2025first, li2025treepo}. VinePPO~\citep{kazemnejad2024vineppo} improves PPO via Monte Carlo estimation. To improve step segmentation, SPO~\citep{guo2025segment}, TreeRL~\citep{hou2025treerl}, and FR3E~\citep{zheng2025first} use low-probability or high-entropy tokens as division points, while AttnRL~\citep{liu2025attention} further proposes to branch at steps with high attention scores. To improve sample efficiency and advantage estimation, SPO~\citep{guo2025segment}, TreeRPO~\citep{yang2025treerpo}, TreeRL~\citep{hou2025treerl} and TreePO~\citep{li2025treepo} explore tree-based structures for fine-grained process reward computation.
    MRT~\citep{qu2025optimizing}, S-GRPO~\citep{dai2025s}, VSRM~\citep{yue2025promoting}, and SSPO~\citep{xu2025sspo} force the LLM to terminate the thinking process at intermediate positions to estimate step-level rewards efficiently.
    PROF~\citep{ye2025beyond} utilizes the consistency between outcome rewards and process rewards to filter noisy data for RL training.
    \cite{feng2025group} propose Group-in-Group Policy Optimization (GiGPO), a novel two-level, critic-free RL method that enables fine-grained step-level credit assignment in multi-turn LLM agent training.
\end{itemize}

\paragraph{Turn-Level Rewards.} 
Turn-level rewards evaluate each complete agent-environment interaction, such as a tool call and its result, providing feedback at the granularity of a single turn in multi-turn tasks. Research on turn-level rewards can be broadly divided into two lines: direct per-turn supervision and deriving turn-level signals from outcome-level rewards. 
\begin{itemize}
    \item For direct per-turn supervision, works provide explicit feedback at each turn. For example, Emotion-sensitive dialogue policy learning~\citep{zhu2024emotion} exploits user emotions as per-turn rewards to guide policy optimization, showing how turn-level feedback can enhance interaction quality in conversational agents. Similarly, ToolRL~\citep{qian2025toolrl} designs structured rewards on format and correctness that are provided at each tool invocation step, offering dense turn-level signals for learning. \citet{zeng2025reinforcing} further leverage verifiable signals with explicit turn-level advantage estimation to improve multi-turn tool use during RL. In addition, SWEET-RL~\citep{zhou2025sweet} learns a step/turn-level critic that provides per-turn rewards and credit assignment, thereby supplying explicit turn-level supervision. More recently, MUA-RL~\citep{zhao2025mua} incorporates simulated user interactions into the RL loop, where each multi-turn exchange produces per-turn feedback, allowing the agent to iteratively refine its policy under realistic user-agent dynamics.
    G-RA~\citep{sun2025stabilizing} extends this line of work by introducing gated reward aggregation, where dense turn-level rewards (e.g., action format, tool call validity, tool choice) are only accumulated if higher-priority outcome-level conditions are satisfied.
    \item For deriving turn-level signals from outcome-level rewards, the idea is to decompose or redistribute outcome-based supervision into finer-grained units. Aligning Dialogue Agents with Global Feedback~\citep{lee2025aligning} transforms session-level scores into turn-level pseudo-rewards, and GELI~\citep{lee2024improving} exploits multimodal cues such as prosody and facial expressions to refine session-level feedback into local turn-level signals. Similarly, SPA-RL~\citep{wang2025spa} redistributes outcome-based rewards into per-step or per-turn contributions through progress attribution. ARPO~\citep{dong2025agentic} follows this line by attributing step/turn-level advantages from trajectory-level outcomes (e.g., after tool use), effectively converting global returns into localized signals.
\end{itemize}

Overall, turn-level rewards, whether directly assigned at each interaction or derived from outcome decomposition, serve as a bridge between process- and outcome-based supervision, and play a central role in stabilizing and improving optimization in multi-turn agent RL, with more details in $\S$~\ref{sec:application_agentic}.

\subsubsection{Unsupervised Rewards}
\label{sec:reward_unsupervised}

\begin{myboxi}[Takeaways]
\begin{itemize}
    \item Unsupervised rewards eliminate the human annotation bottleneck, enabling reward signal generation at the scale of computation and data, not human labor.
    \item Main approaches include deriving signals either from the model's own processes (Model-Specific: consistency, internal confidence, self-generated knowledge) or from automated external sources (model-agnostic: heuristics, data corpora).
\end{itemize}
\end{myboxi}

Frontier language models excel at a wide range of tasks, including many that are exceptionally challenging \citep{li2024numinamath,phan2025humanitysexam,glazer2024frontiermath,jimenez2023swe}.
However, a key limitation in advancing these models is the reliance on human-generated reward signals for RL ($\S$~\ref{sec:reward_verifiable}--\ref{sec:reward_dense}).
For tasks requiring superhuman expertise, human feedback is often slow, expensive, and impractical \citep{burns2023weak}.
To address this, a promising approach is Unsupervised RL, which uses automatically generated, verifiable reward signals instead of ground-truth labels.
This method is fundamental to achieving scalable RL for LLMs.
This section surveys these unsupervised reward mechanisms, categorizing them into two types based on their source: those derived from the model itself (\textit{Model-Specific}) and those from external, non-human sources (\textit{Model-Agnostic}).

\paragraph{Model-Specific Rewards.} This paradigm uses an LLM's internal knowledge as the sole source of supervision. It operates on the assumption that a high-performing model will generate consistent, confident, or evaluatively sound outputs. This method is highly scalable, requiring only the model and computational resources to generate a virtually infinite amount of ``labeled'' data. However, its closed-loop nature risks reward hacking and model collapse.

\begin{itemize}
    \item \textbf{Rewards from Output Consistency}: This approach posits that correct answers will form a dense, consistent cluster among multiple generated outputs. Foundational works like EMPO~\citep{zhang2025right} and Test-Time Reinforcement Learning (TTRL)~\citep{zuo2025ttrl} operationalize this via clustering and majority voting, respectively. Subsequent methods aim to refine this by improving efficiency (ETTRL \citep{liu2025ettrl}), incorporating reasoning trajectories (CoVo \citep{zhang2025consistent}), or using contrastive agreement to combat reward hacking (Co-Reward \citep{zhang2025co}).
    \citet{zhou2025evolving} introduce EVOL-RL, a label-free RL approach that stabilizes training while promoting diversity and generalization through a combination of majority vote selection and novelty-aware rewards.

    \item \textbf{Rewards from Internal Confidence}: An alternative is to derive rewards directly from the model's internal states, using confidence as a proxy for correctness. Signals can be based on cross-attention (CAGSR \citep{kiruluta2025self}), negative entropy (EM-RL \citep{agarwal2025unreasonable}, RENT \citep{prabhudesai2025maximizing}), or generation probabilities (Intuitor \citep{zhao2025learning}, RLSC \citep{li2025confidence}, RLSF \citep{van2025post}). The success of these methods often depends on the base model's initial quality \citep{gandhi2025cognitive} and can be brittle \citep{shumailov2023curse,press2024entropy}, as they rely on priors like low-density separation between correct and incorrect paths \citep{chapelle2005semi,lee2013pseudo}.

    \item \textbf{Rewards from Self-Generated Knowledge}: This paradigm uses the model's knowledge to create learning signals, either by acting as a judge (self-rewarding) or a problem proposer (self-instruction). In self-rewarding, the model evaluates its own outputs to generate a reward, a concept framed by \citet{yuan2024self} and \citet{wu2024meta} and applied in works like SSR-Zero \citep{yang2025ssr} and MINIMO \citep{poesia2024learning}. In self-instruction, a proposer model generates a curriculum for a solver. The proposer is often rewarded for creating tasks of optimal difficulty \citep{zhao2025absolute,huang2025r,chen2025selfques}, while the solver's reward can be model-agnostic (e.g., from a code executor in AZR \citep{zhao2025absolute}) or Model-Specific (e.g., via majority voting in SQLM \citep{chen2025selfques} and SeRL \citep{fang2025serl}).

\end{itemize}

\paragraph{Model-Agnostic Rewards.} In contrast to Model-Specific methods, this paradigm derives rewards from external, automated sources. This approach grounds the learning process in external information, eliminating the need for human labels. Its core principle is that these external signals are readily accessible and do not require manual effort. However, since precise feedback is often unavailable, the quality of the proxy reward is critical, and the risk of reward hacking persists.

\begin{itemize}
    \item \textbf{Heuristic Rewards}: This approach constitutes another form of rule-based reward, employing simple, predefined rules based on output properties such as length or format as proxies for quality. It represents a specific case discussed in $\S$~\ref{sec:reward_verifiable}. This was pioneered by DeepSeek-R1 \citep{guo2025deepseek} and later refined with techniques like dynamic reward scaling \citep{yu2025dapo}. While scalable, these heuristics can be gamed by the model, leading to superficial improvements without advancing true capability \citep{xin2025surrogate,liu2025there}.

    \item \textbf{Data-Centric Rewards}: This approach derives reward signals from the structure of large, unlabeled corpora. Analogous to next-word prediction for large-scale pre-training, RPT \citep{dong2025reinforcement,wang2025thinking,li2025reinforcement} reframes next-token prediction as an RL task, turning web-scale datasets into millions of training examples. At a meta-level, SEAL \citep{zweiger2025self} allows a model to generate its own training data and hyperparameters, using downstream performance as the reward.

\end{itemize}

In summary, unsupervised reward design is essential for creating scalable RL systems for LLMs. The Model-Specific paradigm facilitates self-improvement by leveraging the model's internal knowledge, whereas the Model-Agnostic paradigm grounds learning in external, automated feedback. While both approaches effectively bypass the human annotation bottleneck, they remain susceptible to reward hacking \citep{zhang2025no}. The future of scalable RL will likely involve hybrid systems that strategically combine these methods, for instance, using data-centric rewards for pre-training, Model-Specific self-rewarding for fine-tuning on complex reasoning, and minimal human oversight for safety and alignment.

\subsubsection{Rewards Shaping}
\label{sec:reward_shaping}

\begin{myboxi}[Takeaways]
\begin{itemize}
    \item Reward shaping enriches sparse signals into stable, informative gradients for LLM training.
    \item Combine verifiers with reward models, and use group baselines plus \passk-aligned objectives to stabilize training, expand exploration, and match evaluation metrics at scale.
\end{itemize}
\end{myboxi}

As noted, the primary learning objective of agents in RL is to maximize cumulative rewards, making the design of the reward function particularly critical~\citep{sutton1998introduction}. In previous sections, we introduced various reward functions, such as verifiable rewards ($\S$~\ref{sec:reward_verifiable}), generative rewards ($\S$~\ref{sec:reward_generative}), dense rewards ($\S$~\ref{sec:reward_dense}) and even unsupervised rewards ($\S$~\ref{sec:reward_unsupervised}). Beyond reward engineering, it is equally important to consider how the reward function can be modified or augmented to encourage behaviors that drive progress toward the desired solution. This process, known as reward shaping~\citep{goyal2019using,hu2020learning,gupta2022unpacking,xie2023text2reward}, can be categorized into rule-based and structured-based reward shaping.

\paragraph{Rule-based Reward Shaping.}
The simplest and most commonly adopted approach to reward shaping in LLM-based RL involves combining rewards from both a rule-based verifier and a reward model to generate the overall reward signal, as demonstrated in Qwen2.5 Math~\citep{yang2024qwen2}. Typically, a constant coefficient is used to balance the contributions of the reward model and the rule-based component. Rather than assigning identical rewards to all correct responses, this method allows for further ranking of responses based on the scores from the reward model. This approach is particularly useful for more challenging samples and helps to avoid cases where all reward values are 0 or 1, which would otherwise lead to ineffective learning gradients~\citep{yu2025dapo}. This heuristic combination strategy is widely employed in open-domain tasks, where integrating rule-based rewards and reward models~\citep{liu2025inference,guo2025reward,liao2025rlmr} results in more informative and effective reward signals for the RL of LLM~\citep{zeng2025reviewrl,zhang2024generative,su2025crossing}. Another approach involves combining rule-based rewards, such as outcome-level rewards and format rewards, as implemented in DeepSeek-R1~\citep{guo2025deepseek}, which enables LLMs to learn long chain-of-thought reasoning. These rewards include format-based~\citep{xin2025surrogate} and length-based components~\citep{liu2025learn} to address various exceptions in the outputs of LLMs.
In contrast to using fixed reward weights~\citep{yao2025training,team2025kimi} or heuristic rules for reward interpolation~\citep{zhang2025grpo,aggarwal2025l1}, \cite{lu2025learning} propose dynamic reward weighting, employing both hypervolume-guided weight adaptation and gradient-based weight optimization. This approach achieves superior performance on multi-objective alignment tasks~\citep{liu2024stochastic,li2025gradient}.
Recent work also explores multi-role RL training and assigns different rewards for different roles with different reward functions, such as solver and critic~\citep{li2025jointlyreinforcingdiversityquality}.
Typically, these rewards are combined using manually set constants.
Recent works have also explored multi-role RL training~\citep{li2025chain,li2025jointlyreinforcingdiversityquality}, assigning distinct reward functions to different roles to encourage diverse behaviors and objectives~\citep{li2025jointlyreinforcingdiversityquality}, such as solver and critic.

\paragraph{Structure-based Reward Shaping.}
In contrast to rule-based reward shaping, which relies solely on individual samples, structure-based reward shaping computes rewards across a group of candidates by leveraging list-wise or set-level baselines. One influential method is GRPO~\citep{shao2024deepseekmath}, which uses the group mean of responses to the same question G as a baseline (or variants such as leave-one-out~\citep{ahmadian2024back} or ranking) and constructs advantages accordingly for PPO-style updates~\citep{schulman2017proximal}.
Recent works have further modified the optimization objective or credit allocation strategies to promote stronger exploration and achieve closer alignment with evaluation metrics, such as \passk~\citep{yue2025does}. For example, \citet{walder2025pass} perform a joint transformation on the final reward, making the optimization directly equivalent to set-level objectives like \passk, and provide low-variance, unbiased gradient estimation. \citet{chen2025pass} directly target \passk~in deriving and analyzing advantages and efficient approximations, decomposing set-level targets back into individual sample credit allocation.
Reward shaping methods in this direction aim to stabilize training and encourage the policy to explore more extensively, thereby reducing the risk of premature convergence to suboptimal local solutions.

\subsection{Policy Optimization}
\label{sec:policy}

In this subsection, we first provide a technical overview of the mathematical formulation of the policy gradient objective ($\S$~\ref{sec:policy_gradient}). Next, we divide the on-policy optimization algorithms in RL into two categories based on how the reward is generated for the gradient calculation process: critic-based ($\S$~\ref{sec:policy_critic_based}) and critic-free ($\S$~\ref{sec:policy_critic_free}). In addition, we discuss recent studies that combine on-policy RL with offline datasets for more sophisticated post-training (i.e., off-policy) optimization ($\S$~\ref{sec:policy_off_policy}), as well as various regularization techniques such as entropy and KL ($\S$~\ref{sec:policy_regular}).

\subsubsection{Policy Gradient Objective}
\label{sec:policy_gradient}
As introduced in $\S$~\ref{sec:pre_background}, the context in RL for LLMs is treated as the environment, and the probability distribution of the next-level prediction is treated as a policy. For an RL system, the objective of the system is to find an optimal policy such that the expected cumulative reward generated by the system is maximized. 
The RL policy optimization algorithms for LLMs are mostly first-order gradient-based algorithms, due to the large number of parameters in the LLMs. In general, RL algorithms seek to optimize network parameters such that the expected reward is maximized. Below, we present a general formulation for LLM gradient calculation of RL algorithms.

\paragraph{Notations.}
Although we have introduced the relevant symbols in $\S$~\ref{sec:pre_background}, we revisit these definitions here for the sake of comparative clarity.
Let $x \sim \mathcal{D}$ be a prompt (initial state $s_1 = s$). A stochastic policy $\pi_{\theta}$ generates a sequence $y=(a_1, \cdots, a_T)$, we denote the total sequence length of $y$ as $|y|$, with states defined by $s_{t+1} = (x, s_{\le t})$. We assume a primarily sequence-level reward $R(x,y)$, optionally decomposed into token-level rewards $r_t$.
We collect $G \geq 1$ responses per prompt using a \textit{behavior policy} $\pi_b$ (also denoted as $\pi_\textit{old}$, referring to an earlier version of the current policy).
Optionally, a \textit{reference policy} $\pi_{\textit{ref}}$ (e.g., base, finetuned or instructed models) may be used for regularization.

We revisit the MDP defined in $\S$~\ref{sec:pre_background}. In MDPs, we denote the expected cumulative reward given the current state $s$ as the V (value) function:
\begin{equation}
    \label{eq:v_function}
    V(s) = \mathbb{E}_{a_t \sim \pi_\theta(s_t), s_{t+1} \sim \mathcal{P}(s, a)} [\sum_{t=0}^T \gamma^t r(s_t, a_t) | s_0 = s],
\end{equation}
and the expected cumulative reward for the current state-action pair is denoted as Q (quality) function:
\begin{equation}
    \label{eq:q_function}
    Q(s, a) = \mathbb{E}_{a_t \sim \pi_\theta(s_t), s_{t+1} \sim \mathcal{P}(s, a)} [\sum_{t=0}^T \gamma^t r(s_t, a_t) | s_0 = s, a_0 = a].
\end{equation}

Then the objective of RL can be formulated as a maximization problem for the expected cumulative reward. 
To optimize the objective function, it is a common practice to use the Policy Gradient algorithm~\citep{williams1992simple,sutton1999policy} for gradient estimation:
\begin{equation}
    \nabla_{\theta} \mathcal{J} (\theta) = \mathbb{E}_{x \sim \mathcal{D}, y \sim \pi_{\theta}} \left[ \sum_{t=1}^T \nabla_{\theta}  \pi_{\theta} (y_t | y_{<t}) Q_t \right].
\end{equation}

 The policy gradient can be justified by the intuition that an algorithm following the policy gradient should maximize the probability of better-than-average actions and minimize the probability of worse-than-average actions. This notion led to the introduction of the $A$ (advantage) function $A(s, a) = Q(s, a) - V(s)$. The advantage measures how much the current action improves upon the expected total reward compared to the existing policy.
 The advantage can be estimated in many ways. If we only have rewards for the full trajectory, the vanilla REINFORCE algorithm~\citep{williams1992simple} directly defines $A_t = R(x, y)$.

For the case of training LLMs, the vanilla policy gradient algorithms often suffer from stability issues. Instead, the training is often done with the PPO algorithm \citep{schulman2017proximal}. 
For an algorithm with $N$ samples, we define a general objective with PPO-style updates as follows:
\begin{equation}
\mathcal{J}(\theta) = \mathbb{E}_{\text{data}} \left[ \frac{1}{Z} \sum_{i=1}^{N} \sum_{t=1}^{T_i} \min \left( w_{i,t}(\theta) \hat{A}_{i,t}, \, \text{clip}(w_{i,t}(\theta), 1 - \epsilon_{\text{low}}, 1 + \epsilon_{\text{high}}) \hat{A}_{i,t} \right) \right],
\end{equation}
where:
\begin{itemize}
  \item \( w_{i,t}(\theta) \) is the importance ratio;
  \item \( \hat{A}_{i,t} \) is the advantage (either token-wise or sequence-level);
  \item \( T_i \) is the number of tokens or responses per sample;
  \item \( N \) is the total number of samples under the given prompt;
  \item \( Z \) is the normalization factor (e.g., total tokens, group size, etc.).
\end{itemize}

The PPO algorithm \citep{schulman2017proximal} was first proposed as a computationally efficient approximation for the TRPO algorithm \citep{schulman2015trust}. PPO excels when vanilla policy gradient methods suffer from poor data efficiency and robustness issues. In addition, PPO is shown to be much simpler to implement, more general, and has better sample complexity compared to TRPO.

However, since the complex and long CoT nature of LLMs, the exact objective function, gradient estimation, and update techniques can take a wide range of different forms as shown in Table~\ref{tab:algo}.

\begin{table}[!t]
\centering
\caption{Comparison of representative RL algorithms for reasoning models training.
}
\label{tab:algo}
\begingroup
\renewcommand{\arraystretch}{1.2}
\resizebox{0.88\textwidth}{!}{
\begin{tabular}{llccc}
\toprule
\textbf{Date} & \textbf{Algorithm}  & \textbf{Advantage Estimate} & \textbf{Importance Sampling} & \textbf{ Loss Agg.}\\
\midrule
2017.01  &  PPO       &     Critic-GAE 
&  PPO-Style     & Token-Level  \\
2023.10  &  ReMax    &   Greedy Baseline   &    N/A      & Token-Level  \\
2024.02  &  RLOO     &  Leave-One-Out   &    N/A      & Token-Level \\
2025.01  &  RF++     &  Negative KL + Batch Relative   &  PPO-Style    & Sequence-level  \\
2024.02  &  GRPO     &  Group Relative   &     PPO-Style  & Sequence-level  \\
2025.01  &  PRIME    &  Outcome + Implicit PRM   &    PPO-Style      &Token-Level     \\
2025.03  &  VAPO     &  Value Adjusted GAE   &    Clip-Higher       &Token-Level             \\
2025.03  &  Dr. GRPO  &  Group Baseline   &  PPO-Style   &Token-Level        \\
2025.04  &  DAPO     &  Group Relative   &      Clip-Higher      &Token-Level            \\
2025.05  &  Clip-Cov &  Group Relative   & PPO-Style   & Sequence-level          \\
2025.05  &  KL-Cov   &  Group Relative   &  PPO-Style  & Sequence-level          \\
2025.06  &  CISPO    &  Group Relative   &  Clipped IS-weight   &  Token-Level         \\
2025.07  &  GSPO     &  Group Relative   &   PPO-Style   &  Sequence-level        \\
2025.08  &  GMPO     &  Group Relative   &  Clip-Wider   & Geometric-Avg       \\
2025.08  &  GFPO     &  Filter + Group Relative   &   PPO-Style     &   Token-level      \\
2025.08  &  LitePPO  &  Group-level mean, Batch-level std   &   PPO-Style   &Token-level        \\
2025.08  &  FlashRL  &  Group Relative   &   Truncated IS    &  Token-level   \\
2025.09  &  GPPO     &  Group Relative   &  Grad-Preserving Clip  & Sequence-level \\
2025.09  &  GEPO     &  Group-level mean  &   Group Expectation   &  PPO-Style  \\
2025.09  &  SPO      &  Entire Batch-level  &   PPO-Style   &  Sequence-level \\
\bottomrule
\end{tabular}
}
\endgroup
\end{table}

\subsubsection{Critic-based Algorithms}
\label{sec:policy_critic_based}

\begin{myboxi}[Takeaways]
\begin{itemize}
    \item The critic model is trained on a small subset of labeled data, and provides scalable token-level value signals for unlabeled roll-out data.
    \item The critic is required to run and update alongside the LLM, resulting in a significant computational overhead and scales unfavorably for complex tasks.
\end{itemize}
\end{myboxi}

The first LLM-related works in RL focus on how to effectively align the LLM policy to the external supervision, to make LLMs have better instruction following capabilities while ensuring the models are helpful, honest, and harmless.
The most common approach for LLM alignment is RLHF~\citep{christiano2017deep, ouyang2022training, stiennon2020learning, bai2022training}. This technique utilizes humans as a critic for the learning algorithm; the exact steps are as follows.
First, a selection of model outputs is generated by the LLM and labeled by humans to create a dataset. The dataset is then used to train a reward model to predict which response would be preferred by humans. 
Lastly, the reward model is used to train the LLM along with a value function, acting as the critic in the system. The training is often done with the PPO algorithm \citep{schulman2017proximal}. The PPO algorithm formulates the objective in the following form:
\begin{equation}
\label{eq:ppo}
\mathcal{J}_{\text{PPO}}(\theta) = \mathbb{E}_{x \sim \mathcal{D}, y \sim \pi_{\theta_{\textit{old}}}(\cdot | x)} \left[ \frac{1}{|y|} \sum_{t=1}^{|y|} \min \left( w_t(\theta) \hat{A}_t, \, \text{clip}(w_t(\theta), 1 - \epsilon, 1 + \epsilon) \hat{A}_t \right) \right],
\end{equation}
where \( \hat{A}_t \) is a value-model-based advantage and:
\begin{equation}
    \label{eq:ppo_is}
    w_t(\theta) = \frac{\pi_\theta(y_t | x, y_{<t})}{\pi_{\theta_{\textit{old}}}(y_t | x, y_{<t})}.
\end{equation}

We note that PPO is proposed as a clipped surrogate objective of TRPO, which preserves the conservative policy iteration of TRPO while being unconstrained and having a computational complexity close to traditional policy gradient methods. Due to the discrepancy between the current policy and the sampling distribution, the advantage in TRPO is multiplied by $w_t$, the importance sampling factor in Equation~\ref{eq:ppo}. PPO maximizes the same objective as TRPO, but removes the trust region constraint. Furthermore, PPO adds a clipping mechanism and a KL regularization factor to ensure the current policy does not diverge too far from the rollout policy $pi_{\theta_{old}}$.

In critic-based approaches, the scalability of RL is achieved by the introduction of a critic model. After the reward model is sufficiently trained on the manually labeled small subset of generated data, it can be used to construct the critic model, generating token-level value signals on a much larger scale for the vast majority of unlabeled generated data for RL. However, these works require a critic model to run and optimize along the target LLM, and create a significant computational overhead.

In PPO, the critic model adapts the Generalized Advantage Estimator (GAE) \citep{schulman2015high} from the RL literature. 
GAE is typically constructed with the temporal difference error 
\begin{equation}
    \label{eq:td_error}
    \delta_{t} = r_t + \gamma V(y_{t+1}) - V(y_t),
\end{equation}
which is then accumulated across time steps: 
\begin{equation}
    \label{eq:gae}
    \hat{A}_{GAE, t} = \sum_{l=t}^T (\gamma \lambda)^l \delta_{t+l},
\end{equation}
where $\gamma$ is the discount factor of the MDP and $\lambda$ is a parameter that controls the bias-variance tradeoff.

Recent work has argued that the decay factor scales unfavorably for complex reasoning tasks that require long CoT and proposed a Value-Calibrated PPO \citep{yuan2025s} and VAPO \citep{yue2025vapo}, VRPO \citep{zhu2025vrpo} proposed novel mechanisms for enhancing the robustness of the critic model under noisy reward signals.

In addition, critic-based algorithms \citep{hu2025open} have also demonstrated steady scalability properties for Monte-Carlo estimation with rule-based rewards. Similar approaches have been adapted with fixed external models \citep{wang2023math,lu2024autopsv} by the implementation of PRMs.

Another approach to introduce critic models is done with the introduction of Implicit PRM \citep{yuan2024free}. This approach is also able to provide token-level supervision for scalable RL training. Different from the GAE approach, methods such as Implicit PRM \citep{yuan2024free} and PRIME \citep{cui2025process} adapted a specific reward model formulation to directly generate token-level rewards.

\subsubsection{Critic-Free Algorithms}
\label{sec:policy_critic_free}

\begin{myboxi}[Takeaways]
\begin{itemize}
    \item Critic-free algorithms only require sequence-level rewards for training, making them more sufficient and scalable.
    \item For RLVR tasks, rule-based training signals reliably prevent critic-related issues such as reward hacking.
\end{itemize}
\end{myboxi}
Apart from the critic-based models, which provide token-level feedback signals for model training, many recent works have stated that the response-level rewards are sufficient for scalable reasoning tasks with RL. These critic-free algorithms apply the same rule-based or model-generated response-level reward for all tokens in the response and demonstrate their effectiveness across various tasks.
Compared to the critic-based algorithms, critic-free approaches do not require a separate critic model, significantly reducing the computational requirement and simplifying training. 
Moreover, when training LLMs in rule-based environments where the reward for any response can be clearly defined, critic-free algorithms can avoid reward hacking issues that may arise from an ill-trained critic model. This property makes critic-free algorithms more scalable than critic-based approaches in such settings.

The classic REINFORCE \citep{williams1992simple} algorithm was among the first algorithms developed for RL. It was applied to the LLM problem in \citep{ahmadian2024back}. The exact formulation for REINFORCE is as follows:
\begin{equation}
\mathcal{J}_{\text{REINFORCE}}(\theta) = \mathbb{E}_{x \sim \mathcal{D}, \{y\} \sim \pi_{\textit{old}} (\cdot | x)} \left[   R(x, y) \nabla_\theta \log(\pi_\theta (y|x)) \right],
\end{equation}
where $R(x,y)$ usually takes the form of $\pm 1$ for RLVR tasks. This naive formulation takes the entire sequence as a single action and considers the response task as a bandit.
However, the vanilla algorithm usually suffers from severe instability issues due to high variance. ReMax~\citep{li2023remax} introduced a variance reduction mechanism to REINFORCE with a greedy baseline estimation. \citet{ahmadian2024back} also introduced RLOO, which further provides an unbiased baseline with more stable results. REINFORCE++~\citep{hu2025reinforce++} adapts techniques such as clipping and global advantage normalization from PPO and GRPO style algorithms to provide a more accurate advantage and gradient estimations.

One of the most popular critic-free approaches for RL is GRPO \citep{shao2024deepseekmath}. The objective formulation for GRPO is as follows:
\begin{equation}
\mathcal{J}_{\text{GRPO}}(\theta) = \mathbb{E}_{x \sim \mathcal{D}, \{y_i\}_{i=1}^G \sim \pi_{\theta_{\textit{old}}} (\cdot | x)} \left[ \frac{1}{G} \sum_{i=1}^{G} \frac{1}{|y_i|} \sum_{t=1}^{|y_i|} \min \left( w_{i,t}(\theta) \hat{A}_{i,t}, \, \text{clip}(w_{i,t}(\theta), 1 - \epsilon, 1 + \epsilon) \hat{A}_{i,t} \right) \right],
\end{equation}
\begin{equation}
    \label{eq:w_and_a}
    w_{i,t}(\theta) = \frac{\pi_\theta(y_{i,t} | x, y_{i,<t})}{\pi_{\theta_{\textit{old}}}(y_{i,t} | x, y_{i,<t})}, \quad
    \hat{A}_{i,t} = \hat{A}_i = \frac{R(x, y_i) - \text{mean}(\{R(x, y_i)\}_{i=1}^G)}{\text{std}(\{R(x, y_i)\}_{i=1}^G)},
\end{equation}
where all the tokens in $y_i$ share the same advantage as $\hat{A_i}$.

GRPO is a critic-free modification of PPO, where instead of GAE provided by a critic, the entire sequence uses the same advantage estimate, which is calculated by a group-relative normalization as a better estimation than the binary rule-based reward. 
Compared to PPO and REINFORCE-style methods, the group-based advantage calculation of GRPO effectively reduces variance from training signals and has been shown to speed up the training process.
Other recent approaches, including DAPO \citep{yu2025dapo}, CISPO \citep{chen2025minimax}, Dr. GRPO \citep{liu2025understanding}, LitePPO \citep{liu2025part}, made further modifications to GRPO with careful tuning of sampling strategy, clipping threshold, and loss normalization to further enhance the stability of the RL training process. Another recent approach, GSPO \citep{zheng2025group}, replaces the token-wise clipped importance sampling ratio with a sequence-level clipping. 

Apart from REINFORCE and GRPO-related algorithms, there are other critic-free approaches.
VinePPO modifies PPO by replacing the learned critic with a Monte Carlo advantage estimation.
CPGD~\citep{liu2025cpgd} proposed a novel policy gradient objective, along with a drift regularization mechanism.
K1.5 \citep{team2025kimi} utilizes RL with an adaptation of mirror descent in the training of foundational models, which successfully enhanced the long-context reasoning capabilities of LLMs.
\citet{lv2025towards} have recently introduced a unified policy gradient estimator with a hybrid post-training algorithm, providing a unified framework for policy gradient estimation for RL in LLMs.
SPO~\citep{xu2025singlestream} introduces a group-free, single-stream policy optimization that replaces per-group baselines with a persistent KL-adaptive value tracker and global advantage normalization, yielding smoother convergence and higher accuracy than GRPO while scaling efficiently in long-horizon and tool-integrated settings.
HeteroRL~\citep{zhang2025group} decouples rollout sampling from parameter learning for decentralized asynchronous training and, via GEPO, reduces importance-weight variance under latency-induced KL drift (theoretically exponential), maintaining stability even under severe delays (e.g., <3\% degradation at 1,800s).
GPPO~\citep{su2025klearreasoner} introduce a gradient-preserving clipping scheme for GRPO/PPO that keeps the forward clipped objective unchanged while—via stop-gradient decoupling—replacing zeroed gradients outside the clip range with bounded constants, thereby retaining informative out-of-bound signals and maintaining PPO-style stability.
\citet{dwyer2025s} propose Probability Smoothing Policy Optimisation (PSPO) for stabilizing policy updates in RL fine-tuning of LLMs using soft trust regions and improving performance.
FlowRL~\citep{zhu2025flowrl} introduces a flow-balanced optimization approach that matches complete reward distributions rather than maximizing scalar rewards, resulting in more diverse reasoning patterns. This fundamental shift addresses the mode collapse problem inherent in reward-maximizing RL methods.

\paragraph{Importance Sampling for Policy Optimization.}
Due to the rollout-reward-training cycle for RL, it is generally computationally intractable to ensure the rollout data follows the exact policy distribution of the current model. Therefore, importance sampling was introduced to reduce bias in training. The first version of importance sampling in RL was introduced in TRPO, where a token-wise importance ratio $w_{i, t}$ was introduced into the objective. This approach is widely adopted among recent works, such as GRPO. 
This approach is restricted to the token-wise importance ratio since the actual distribution ratio can not be effectively calculated over the long context of CoT.
However, token-level importance sampling introduces another bias into RL algorithms, since the actual sampling distribution given policy is defined with respect to the state-action pair, whereas the token-level approach only considers the current action.
GMPO~\citep{zhao2025geometric} seeks mitigation by introducing a geometric averaging to increase training robustness for tokens with extreme importance sampling ratios.
In the recent work of GSPO~\citep{zheng2025group}, a sequence-level importance sampling factor was calculated. GSPO adds a unique normalization factor to ensure that the probability ratio can be calculated, but this approach is also a biased estimation of the actual importance sampling factor.
A promising new direction is to move beyond the theoretical framework of standard on-policy policy gradient methods and instead derive inherently off-policy algorithms directly from supervised learning theory~\citep{chen2025bridging}.
We will provide a detailed introduction to off-policy optimization in the next section.

\subsubsection{Off-policy Optimization}
\label{sec:policy_off_policy}
\begin{myboxi}[Takeaways]
\begin{itemize}
    \item Off-policy RL boosts sample efficiency by decoupling data collection from policy learning, enabling training from historical, asynchronous, or offline datasets.
	\item Modern practice mixes off-policy, offline, and on-policy methods (e.g., SFT+RL or large-scale offline learning) to improve stability and performance.
\end{itemize}
\end{myboxi}

In RL, off-policy methods address the scenario where the policy being learned (the target policy) differs from the policy generating the data (the behavior policy). This core distinction allows an agent to learn about an optimal course of action without having to follow it during data collection. This flexibility is a key advantage, often leading to more sample-efficient algorithms than on-policy counterparts, which require new data sampled directly from the current policy for each update.
A core challenge in these methods is correcting for the distributional shift between the behavior policy and the target policy, often addressed using importance sampling with a weighted objective function:
\begin{equation}
    \label{eq:off_policy}
    \mathcal{L}_{\text{policy}}(\theta) = -\mathbb{E}_{x \sim \mathcal{D}, y \sim \pi_{\textit{b}}(y|x)} \left[ \frac{\pi_{\theta}(y|x)}{\pi_{\textit{b}}(y|x)} \cdot r(x,y) \right],
\end{equation}
where the fraction $\frac{\pi_{\theta} (y|x)}{\pi_{b} (y|x)}$ serves as the importance weight between the target policy $\pi_{\theta}$ and the behavior policy $\pi_{\textit{b}}$. 

In practical large-scale model training, off-policy learning often manifests in different forms. Recent works can be broadly grouped into three aspects: 1) training–inference precision discrepancies, where models are trained with high precision but deployed in lower precision, creating a gap between the target and behavior policies; 2) asynchronous experience replay mechanisms, which enhance efficiency and stability by reusing past trajectories during learning; and 3) broader off-policy optimization approaches, including optimizer-level improvements, data-level offline learning, and hybrid methods that combine supervised fine-tuning with RL.

\paragraph{Training-Inference Precision Discrepancy.}
A notable off-policy scenario arises from the difference in parameter precision between the training model and the inference model, employing different frameworks for training and inference~\citep{yao2025offpolicy} (e.g., vLLM vs. FSDP), or of model quantization to accelerate inference~\citep{lin2016fixed}, which are the manifestations of nondeterminism in LLM inference~\citep{he2025nondeterminism}. It is common practice to train a model using high-precision parameters (e.g., 32-bit floating point) and then deploy a quantized version with lower-precision parameters (e.g., 8-bit integers)~\citep{yao2025flashrl}. This creates a discrepancy where the deployed, low-precision model acts as the behavior policy, generating real-world interaction data, while the high-precision model remains the target policy being updated during training. While this mismatch establishes an off-policy learning problem, research indicates that the policy divergence due to quantization is often minimal. Consequently, this difference can be effectively managed with simple correction techniques, such as truncated importance sampling (TIS)~\citep{ionides2008truncated,yao2025offpolicy}, allowing for stable training while retaining the benefits of accelerated inference.

\paragraph{Asynchronous Off-Policy Training.}
Asynchronous training pairs naturally with off-policy RL for LLMs. Many actors generate trajectories concurrently and append them to a shared replay buffer, while a centralized learner samples mini-batches from this buffer to update the target policy. Building on this view, several recent methods deliberately reuse past trajectories to improve efficiency and stability. One example is Retrospective Replay \citep{dou2025improving}, which enhances exploration for LLM reasoning by selectively replaying earlier reasoning traces to guide current policy updates. Similarly, EFRame \citep{wang2025eframe} adopts an exploration-filter-replay mechanism, interleaving filtered responses with fresh rollouts to encourage deeper reasoning. In the domain of code generation, Possibility- and Pass-rate Prioritized Experience Replay (PPER) \citep{chen2024enhancing} takes this further by prioritizing high-value code samples in the buffer, leading to more stable optimization. Extending these ideas to multimodal interaction, ARPO \citep{lu2025arpo} applies replay to GUI agents, where successful trajectories are reused to provide reliable learning signals under sparse rewards. Finally, RLEP \citep{zhang2025rlep} anchors exploration with an experience buffer of verified successful trajectories from earlier runs, which are blended with new rollouts to balance reliability with discovery. Together, these approaches illustrate how replay buffers have become a cornerstone of modern, asynchronous off-policy training for LLM-based agents.

\paragraph{Off-Policy Optimization.} 
Recent advancements in fine-tuning LLMs have explored sophisticated optimization strategies beyond traditional on-policy RL. These methods, broadly categorized as off-policy and mixed-policy optimization, aim to improve sample efficiency, training stability, and overall performance by creatively using data from various sources. We introduce this topic below:

\begin{itemize}
    \item \textbf{Optimizer-Level Off-Policy Methods:}  
    These approaches focus on improving the optimization procedure itself, emphasizing stability and efficiency in policy updates. For example, SPO \citep{cohen2025soft} introduces a soft policy optimization method that enables stable online, off-policy RL, while TOPR \citep{roux2025tapered} proposes a tapered off-policy REINFORCE algorithm for improved stability and efficiency. ReMix \citep{liang2025squeeze} further highlights this by focusing on efficiently leveraging off-policy data to maximize the utility of available information. 

    \item \textbf{Data-Level Off-Policy Methods:}  
    A class of off-policy algorithms learns entirely from large-scale, external offline data \citep{zhang2025stephint}. 
    For instance, the Dynamic Fine-Tuning (DFT) framework~\citep{wu2025generalization} generalizes the SFT loss to an RL formulation and introduces a stop-gradient mechanism, enabling training on offline data as in SFT, while yielding improved performance. Building on offline data as well, Intuitive Fine-Tuning (IFT)~\citep{hua2024intuitive} adds a temporal residual connection that fuses SFT and RLHF objectives and explicitly models and optimizes the influence of the current token on all future generations. Another pertinent approach is Direct Preference Optimization (DPO)~\citep{rafailov2023direct}, which directly optimizes the policy from preference data.
    These methodologies collectively represent a move towards more data-centric approaches in RL, enabling the development of sophisticated policies from vast and diverse sources of offline data. 

    \item \textbf{Mix-Policy Methods:} 
    In parallel with reusing past data more efficiently, mixed-policy optimization represents another significant trend, which combines the strengths of SFT and RL. This hybrid approach leverages the stability from SFT on expert data while using RL to optimize for specific reward functions, integrating the supervised data in two primary ways. One strategy is at the loss-level, where SFT and RL objectives are combined directly in the loss function \citep{lv2025towards, zhang2025policy, xiao2025connection}.
    Methods like UFT \citep{liu2025uft}, SRFT \citep{fu2025srft}, LUFFY \citep{yan2025learning}, RED~\citep{guan2025recall}, and ReLIFT \citep{ma2025learning} all exemplify this by creating unified or single-stage training processes that learn from both expert demonstrations and RL feedback simultaneously. A second strategy operates at the data level, using expert data to structure the generation process itself. Here, high-quality data serves as a prefix or anchor to guide the model's exploration~\citep{guo2025g}. For instance, BREAD \citep{zhang2025bread} generates branched rollouts from expert anchors, and Prefix-RFT \citep{huang2025blending} blends the training regimes via prefix sampling. By mixing policies at either the loss or data level, these methods prevent reward hacking and ensure the model retains knowledge from SFT, leading to more robust and capable models for complex reasoning.
\end{itemize}

\subsubsection{Regularization Objectives}
\label{sec:policy_regular}

\begin{myboxi}[Takeaways]
\begin{itemize}
    \item Objective-specific regularization helps balance exploration and exploitation, boosting RL efficiency and policy performance.
	\item The optimal choice and form of KL, entropy, and length regularization remain open questions, each affecting policy optimization and scalability.
\end{itemize}
\end{myboxi}

As introduced in previous sections, ensuring stability and preventing catastrophic policy drift is paramount. In particular, for long-horizon training, techniques such as KL regularization and entropy regularization are widely employed.

\paragraph{\textbf{KL Regularization}.}
The role of KL divergence regularization is a highly controversial topic in this area.
In most studies, KL regularization is applied to 1). current policy $\pi_\theta$ and the reference policy $\pi_\textit{ref}$, 2). current policy $\pi_\theta$ and the old policy $\pi_\textit{old}$.
We provide a unified formulation in Equation~\ref{eq:kl}.
\begin{equation}
\label{eq:kl}
    \mathcal{L}_{\textit{KL}} = \beta \sum_{t=1}^{|y|} KL(\pi_{\theta} (\cdot | y_t) || \pi_{\textit{ref/old}} (\cdot | y_t)).
\end{equation}
\begin{itemize}
    \item
    For the former, this is a commonly used technique in RLHF~\citep{ouyang2022training, touvron2023llama}.
    It was initially introduced to prevent the model from being destructively updated.
    Prior work argues that incorporating a KL penalty is essential for maintaining stability and avoiding entropy collapse over thousands of training steps.
    To reduce the risk of the KL term excessively constraining progress, \citet{liu2025prorl} use this method combined with a periodic reference policy reset, in which the reference model is updated to a recent snapshot of the training policy.
    To simultaneously maintain knowledge and enhance reasoning capabilities, \citet{wang2025stabilizing} apply stronger KL regularization to low-entropy tokens and weaker regularization to high-entropy tokens.
    However, in the context of RL for reasoning with LLMs, which is more challenging than standard RLHF, the necessity of this kind of KL regularization needs to be reconsidered.
    Recently, many studies have identified that the policy is expected to explore freely during training, thus may diverge significantly from its initialization to discover new CoT structures, making the KL constraint an unnecessary restriction. Thus, a majority of other recent works advocate for removing the KL penalty entirely~\citep{yu2025dapo, fan2025truncated, liu2025understanding, he2025skywork, liao2025enhancing, chen2025acereason, an2025polaris, cui2025process, arora2025training, yan2025learning} to simplify implementation, reduce memory cost and achieve more scalable GRPO.

    \item For the latter case, it can serve as a substitute for the clip form of the policy loss~\citep{schulman2017proximal}.
    \citet{zhang2025design} discusse the differences between forward KL, reverse KL, normalized KL, and Normalized forms.
    This approach has also been adopted in~\citet{cui2025entropy, team2025kimi, lyu2025exploring}, demonstrating its potential across different RL training scales.
    Nevertheless, its deeper mechanisms and its significance for scalable RL remain under exploration.
\end{itemize}

\paragraph{\textbf{Entropy Regularization}.}
In the RL literature, preserving policy entropy is widely considered a critical aspect of many algorithms~\citep{williams1991function, williams1992simple, eysenbach2021maximum}.
To this end, policy entropy is actively controlled through regularization techniques~\citep{ziebart2008maximum, schulman2017proximal, haarnoja2018soft}.
\begin{equation}
    \mathcal{L}_{\text{ent}} = - \alpha \sum_{t=1}^{|y|} H [\pi_{\theta}(\cdot | y_t)] = \alpha \sum_{t=1}^{|y|}{\sum_{v=1}^{|\mathcal{V}|} {\pi_{\theta} (y_t^v | y_t) \log \pi_{\theta} (y_t^v | y_t).}}
\end{equation}

However, in RL for LLMs, directly applying entropy regularization is neither common nor effective~\citep{cui2025entropy, he2025skywork}.
The use of an explicit entropy regularization term in the loss function remains a point of contention. While some find it beneficial, using either a standard coefficient \citep{shrivastava2025sample} or a targeted loss function \citep{wu2025confucius3}, others argue against it, finding it can lead to instability or even training collapse, especially with sparse rewards \citep{liao2025enhancing, an2025polaris}.
Many studies have shown the phenomenon of entropy collapse when no intervention is applied~\citep{cui2025entropy, yu2025dapo, cheng2025reasoning}, which hinders effective policy exploration during training.
To address it, \citet{he2025skywork} dynamically adjust the coefficient of the entropy loss, \citet{yu2025dapo} employs the clip-higher technique to involve more low-probability tokens in the policy update, \citet{wang2025beyond} directly train on 20$\%$ high-entropy tokens, \citet{cheng2025reasoning} and \citet{chen2025seed} emphasize entropy through incorporate it into the advantage computation.
Beyond these techniques, which explicitly maximize entropy, \citet{cui2025entropy} provide a theoretical explanation for the underlying mechanism of entropy dynamics, identifying the covariance between an action’s output probability and its advantage as the entropy ``driver''. Built on this insight, Clip-Cov and KL-Cov are proposed to regulate entropy by selectively constraining a small portion of tokens exhibiting exceptionally high covariance.

\paragraph{\textbf{Length Penalty}.}
Recent successes of LRMs on complex tasks have validated the effectiveness of long-CoT reasoning. Yet longer reasoning traces incur higher inference costs.
To balance the reasoning budget and performance~\citep{he2025thinkdial, openai2025gpt-oss-120b}, many works seek to reduce the reasoning cost while retaining the model performance~\citep{xiang2025just, liu2025learn, su2025between, aggarwal2025l1, luo2025o1}. For example, \citet{aggarwal2025l1} control reasoning length by ensuring adherence to user-specified length constraints, while \citet{yuan2025efficient} and \citet{luo2025o1} design relative-length regularization and an accuracy-preservation constraint to the optimization objective, \citet{xiang2025just} and \citet{liu2025learn} propose to apply adaptive length penalties conditioned on problem difficulty to preserve the model ability.

\subsection{Sampling Strategy}
\label{sec:sampling}
Unlike static datasets, RL depends on actively curated rollouts, where decisions about what and how to sample directly influence learning efficiency, stability, and the quality of acquired reasoning behaviors. Effective sampling strategies not only ensure diverse and informative training signals but also align the learning process with the intended reward structure and policy objectives. 
In this subsection, we survey recent advances in dynamic and structured sampling ($\S$~\ref{sec:sampling_dynamic}), as well as hyperparameter adjustment techniques that further optimize sampling and policy improvement ($\S$~\ref{sec:sampling_hyper_params}).

\subsubsection{Dynamic and Structured Sampling}
\label{sec:sampling_dynamic}

\begin{myboxi}[Takeaways]
\begin{itemize}
    \item High-quality, diverse rollouts stabilize RL training and enhance overall performance by exposing agents to a broader range of meaningful experiences.
    \item  Balancing the exploration of diverse trajectories with maintaining high sampling efficiency presents a fundamental trade-off in RL.
\end{itemize}
\end{myboxi}

Sampling has become a first-class lever in RL fine-tuning for reasoning LLMs, serving as an efficient and adaptive mechanism to maximize data utilization, reduce wasted computation, and enhance training effectiveness or a control and a guidance for LLMs to sample in a structured format.

\paragraph{Dynamic Sampling.}
Dynamic sampling adapts both the selection of prompts for rollout and the computational budget allocated to each, based on online learning signals such as success rate, advantage, uncertainty, or estimated difficulty. The primary goal is to concentrate computing on informative examples while avoiding saturated or unproductive ones. Existing methods generally fall into two categories:
\begin{itemize}
    \item \textbf{Efficiency-oriented Sampling:} Some works use online-filtering to concentrate training on questions of medium difficulty to ensure training effectiveness and efficiency. A representative design is PRIME \citep{cui2025process}, which applies an online filter to drop out too easy or too difficult problems. Another example is DAPO~\citep{yu2025dapo}, which over-samples and filters prompts whose rollouts are saturated (all-correct) or degenerate (all-wrong), then repeatedly samples until each mini-batch contains prompts with non-zero advantage, focusing on medium-difficulty cases to maintain informative gradients.
    Building on this foundation, prioritized schemes allocate rollout budget toward under-mastered items by sampling proportional to failure rates, as $p(i)\!\propto\!(1-s_i)$ rule \citep{team2025kimi}. 
    Curriculum learning approaches operate at multiple scales: category-level selection~\citep{chen2025self} uses non-stationary bandits, while E2H~\citep{parashar2025curriculum} follows easy-to-hard schedules with convergence guarantees for small models.
    Efficiency methods include pre-rollout selection to skip unhelpful prompts and difficulty-based online selection with rollout replay~\citep{zheng2025act,sun2025improving}. POLARIS~\citep{an2025polaris} formalizes this via offline difficulty estimation, constructing ``mirror-J'' distributions by model scale, continuously removing mastered items, and applying in-batch information replacement.
    AttnRL~\citep{liu2025attention} estimates problem difficulty online and leverages attention scores to filter out some easy problems for sampling. Additionally, AttnRL introduces an adaptive batch sampling mechanism, which estimates the prompt batch size based on historical valid training batch size, which is computed after filtering all responses with zero advantage values.
    Extending these efficiency gains, recent advances use lightweight controllers for adaptive sampling~\citep{shi2025efficient,do2025sparft} without modifying algorithms, while experience replay with random reshuffling~\citep{fujita2025experience} reduces variance through balanced utilization, and enhanced prioritized methods~\citep{li2024prioritized} dynamically adjust priority weights based on experience pool features.
    Sampling efficiency can also be improved by structuring the generation process with expert data: high-quality demonstrations are used as prefix anchors to bias exploration toward promising regions of the search space \citep{guo2025g, zhang2025bread, huang2025blending}.
    \cite{zhou2025april} introduce APRIL, a strategy to reduce GPU idle time and improve rollout efficiency in RL training by over-provisioning requests and recycling incomplete responses.
    The field shifts from uniform sampling to model-aware strategies combining item-, category-, and difficulty-level choices for stronger learning signals per rollout.
    \item \textbf{Exploration-oriented Sampling:} There are other works aiming for exploration using dynamic rollout. 
    ARPO \citep{dong2025agentic} is proposed to implement entropy-guided rollout to ensure high uncertainty so that the model will call external tools, improving diversity, while AttnRL~\citep{liu2025attention} finds that steps with high attention scores are related to reasoning behaviors and branches at these steps for better exploration. DARS \citep{yang2025depth} proposes a rollout mechanism to dynamically assign sample numbers for questions of different difficulty. \citet{zhou2025breaking} propose RuscaRL by providing the policy with different rubrics during rollout to enhance exploration. Different from above, G$^2$RPO-A \citep{guo2025g} does not drop all-wrong questions, but add a guidance during the thinking process to generate correct samples for hard questions. Besides, \citet{li2025temporal} utilize the latest $k$ checkpoints to generate $k$ responses to prevent forgetting during training.
    Meanwhile, Parallel-R1 \citep{zheng2025parallel} instills ``parallel thinking'' via a progressive curriculum learning.
\end{itemize}

\paragraph{Structured Sampling.}
Structured sampling controls not only what is sampled but also the topology of reasoning traces, aligning generation, credit assignment, and compute reuse with the underlying structure of problem solving. By organizing rollouts as trees or through shared and segmented prefixes, these methods enable node-level rewards, improved reuse of partial computations (e.g., KV caches), and greater sample efficiency under memory and budget constraints. We highlight two representative approaches:
\begin{itemize}
    \item \textbf{Search-driven Tree Rollouts:} Other works leverage Monte Carlo Tree Search (MCTS) for tree-format response generation using the classic phases: initialization, selection, expansion, and backpropagation. They view a single inference as a tree rather than a single chain, and assign rewards at the node level, which can produce a more dense/fine-grained process signal. \citet{hou2025treerl} propose TreeRL, an on-policy tree search framework that outperforms traditional Chain-of-Thought RL (ChainRL) while substantially reducing computational overhead through more efficient search strategies. Concurrently, ToTRL \citep{wu2025totrl} introduces a Tree-of-Thought–guided training paradigm in synthetic puzzle environments, enabling emergent generalization to out-of-distribution tasks such as mathematical reasoning. Additionally, \citet{yang2025treerpo} integrate MCTS into training pipelines to generate rule-based, fine-grained process rewards, improving reward signal granularity and fidelity in policy optimization.
    \item \textbf{Shared-prefix or Segment-wise Schemes:} While these tree search methods enrich exploration and provide fine-grained rewards, their sample efficiency remains a limitation. Some works design segmented/shared prefix sampling to improve generation efficiency~\citep{guo2025segment, yang2025treerpo, hou2025treerl, li2025treepo}.
    SPO~\citep{guo2025segment}, TreeRPO~\citep{yang2025treerpo}, TreeRL~\citep{hou2025treerl}, FR3E~\citep{zheng2025first}, and ARPO~\citep{dong2025agentic} conduct additional sampling starting from previously generated prefix. 
    TreePO~\citep{li2025treepo} implements a segment-wise tree sampling algorithm that alleviates the KV cache burden, reducing the GPU hours for training, and improving sampling efficiency.
    \cite{ji2025tree} introduce Tree-GRPO, a tree-based RL method for improving agentic tasks in LLMs using structured sampling and process-level rewards derived from sparse outcome supervision.
\end{itemize}

\subsubsection{Sampling Hyper-parameters}
\label{sec:sampling_hyper_params}
\begin{myboxi}[Takeaways]
\begin{itemize}
    \item Careful hyperparameter tuning is essential for scalable RL, as naive settings can lead to inefficiency and unstable training (e.g., entropy collapse).
	\item Scalable RL relies on a holistic combination of strategies to balance cost and stability, such as staged context lengthening and dynamic exploration controls.
\end{itemize}
\end{myboxi}

This subsection summarizes the hyperparameter adjustment strategies for sampling from recent works.
Effective RL training requires a delicate balance between several competing objectives, and recent literature has focused on techniques across two primary axes:
1) managing the exploration-exploitation trade-off to ensure the model discovers and refines effective reasoning paths;
2) efficiently managing sequence length to balance reasoning depth with computational cost.

\paragraph{Exploration and Exploitation Dynamics.}
A central challenge is balancing exploration (discovering novel reasoning strategies) with exploitation (refining high-reward solutions). The primary levers for this are temperature, entropy regularization, and PPO's clipping mechanism. For temperature, strategies vary significantly. Some works propose a dynamic approach, such as staged temperature increases (e.g., $1.40 \rightarrow 1.45 \rightarrow 1.50$ for a 4B model, $0.7 \rightarrow 1.0 \rightarrow 1.1$ for a 7B model) to gradually expand trajectory diversity as training progresses \citep{an2025polaris}, or using a scheduler to dynamically adjust temperature to maintain a stable entropy level \citep{liao2025enhancing}. A more prescriptive approach recommends tuning the training temperature to keep the post-scaling entropy around a target of 0.3, which is found to strike an optimal balance \citep{liu2025acereason, wu2025confucius3}. Other works simply advocate for a high, fixed temperature (e.g., $1.0$ or $1.2$) to encourage initial exploration, while noting it is insufficient on its own to prevent long-term entropy decline \citep{shrivastava2025sample, arora2025training, liu2025prorl}.

\paragraph{Length Budgeting and Sequence Management.} Nearly all works grapple with managing the length of generated responses to balance performance and cost. The most prevalent strategy is staged context lengthening \citep{luo2025deepscaler}. This involves starting RL with a short context window (e.g., $8k$) before progressively increasing it to $16k$, $24k$, or $32k$ in later stages \citep{luo2025deepscaler, chen2025acereason, liu2025prorl, liu2025acereason}. The initial short-context stage is considered essential, as it forces the model to learn more concise and token-efficient reasoning patterns \citep{luo2025deepscaler, chen2025acereason, liu2025acereason}. An alternative to training on very long contexts is to apply inference-time length extrapolation techniques like Yarn at inference time, allowing a model trained on shorter sequences to generalize to longer ones \citep{an2025polaris}. For handling responses that exceed the length budget, there is no consensus. Some works apply a soft, linear penalty as the response approaches the maximum length \citep{yu2025dapo} or a tunable penalty ($\alpha$) directly in the reward function \citep{arora2025training}. A more nuanced, stage-dependent strategy is to filter (mask the loss of) overlong samples when the length budget is short ($8k$-$16k$) but to penalize them when the budget is large ($32k$), as filtering can become detrimental at very long contexts \citep{liu2025acereason, wu2025confucius3}.

Across these works, effective hyperparameter adjustment emerges as the joint tuning of exploration (temperature, entropy targets, clipping), efficiency (staged length curricula), and sequence management (overlength filters, penalties, or inference-time extrapolation). These methods are directly applicable to most GRPO/PPO-style RL pipelines for LLMs.

\section{Foundational Problems}
\label{sec:problems}

Having reviewed the key components of RL pipelines for LLMs, we now turn to several foundational problems that remain central and often unresolved in the field.
In this section, we articulate the core issues, present contrasting perspectives, and summarize recent progress on each open question. Specifically, we discuss challenges such as the fundamental role of RL (sharpening versus discovery) in $\S$~\ref{sec:problems_role}, the boundary between RL and SFT (generalization versus memorization) in $\S$~\ref{sec:problems_rl_vs_sft}, the selection of model priors (weak versus strong models) in $\S$~\ref{sec:problems_model_prior}, the effectiveness of training algorithms (tricks versus traps) in $\S$~\ref{sec:problems_training_recipes}, and the granularity of reward signals (process versus outcome) in $\S$~\ref{sec:problems_reward_type}. By highlighting these open questions, we aim to clarify the current landscape and motivate further investigation into the foundational underpinnings of RL for LRMs.

\subsection{RL's Role: Sharpening or Discovery}
\label{sec:problems_role}

We begin by summarizing the two prevailing perspectives on the role of RL: \textcolor{cyan!40!black}{Sharpening} and \textcolor{cyan!40!red}{Discovery}.
These perspectives appear to be in direct opposition.
The \textcolor{cyan!40!black}{Sharpening} view suggests that RL does not create genuinely novel patterns, but instead refines and reweights correct responses already contained within the base model.
By contrast, the \textcolor{cyan!40!red}{Discovery} view claims that RL is capable of uncovering new patterns that the base model does not acquire during pre-training and would not generate through repeated sampling.

The divergence between the \textcolor{cyan!40!black}{Sharpening} and \textcolor{cyan!40!red}{Discovery} perspectives can be understood through multiple theoretical lenses. First, from the KL divergence optimization viewpoint, SFT typically optimizes the forward KL divergence $D_{KL}(p_{data} || p_{model})$, exhibiting \emph{mode-covering} behavior: the model attempts to cover all modes in the data distribution.
In contrast, RL methods optimize the reverse KL divergence $D_{KL}(p_{model} || p_{reward})$, which exhibits \emph{mode-seeking} behavior: concentrating probability mass on high-reward regions~\citep{ji2024towards,sun2024supervised}.
Recent theoretical advances have further enriched this understanding. \cite{xiao2025connection} demonstrate that RLHF can be viewed as implicit imitation learning on preference data, establishing a deep connection between RL-based alignment and behavioral cloning. Similarly, \cite{sun2024supervised} frames SFT itself as a form of inverse RL, revealing that even supervised approaches implicitly involve reward modeling. These perspectives suggest that the \textcolor{cyan!40!black}{Sharpening} vs. \textcolor{cyan!40!red}{Discovery} debate may be addressing different aspects of a unified learning process: while the mode-seeking nature of RL provides a mechanism for sharpening, the implicit reward learning and compositional capabilities could enable discovery through extended training.

\begin{itemize}
    \item Initially, DeepSeek-R1~\citep{guo2025deepseek} demonstrated promising ``Aha'' behaviors through RLVR, inspiring lightweight reproductions such as TinyZero~\citep{tinyzero}, which reported similar phenomena with simplified training recipes and minimal code.
    Domain-specific adaptations soon followed, including Logic-RL~\citep{xie2025logic}, which showcased rule-based RL that fosters reflection and verification skills with transfer to mathematical reasoning.

    \item However, Limit-of-RLVR~\citep{yue2025does} provides a sharpening-oriented counterargument: \passk~evaluations indicate that RL enhances \passo~performance, yet tends to underperform relative to base models when sampling broadly at large-$k$ \passk. This suggests that RL predominantly narrows the search space rather than uncovering fundamentally novel solution trajectories.
    Concurrent debates questioned whether the observed ``Aha'' behaviors were genuinely induced by RL or merely latent capabilities already embedded during pre-training~\citep{liu2025there,setlur2025e3}.
    Mechanistic analyses further argued that RL gains often arise from entropy shaping or reward proxies.
    For instance, high-entropy ``forking'' tokens appear to dominate improvements~\citep{wang2025beyond}; maximizing model confidence~(RENT) and TTRL enhance reasoning without relying on external rewards~\citep{prabhudesai2025maximizing,zuo2025ttrl}; and even spurious or random reward signals can shift Qwen models~\citep{shao2025spurious}, implying that RL often surfaces pre-trained reasoning features rather than learning entirely new ones.
    A parallel line of work frames test-time search and compute as a meta-RL problem, proposing MRT to densify progress signals and yield better scaling of ``thinking time'' than outcome-only RL~\citep{qu2025optimizing}.
    Data-efficiency studies have also shown that even extreme cases such as \emph{1-shot} RLVR can substantially improve mathematical reasoning, again aligning with the sharpening view of eliciting latent capabilities~\citep{wang2025reinforcement}.
    Complementing these perspectives, a systematic study of exploration in RLVR~\citep{deng2025trial} formalizes \passk~as a measure of exploration boundaries and uncovers nuanced entropy–performance trade-offs across training, instance, and token levels, thereby situating the sharpening view within a unified analytic framework.
    Recently, \cite{shenfeld2025rl} introduce the principle of ``RL's Razor,'' demonstrating that online RL preserves prior knowledge significantly better than supervised fine-tuning. They show that RL's advantages stem from its ability to maintain existing capabilities while adapting to new tasks, rather than discovering entirely novel behaviors.

    \item Recently, however, several works have reopened the case for discovery. ProRL~\citep{liu2025prorl} reports that sufficiently prolonged and stabilized RL can extend a base model’s reasoning frontier, improving both \passo~and \passk.
    Continued scaling evidence is provided by ProRL~v2~\citep{liu2025prorl}, which incorporates engineering advances and demonstrates stronger results.
    Meanwhile, critiques of \passk~metrics have led to alternatives such as \emph{CoT-Pass@$k$}, supported by theoretical arguments that RLVR implicitly incentivizes correct reasoning paths rather than merely rewarding lucky endpoints~\citep{wen2025reinforcement}.
    Complementary approaches sustain RLVR's benefits by employing self-play problem synthesis to preserve entropy and enhance \passk~\citep{liang2025beyond}, or by directly optimizing \passk~through novel policy objectives~\citep{walder2025pass,chen2025pass}.
    \cite{yuan2025llms} further provide compelling evidence for the discovery view by demonstrating that LLMs can learn new skills in RL through the composition of existing capabilities, suggesting that RL enables emergent behaviors beyond simple refinement of pre-existing patterns.
\end{itemize}

The apparent dichotomy between \textcolor{cyan!40!black}{Sharpening} and \textcolor{cyan!40!red}{Discovery} may be reconciled through recent theoretical advances that reveal deeper connections between different alignment paradigms. The work of \cite{xiao2025connection} shows that RLHF implicitly performs imitation learning, while \cite{sun2024supervised} demonstrates that SFT can be understood as inverse RL.
These insights suggest that both supervised and RL approaches are operating within a shared theoretical framework of distribution matching and reward optimization.
The key distinction lies not in whether these methods can discover new capabilities, but rather in how they navigate the trade-off between exploration and exploitation~\citep{schmied2025llms}. The mode-seeking property of reverse KL in RL provides a mechanism for efficient convergence to high-performance regions (\textcolor{cyan!40!black}{Sharpening}), while the implicit reward learning and sequential decision-making aspects enable the composition of existing capabilities into novel behaviors (\textcolor{cyan!40!red}{Discovery}) when given sufficient training time and appropriate regularization~\citep{yuan2025llms,liu2025prorl}. This unified perspective suggests that the debate should shift from ``\textcolor{cyan!40!black}{Sharpening} or \textcolor{cyan!40!red}{Discovery}'' to understanding the conditions under which each phenomenon dominates.

\subsection{RL vs. SFT: Generalize or Memorize}
\label{sec:problems_rl_vs_sft}

In this subsection, we discuss the roles of RL and supervised fine-tuning, focusing on the interplay between generalization and memorization.
There are two primary approaches to post-training LLMs: SFT and RL. Current debates focus on two main questions:
1) Which method better enables out-of-distribution generalization?
2) Does behavior cloning via SFT set an upper bound on generalization capabilities?
Recently, significant research attention has been devoted to this topic. Notably, \citet{chu2025sft} provide a direct conclusion across both textual and vision environments, stating that ``SFT memorizes, RL generalizes.''

Two recent studies sharpen this contrast.
\citet{huan2025does} find that RL on math tasks (RL-on-math) tends to preserve, or even enhance, performance on non-math tasks and instruction following, whereas supervised fine-tuning on math (SFT-on-math) often leads to negative transfer and catastrophic forgetting.
Their diagnostic analyses based on latent-space PCA and token-distribution (KL) measures, as well as those by~\citet{mukherjee2025reinforcement}, suggest that SFT induces representation and output drift (memorization), while RL better preserves the base-domain structure (generalization).
Complementarily, \citet{zhou2025does} dissect five math problem-solving training routes and observe that 1) continual pretraining on math text provides only modest transfer, 2) conventional \emph{short}-CoT SFT frequently harms generalization, yet 3) \emph{long}-CoT SFT and rule-based RL (with format/correctness rewards) expand reasoning depth and self-reflection and thus improve broader reasoning; moreover, an SFT warmup \emph{before} RL stabilizes the policy and further boosts cross-domain transfer.
These results suggest that on-policy objectives and longer, self-reflective traces foster transferable patterns that remain robust under distribution shift, whereas short-CoT SFT tends to overfit to surface patterns, mirroring the classic RL-vs.-SFT divide between generalization and memorization.
There are three main research directions on this topic:
\begin{itemize}
    \item \textbf{RL demonstrates superior generalization}:
    \citet{chu2025sft} show that RL outperforms SFT in terms of Out-of-Distribution (OOD) performance, while SFT tends to memorize data on the GeneralPoints and V-IRL tasks.
    Previous studies~\citep{kirk2023understanding} have also indicated that RLHF, particularly under greater distribution shifts, can generalize more effectively than SFT, though this may come at the cost of reduced output diversity.
    Additionally, DeepSeek-R1~\citep{guo2025deepseek} demonstrates that pure RL training can lead to the spontaneous emergence of advanced reasoning behaviors, such as reflection and verification.
    \item \textbf{RL is not a panacea}:
    The generalization ability of RL is strongly influenced by the initial data distribution and the design of verification rewards.
    \citet{jin2025rl} find that RL can partially mitigate overfitting; however, it remains ineffective in cases of severe overfitting or abrupt distributional shifts, as observed in OOD ``24 points'' and spectrum analysis tasks.
    The primary value of RL lies in its ability to facilitate ``proper learning''~\citep{swamy2025all}.
    SFT can significantly improve generalization when appropriate reweighting, trust-region constraints, or dynamic rescaling are applied, and it often better prepares models for subsequent RL~\citep{qin2025supervised}.
    In practice, SFT may serve as a lower bound for sparse reward RL.
    \item \textbf{Unified or alternating paradigms of SFT and RL}:
    \citet{yan2025learning} present a framework that enhances RLVR by incorporating off-policy reasoning traces.
    \citet{liu2025uft} integrates SFT and RL into a single-stage target, theoretically overcoming the bottleneck of long-horizon sample complexity and empirically demonstrating superiority over using either approach alone.
    \citet{fu2025srft} propose a joint single-stage integration of demonstration imitation (SFT) and strategy improvement (RL) using entropy perception weights.
    \citet{zhang2025bread} provide theoretical evidence that in scenarios involving small models, high difficulty, or sparse successful trajectories, the traditional from SFT to RL two-stage approach may fail entirely. They address this by employing a branch rollout mechanism that begins from expert anchors to effectively link the two stages.
    \citet{ma2025learning} find that RL excels at consolidating and enhancing existing abilities, whereas SFT is more effective at introducing new knowledge or novel model capabilities.
    \citet{matsutani2025rl} analyze how RL and SFT influence reasoning paths and graph topologies in LLMs, revealing complementary effects on reasoning functionality.
\end{itemize}

However, several challenges remain unresolved.
One major issue is distinguishing between genuine problem-solving ability and mere memorization of answers, while simultaneously avoiding data contamination~\citep{satvaty2024undesirable}.
There is still a lack of standardized, reproducible out-of-distribution benchmarks. Additionally, RL training is highly sensitive to the initial data distribution; when SFT induces significant representation drift, the ability of RL to recover and generalize is limited~\citep{jin2025rl}.
To address these challenges, there is a need to promote frameworks such as UFT~\citep{liu2025uft}, SRFT~\citep{fu2025srft}, and Interleaved~\citep{ma2025learning}, which mechanize the integration of SFT for incorporating new knowledge with RL for amplification and robustness.
\citet{lv2025towards} also explore automated scheduling strategies to determine when to switch between SFT and RL and how to allocate their proportions effectively.

In conclusion, RL tends to achieve ``true generalization'' on verifiable tasks and under substantial distribution shifts, but it is not a panacea. Modified SFT can help bridge the remaining gaps in generalization. Consequently, best practices are converging towards unified or alternating hybrid paradigms that combine the strengths of both approaches~\citep{chen2025bridging,liu2025uft,zhu2025proximal,wu2025generalization,lv2025towards,chen2025twostagetrainingcooperativesft}.

\subsection{Model Prior: Weak and Strong}
\label{sec:problems_model_prior}

Recent studies have shown that RL can now perform well across a wide range of tasks when coupled with sufficiently powerful model priors and verifiable reward signals, thereby shifting the primary bottleneck from scale to the design of environments and evaluation protocols\footnote{\url{https://ysymyth.github.io/The-Second-Half/}}.
From this perspective, RL serves chiefly to resharpen latent competencies already encoded during pretraining, rather than to generate novel abilities entirely from scratch.

In this subsection, we examine three key dimensions of this dependency: the comparative advantages of applying RL to base versus instruction-tuned models, the substantial variations in RL responsiveness across different model families (particularly between Qwen and Llama architectures), and the emerging strategies that can enhance RL outcomes for both weak-prior and strong-prior models, including mid-training and curriculum design.

\paragraph{Base vs. Instruct Models.}
DeepSeek-R1 first introduced a discussion on applying RL to either base models or instruct-tuned models, and it introduced two viable paradigms for post-training:
1) \emph{R1-Zero}, which applies large-scale rule-based RL directly to a base model, yielding emergent long-horizon reasoning;
and 2) \emph{R1}, which incorporates a brief cold-start SFT stage to stabilize output format and readability prior to RL.
Independently, Open-Reasoner-Zero \citep{hu2025open} demonstrated that a minimalist training recipe applied to base Qwen models is sufficient to scale both response length and benchmark accuracy, mirroring the training dynamics of R1-Zero.
These findings suggest that base model priors are better suited to RL than those of instruct models, often producing smoother improvement trajectories than those observed when starting from heavily aligned Instruct models, where entrenched formatting and obedience priors may interfere with reward shaping.

\paragraph{Model Family Differences.}
More recent studies highlight that the choice of base model can critically shape RL outcomes.
For instance, One-shot RLVR~\citep{wang2025reinforcement} shows that introducing a single, carefully selected mathematical example can more than double MATH500 accuracy for Qwen2.5-Math-1.5B, delivering substantial average improvements across multiple benchmarks.
Yet, Spurious Rewards~\citep{shao2025spurious} uncovers a contrasting pattern: Qwen-family models register significant gains even under random or spurious reward signals, whereas Llama and OLMo models often do not.
This divergence underscores the influence of model priors and emphasizes the importance of validating RL claims across models with differing priors.
The observed asymmetries suggest differences in pretraining exposure to reasoning patterns (e.g., mathematical or code CoT). Qwen models, having been extensively exposed to such distributions, tend to be more ``RL-friendly'', whereas comparable Llama models often exhibit brittleness when subjected to the same RLVR procedure.

\paragraph{Mid-training Solutions.}
In practice, researchers have found that this performance gap can be addressed through mid-training or annealing training strategies.
In recent LLM research, annealing denotes a late-stage pre-training phase during which the learning rate decays while the data distribution is reweighted to emphasize smaller, high-quality sources such as code, mathematics, and curated QA corpora.
Llama 3~\citep{grattafiori2024llama} explicitly names this phase Annealing Data, describing both a shift in the data mixture and a linear LR decay to zero. They further report that injecting small amounts of high-quality math and code at this stage substantially improves reasoning-oriented benchmarks.
Earlier, MiniCPM~\citep{hu2024minicpm} articulated a comparable two-stage curriculum, termed stable-then-decay. During the decay (annealing) stage, they interleave SFT-style, high-quality knowledge and skill data with standard pre-training corpora, observing larger improvements than applying the same SFT only after pre-training.
Similarly, OLMo 2~\citep{olmo20242} makes public a modern mid-training recipe: pre-training is split into a long, web-heavy stage followed by a shorter mid-training phase that up-samples high-quality and domain-specific sources, especially mathematics, while linearly decaying the LR to zero.
More generally, contemporary mid-training strategies treat the joint design of learning rate schedules and data distribution switches as a first-class concern. For instance, \citet{parmar2024reuse} show that optimal continued-pretraining requires: 1) a two-distribution curriculum that emphasizes the target capabilities during the late stage, and 2 an annealed, non-rewarmed LR schedule where the timing of the distribution switch is determined by the LR fraction rather than a fixed token count.
A recent systematic study extends this line of work, demonstrating that a stable-then-decay mid-training curriculum that injects high-quality mathematics and chain-of-thought QA corpora makes Llama models substantially more scalable under RL-based fine-tuning, effectively narrowing the performance gap with Qwen models~\citep{wang2025octothinker}.
Taken together, these findings suggest a practical recipe for weak-prior model families: strengthen reasoning priors through mid-training, and subsequently apply RLVR.

\paragraph{Strong Model Improvements.}
While many replications favor base models, there is mounting evidence that RL can further improve strong distilled/Instruct models when curriculum, verification, and length control are carefully designed.
For example, AceReason-Nemotron~\citep{chen2025acereason} reports consistent gains from math-first then code-only RL atop distilled Qwen models, with analyses showing improvements in both \passo~and \passk~regimes. These findings nuance a simplistic ``base-only'' narrative: with the right constraints, Instruct/distilled starts can also benefit, but optimization is less forgiving.
A parallel line evaluates the controllability of reasoning models. MathIF~\citep{fu2025scaling} highlights a systematic tension: scaling up reasoning capabilities frequently undermines instruction-following performance, particularly in the context of long-form outputs.
Complementary evidence shows that explicit CoT prompting can reduce instruction-following accuracy and proposes selective-reasoning mitigations~\citep{li2025thinking}. Together, these works motivate multi-objective training (format, brevity, obedience) alongside correctness/verifiability in RL.

We can summarize how model priors fundamentally shape RL outcomes in LLM training from three perspectives:
1) Base models consistently outperform instruct-tuned models as RL starting points, with DeepSeek-R1 and Open-Reasoner-Zero demonstrating emergent reasoning from minimal recipes;
2) Model families exhibit asymmetric RL responsiveness: Qwen models show gains even under spurious rewards while Llama/OLMo models require careful mid-training with annealed learning rates and high-quality math/code data injection;
3) Strong distilled models can benefit from RL but demand more sophisticated curriculum design and multi-objective optimization.

As RL increasingly serves to resharpen latent pretraining competencies rather than create novel abilities, the focus shifts toward optimizing the pretraining-to-RL pipeline holistically rather than treating these stages independently.

\subsection{Training Recipes: Tricks or Traps}
\label{sec:problems_training_recipes}

RL training for large models has primarily evolved from the PPO~\citep{schulman2017proximal} series, maintaining stability through a variety of engineering techniques~\citep{shengyi2022the37implementation} such as trimming, baseline correction, normalization, and KL regularization. In the context of RL for LLM reasoning, DeepSeek-Math and DeepSeek-R1 introduce critic-free GRPO~\citep{shao2024deepseekmath}, which simplifies the training process by reducing complexity. Despite these advances, challenges related to training stability and efficiency persist, motivating a range of new methods, including dynamic sampling, various importance sampling ratios, and multi-level normalization.

A more widely adopted technique to boost exploration is to use decoupled PPO clipping (``Clip-Higher''), where the upper clipping bound is set higher than the lower one (e.g., $\epsilon_{\text{low}}=0.2, \epsilon_{\text{high}}=0.28$) to allow the probabilities of unlikely but potentially useful tokens to increase more freely \citep{yu2025dapo, liu2025prorl, an2025polaris}. Archer~\citep{wang2025stabilizing} proposes a dual-clipping mechanism for tokens with different entropy levels and ASPO~\citep{ASPO} further uses asymmetric importance sampling for tokens with opposite advantage values.

\begin{itemize}
    \item \textbf{Minimalism in Data and Sampling}: \citet{xiong2025minimalist} decompose GRPO and finds that the largest performance gains come from discarding all incorrect samples, rather than relying on complex reward normalization techniques. They propose that methods like RAFT~\citep{dong2023raft} or ``Reinforce-Rej''~\citep{liu2023statistical} can achieve stability and KL efficiency comparable to GRPO/PPO using much simpler mechanisms. DAPO~\citep{yu2025dapo} systematizes ``dynamic sampling + decoupled pruning'' into a reproducible large-scale approach, and incorporates decoupled PPO clipping (``Clip-Higher'') where the upper clipping bound is set higher than the lower one (e.g., $\epsilon_{\text{low}}=0.2, \epsilon_{\text{high}}=0.28$) to allow the probabilities of unlikely but potentially useful tokens to increase more freely, demonstrating state-of-the-art results on strong baselines for the AIME24 benchmark. Similarly, GRESO~\citep{zheng2025act} shows that pre-filtering can speed up rollout time by 2.4× and overall training by 2.0× with minimal loss in performance.
    \item \textbf{Structural Modification of the Objective Function}: GSPO~\citep{zheng2025group} shifts ratio and cropping operations to the sequence level, resulting in improved stability and efficiency over GRPO, especially for stable RL training of Mixture-of-Experts (MoE) models.
    S-GRPO~\citep{dai2025s} further reduces redundant reasoning, mitigating the tendency for longer and unnecessary reasoning chains and shortening sequence length by 35–61\% across multiple benchmarks, with slight improvements in accuracy.
    \item \textbf{The Struggle Between De-biasing and Normalization}: Dr. GRPO~\citep{liu2025understanding} identifies a key deviation in GRPO where ``the longer it's wrong, the more wrong it gets,'' and introduces minor algorithmic modifications to improve token efficiency. At the same time, other studies (e.g., BNPO~\citep{xiao2025bnpo}) revisit the importance of reward normalization from an adaptive distribution perspective, proposing new normalization families. The evidence from these two camps is contradictory, indicating that viewing normalization as a universal solution may be misleading.
\end{itemize}

\citet{liu2025part} present a recent review with unified evaluation, incorporating common techniques into a single open-source framework~\citep{wang2025reinforcement-roll} to enable isolated and reproducible experiments. This work provides a roadmap outlining ``which techniques are effective under what settings'' and demonstrates that a minimalist combination of methods can outperform GRPO and DAPO across multiple configurations. Crucially, it highlights the field's most pressing challenges: inconsistent experimental settings, incomplete reporting, and conflicting conclusions. This constitutes a fundamental limitation in the current application of RL within the research community.
In summary, while practical ``tricks'' are valuable for stabilizing RL training, the essence of ``scientific training'' lies in verification and scalability. Progress in the field requires unified experimental protocols, verifiable reward structures, and explicit scalability–performance–cost curves~\citep{nimmaturi2025predictive} to show that a method remains effective as it scales, rather than only at specific data or models.

\subsection{Reward Type: Process or Outcome}
\label{sec:problems_reward_type}

In standard RL, the objective of the policy is to maximize the expected cumulative reward~\citep{sutton1998introduction}. The ``Reward is Enough'' hypothesis~\citep{silver2021reward,bowling2023settling} further posits that appropriately designed rewards are sufficient and that maximizing returns can, in principle, give rise to all aspects of intelligence. In the context of RL for LLMs, the core challenge is how to provide meaningful rewards, such as training a reward model or verifier to score outputs and using these scores for RL or search. Common approaches include outcome rewards, which evaluate only the final result (e.g., correctness or passing individual tests), and process rewards, which provide step-by-step scoring through dense feedback on intermediate steps~\citep{lightman2023let}.

\begin{itemize}
    \item As shown in $\S$~\ref{sec:reward_verifiable}, when task answers are verifiable, outcome rewards are the simplest and most scalable for challenging mathematical and coding tasks.
    However, outcome-only approaches may tacitly encourage unfaithful chain-of-thought~\citep{arcuschin2025chain}, such as ``answer first, hallucinate later,'' and reward speculation. Recent research~\citep{baker2025monitoring} indicates that state-of-the-art models also exhibit unfaithful reasoning and post-hoc rationalization in real-world scenarios. Other work has highlighted that rule-based RL is prone to reward hacking and the development of reasoning illusions~\citep{sun2025detection}.
    \item PRMs~\citep{zhang2025openprm} naturally facilitate long-chain credit assignment. \citet{lightman2023let} clearly compare the two reward approaches: for mathematical reasoning, PRMs trained with process supervision are more stable and reliable, significantly outperforming those supervised solely by results. Nevertheless, step-wise annotation is extremely costly, and quality often declines across different domains~\citep{zhang2025lessons}. Relevant studies suggest that heuristic or Monte Carlo–based synthesis approaches tend to generalize poorly and introduce bias~\citep{yin2025dynamic}.
\end{itemize}

Overall, outcome rewards provide ``scalable goal alignment with automated verification'', while process rewards offer ``interpretable dense guidance.'' Combining the two, for example via implicit process modeling~\citep{cui2025process} or generative verifiers~\citep{zhang2024generative}, may represent a promising future direction in reward design.

\section{Training Resources}
\label{sec:resource}

Effective RL for LLMs depends not only on algorithms and objective design, but also on the quality and structure of the underlying training resources. The selection of resources ranging from static corpora to dynamic environments and specialized RL infrastructure, profoundly influences both the stability and scalability of large-scale training.
In this section, we survey the key categories of training resources leveraged in current practice. We first examine the role and limitations of static corpora as a foundation for RL ($\S$~\ref{sec:resource_static_corpus}), then discuss the growing importance of dynamic, interactive environments that provide richer learning signals and more realistic task distributions ($\S$~\ref{sec:resource_dynamic_environment}). Finally, we review the RL infrastructure that enables scalable and efficient training pipelines for LLMs ($\S$~\ref{sec:resource_rl_infra}).

\subsection{Static Corpus}
\label{sec:resource_static_corpus}
\begin{myboxi}[Takeaways]
\begin{itemize}
	\item RL reasoning datasets are moving from large-scale raw data to higher-quality, verifiable supervision using distillation, filtering, and automated evaluation to boost sample effectiveness and process fidelity.
	\item Data coverage has expanded beyond single domains (math/code/STEM) to include search, tool use, and agentic tasks with traceable, plan–act–verify trajectories.
\end{itemize}
\end{myboxi}

This section surveys static corpora for RL with LLMs. Data construction is shifting from ``scale-first'' to ``quality- and verifiability-first'', explicitly to support verifiable rewards (see \S~\ref{sec:reward_verifiable}).
As shown in Table~\ref{tab:dataset}, the dataset coverage spans four major tracks: mathematics, coding, STEM, and agentic tasks (e.g., search and tool use). All corpora are directly compatible with RLVR, enabling process-aware evaluation. These datasets support key components of the RL pipeline, including policy pretraining, reward modeling, and difficulty-aware sampling.

\afterpage{
\clearpage
\footnotesize
\begin{longtable}{llccccc}
\caption{Static datasets for RL training of LLMs, including Math, Code, STEM, and Agent domains. For data acquisition methods, ``Distil'' and ``Anno'' indicate distillation and annotation, respectively. ``Merge’’ indicates the integration of existing datasets, including difficulty and quality filtering.}
\label{tab:dataset}\\
\toprule
\textbf{Domain} & \textbf{Date} & \textbf{Name} & \textbf{\#Sample} & \textbf{Format} & \textbf{Type} & \textbf{Link} \\
\midrule
\endfirsthead
\multicolumn{7}{c}%
{\tablename\ \thetable\ -- \textit{Continued from previous page}} \\
\toprule
\textbf{Domain} & \textbf{Date} & \textbf{Name} & \textbf{\#Sample} & \textbf{Format} & \textbf{Type} & \textbf{Link} \\
\midrule
\endhead
\midrule \multicolumn{7}{r}{\textit{Continued on next page}} \\
\endfoot
\bottomrule
\endlastfoot
\multirow{15}{*}{Math}
& 2025.02 & DAPO             & 17k   & Q-A   & Anno          & \href{https://github.com/BytedTsinghua-SIA/DAPO}{\faGithub}\; \href{https://huggingface.co/datasets/BytedTsinghua-SIA/DAPO-Math-17k}{\faHuggingFace} \\
& 2025.02 & PRIME            & 481k  & Q-A   & Merge\&Distil & \href{https://github.com/PRIME-RL/PRIME}{\faGithub}\; \href{https://huggingface.co/datasets/PRIME-RL/Eurus-2-RL-Data}{\faHuggingFace} \\
& 2025.02 & Big-MATH         & 47k   & Q-A   & Anno          & \href{https://huggingface.co/datasets/SynthLabsAI/Big-Math-RL-Verified}{\faHuggingFace} \\
& 2025.02 & LIMO             & 800   & Q-C-A & Anno          & \href{https://github.com/GAIR-NLP/LIMO}{\faGithub}\; \href{https://huggingface.co/datasets/GAIR/LIMO-v2}{\faHuggingFace} \\
& 2025.02 & LIMR             & 1.39k & Q-A   & Anno          & \href{https://github.com/GAIR-NLP/LIMR}{\faGithub}\; \href{https://huggingface.co/datasets/GAIR/LIMR}{\faHuggingFace} \\
& 2025.02 & DeepScaleR       & 40.3k & Q-C-A & Distil        & \href{https://huggingface.co/datasets/agentica-org/DeepScaleR-Preview-Dataset}{\faHuggingFace} \\
& 2025.02 & NuminaMath 1.5       & 896k & Q-C-A & Anno   & \href{https://github.com/project-numina/aimo-progress-prize}{\faGithub}\; \href{https://huggingface.co/datasets/AI-MO/NuminaMath-1.5}{\faHuggingFace} \\
& 2025.02 & OpenReasoningZero     & 72k & Q-A & Merge\&Distil   & \href{https://github.com/Open-Reasoner-Zero/Open-Reasoner-Zero}{\faGithub}\; \href{https://huggingface.co/datasets/Open-Reasoner-Zero/orz_math_72k_collection_extended}{\faHuggingFace} \\
& 2025.02 & STILL-3-RL     & 90k & Q-A & Merge\&Distil   & \href{https://github.com/RUCAIBox/Slow_Thinking_with_LLMs}{\faGithub}\; \href{https://huggingface.co/datasets/RUC-AIBOX/STILL-3-RL-90K}{\faHuggingFace} \\
& 2025.02 & OpenR1-Math     & 220k & Q-C-A & Distil   & \href{https://github.com/huggingface/open-r1}{\faGithub}\; \href{https://huggingface.co/datasets/open-r1/OpenR1-Math-220k}{\faHuggingFace} \\
& 2025.03 & Light-R1         & 79.4k & Q-C-A & Merge               & \href{https://huggingface.co/datasets/qihoo360/Light-R1-SFTData}{\faHuggingFace} \\
& 2025.04 & DeepMath       & 103k & Q-C-A & Distil\&Anno   & \href{https://github.com/zwhe99/DeepMath}{\faGithub}\; \href{https://huggingface.co/datasets/zwhe99/DeepMath-103K}{\faHuggingFace} \\
& 2025.04 & OpenMathReasoning     & 5.5M & Q-C-A & Distil   & \href{https://github.com/NVIDIA/NeMo-Skills}{\faGithub}\; \href{https://huggingface.co/datasets/nvidia/OpenMathReasoning}{\faHuggingFace} \\
& 2025.07 & MiroMind-M1-RL-62K    & 62k & Q-A & Merge   & \href{https://github.com/MiroMindAI/MiroMind-M1}{\faGithub}\; \href{https://huggingface.co/datasets/miromind-ai/MiroMind-M1-RL-62K}{\faHuggingFace} \\
\midrule
\multirow{10}{*}{Code}
& 2024.12 & SWE-Gym  & 2.4k & Q-A & Anno     &  \href{https://github.com/SWE-Gym/SWE-Gym}{\faGithub}\; \href{https://huggingface.co/datasets/SWE-Gym/SWE-Gym}{\faHuggingFace} \\
& 2025.01 & codeforces-cots  & 47.8k & Q-C-A & Distil        & \href{https://huggingface.co/datasets/open-r1/codeforces-cots}{\faHuggingFace} \\
& 2025.01 & SWE-Fixer        & 110k  & Q-A   & Anno        & \href{https://github.com/InternLM/SWE-Fixer}{\faGithub}\; \href{https://huggingface.co/datasets/internlm/SWE-Fixer-Train-110K}{\faHuggingFace} \\
& 2025.03 & KodCode          & 268k  & Q-A   & Distil        & \href{https://kodcode-ai.github.io/}{\faGithub}\; \href{https://huggingface.co/datasets/KodCode/KodCode-V1-SFT-R1}{\faHuggingFace} \\
& 2025.03 & Code-R1          & 12k  & Q-A   & Merge        & \href{https://github.com/ganler/code-r1}{\faGithub}\; \href{https://huggingface.co/datasets/ganler/code-r1-12k}{\faHuggingFace} \\
& 2025.04 & Z1               & 107k  & Q-C-A & Distil        & \href{https://github.com/efficientscaling/Z1}{\faGithub}\; \href{https://huggingface.co/datasets/efficientscaling/Z1-Code-Reasoning-107K}{\faHuggingFace} \\
& 2025.04 & LeetCodeDataset               & 2.9k  & Q-A & Anno        & \href{https://github.com/newfacade/LeetCodeDataset}{\faGithub}\; \href{https://huggingface.co/datasets/newfacade/LeetCodeDataset}{\faHuggingFace} \\
& 2025.04 & OpenCodeReasoning& 735k  & Q-C-A & Distil        & \href{https://huggingface.co/datasets/nvidia/OpenCodeReasoning}{\faHuggingFace} \\
& 2025.04 & DeepCoder& 24k  & Q-A & Merge        &  \href{https://github.com/agentica-project/rllm}{\faGithub}\; \href{https://huggingface.co/datasets/agentica-org/DeepCoder-Preview-Dataset}{\faHuggingFace} \\
& 2025.05 & rStar-Coder      & 592k  & Q-C-A & Distil\&Anno & \href{https://github.com/microsoft/rStar}{\faGithub}\; \href{https://huggingface.co/datasets/microsoft/rStar-Coder}{\faHuggingFace} \\
\midrule
\multirow{6}{*}{STEM}
& 2025.01 & SCP-116K         & 182k  & Q-C-A & Distil        & \href{https://huggingface.co/datasets/EricLu/SCP-116K}{\faHuggingFace} \\
& 2025.02 & NaturalReasoning & 2.15M & Q-C-A & Distil        & \href{https://huggingface.co/datasets/facebook/natural_reasoning}{\faHuggingFace} \\
& 2025.05 & ChemCoTDataset   & 5k    & Q-C-A & Distil        & \href{https://huggingface.co/datasets/OpenMol/ChemCoTDataset}{\faHuggingFace} \\
& 2025.06 & ReasonMed        & 1.11M & Q-C-A & Distil        & \href{https://github.com/YuSun-Work/ReasonMed}{\faGithub}\; \href{https://huggingface.co/datasets/lingshu-medical-mllm/ReasonMed}{\faHuggingFace} \\
& 2025.07 & MegaScience      & 2.25M & Q-C-A & Merge\&Distil & \href{https://huggingface.co/datasets/MegaScience/MegaScience}{\faHuggingFace} \\
& 2025.09 & SSMR-Bench      & 16k & Q-A & Anno & \href{https://github.com/Linzwcs/AutoMusicTheoryQA}{\faGithub}\; \href{https://huggingface.co/datasets/Sylence/SSMR-Bench}{\faHuggingFace} \\
\midrule
\multirow{7}{*}{Agent}
& 2025.03 & Search-R1 & 221K & Q-A & Anno &  \href{https://huggingface.co/datasets/PeterJinGo/nq_hotpotqa_train}{\faHuggingFace}\\
& 2025.03 & ToRL & 28K & Q-A & Merge & \href{https://github.com/GAIR-NLP/ToRL}{\faGithub}\\
& 2025.03 & ToolRL & 4K & Q-C-A & Distil & \href{https://github.com/qiancheng0/ToolRL}{\faGithub}\\
& 2025.05 & ZeroSearch & 170K & Q-A & Anno & \href{https://github.com/Alibaba-NLP/ZeroSearch}{\faGithub}\; \href{https://huggingface.co/datasets/Alibaba-NLP/ZeroSearch_dataset}{\faHuggingFace} \\
& 2025.07 & WebShaper & 0.5K & Q-A & Anno & \href{https://huggingface.co/datasets/Alibaba-NLP/WebShaper}{\faHuggingFace} \\
& 2025.08 & MicroThinker & 67.2K & Q-A & Anno & \href{https://huggingface.co/datasets/miromind-ai/MiroVerse-v0.1}{\faHuggingFace} \\
& 2025.08 & ASearcher & 70K & Q-A & Anno & \href{https://huggingface.co/datasets/inclusionAI/ASearcher-train-data/viewer/default/ASearcherLRM35k}{\faHuggingFace} \\
\midrule
\multirow{6}{*}{Mix}
& 2025.01 & dolphin-r1     & 300k & Q-C-A & Distil   & \href{https://huggingface.co/datasets/QuixiAI/dolphin-r1}{\faHuggingFace} \\
& 2025.02 & SYNTHETIC-1/2    & 2M/156K & Q-C-A & Distil   & \href{https://huggingface.co/datasets/PrimeIntellect/SYNTHETIC-1}{\faHuggingFace}\; \href{https://huggingface.co/datasets/PrimeIntellect/SYNTHETIC-2-RL}{\faHuggingFace} \\
& 2025.04 & SkyWork OR1& 14k  & Q-A & Merge        &  \href{https://github.com/SkyworkAI/Skywork-OR1}{\faGithub}\; \href{https://huggingface.co/datasets/Skywork/Skywork-OR1-RL-Data/}{\faHuggingFace} \\
& 2025.05 & Llama-Nemotron-PT     & 30M & Q-C-A & Distil   & \href{https://huggingface.co/datasets/nvidia/Llama-Nemotron-Post-Training-Dataset}{\faHuggingFace} \\
& 2025.06 & AM-DS-R1-0528-Distilled    & 2.6M & Q-C-A & Distil &  \href{https://github.com/a-m-team/a-m-models}{\faGithub}\; \href{https://huggingface.co/datasets/a-m-team/AM-DeepSeek-R1-0528-Distilled}{\faHuggingFace} \\
& 2025.06 & guru-RL-92k      & 91.9k & Q-A   &    Distil     & \href{https://huggingface.co/datasets/LLM360/guru-RL-92k}{\faHuggingFace} \\
\end{longtable}
}

Math-focused RL datasets coalesce around three construction pipelines, including annotation/verification, distillation, and multi-source merging, while widely exposing intermediate reasoning traces and spanning sizes from hundreds to millions of examples. Compact, carefully curated sets such as LIMO~\citep{ye2025limo} and LIMR~\citep{li2025limr} emphasize high-quality problems with explicit process feedback; annotated/verified resources like DAPO~\citep{yu2025dapo}, Big-MATH~\citep{albalak2025big}, and DeepMath~\citep{he2025deepmath} deliver reliable solution trajectories suitable for reward modeling and value alignment; at larger scale, NuminaMath 1.5~\citep{li2024numinamath} extends process-rich samples; distillation-centric corpora including DeepScaleR~\citep{luo2025deepscaler}, OpenR1-Math~\citep{openr1}, and OpenMathReasoning~\citep{moshkov2025aimo} inherit strong-teacher or ``R1-style'' long-chain reasoning, supporting policy pretraining and RL-stage selection; merge-and-distill collections such as PRIME~\citep{cui2025process}, OpenReasoningZero~\citep{hu2025open}, and STILL-3-RL~\citep{chen2025empirical} integrate open problems with self-generated candidates, offering difficulty stratification and high-quality filtering signals; community-leaning releases like Light-R1~\citep{wen2025light} and MiroMind-M1-RL-62K~\citep{li2025miromind} package lightweight, RL-ready formats for rapid iteration under compute constraints. Collectively, these resources span basic computation to competition-level problems and provide both final answers and measurable intermediate steps, enabling scalable policy learning, reward modeling, and process-based reinforcement.

Code-oriented RL datasets primarily fall into three categories: program repair/editing, algorithmic competition problems, and general code synthesis with reasoning. These datasets typically provide executable unit tests and intermediate execution traces, facilitating reward shaping and process-level evaluation.
Interactive, test-driven resources such as SWE-Gym~\citep{pan2024training} target fine-grained editing policies; human-verified repair pairs like SWE-Fixer~\citep{xie2025swe} and LeetCodeDataset~\citep{xia2025leetcodedataset} support value alignment and reward modeling. For competition-style and algorithmic reasoning, codeforces-cots~\citep{penedo2025codeforces}, Z1~\citep{yu2025z1}, and OpenCodeReasoning~\citep{ahmad2025opencodereasoning} emphasize long-chain trajectories and difficulty stratification. In large-scale, ``R1-style'' distillation for general code generation, KodCode~\citep{xu2025kodcode} and rStar-Coder~\citep{liu2025rstar} provide process-rich samples that aid policy pretraining and RL-stage selection. Lightweight, merge-centric releases such as Code-R1~\citep{code-r1} and DeepCoder~\citep{luo2025deepcoder} are convenient for rapid iteration under compute constraints. Collectively, these corpora span single-function repair through competition-level problem solving, offering both automatically checkable end artifacts and stepwise plans/edits, thereby enabling scalable policy learning, reward modeling, and process-based reinforcement for code agents.

STEM-oriented RL datasets generally converge on three themes: textbook or curriculum extraction, cross-disciplinary large-scale reasoning, and domain-specialized corpora (e.g., chemistry and medicine) featuring merge-and-distill pipelines. These datasets commonly release chain-of-thought rationales and evidence-aligned signals, enabling process-level rewards.
SCP-116K~\citep{lu2025scp} targets undergraduate-to-doctoral science with automatically extracted problem–solution pairs plus model-generated reasoning. NaturalReasoning~\citep{yuan2025naturalreasoning} offers multi-discipline questions decontaminated from popular benchmarks with extracted reference answers. ChemCoTDataset~\citep{li2025beyond} contributes chemistry-specific CoT exemplars spanning molecular editing/optimization and reaction prediction. ReasonMed~\citep{sun2025reasonmed} provides multi-agent–distilled medical QA with multi-step CoT rationales and concise summaries. SSMR-Bench~\citep{wang2025synthesizing} programmatically synthesizes music-theory-grounded sheet-music reasoning questions in both textual (ABC notation) and visual formats, releasing 16k training pairs per modality, and supporting evaluation as well as RL with verifiable rewards. MegaScience~\citep{fan2025megascience} aggregates public scientific corpora via ablation-based selection and annotates step-by-step solutions for most constituent sets, forming a large training pool for RL on scientific reasoning.

Mixed-domain RL datasets unify math, code, and scientific reasoning through distillation-first and merge-centric pipelines, while broadly releasing chain-of-thought traces, verifier signals, and multi-trajectory candidates that enable process rewards and difficulty-aware selection. In R1-style mixtures, dolphin-r1~\citep{QuixiAI_DolphinR1_2025} blends DeepSeek-R1, Gemini-thinking, and curated chat data for general reasoning. The SYNTHETIC suite couples large-scale SFT-style traces with RL-ready multi-trace samples: SYNTHETIC-1~\citep{2025synthetic1} aggregates DeepSeek-R1 reasoning with diverse verifiers, and SYNTHETIC-2-RL~\citep{2025synthetic1} provides multi-domain tasks with multiple trajectories for preference/reward learning. SkyWork OR1-RL-Data~\citep{he2025skywork} emphasizes verifiable math and code problems with difficulty labels, serving as a lightweight RL pool. Llama-Nemotron Post-Training~\citep{bercovich2025llama-nemotron} compiles instruction/R1-style data spanning math, code, STEM, general reasoning, and tool use for post-training. AM-DeepSeek-R1-0528-Distilled~\citep{AM-DeepSeek-R1-0528-Distilled} offers cross-domain distilled traces with documented quality filtering, and guru-RL-92k~\citep{cheng2025revisiting} curates six high-intensity reasoning domains via a five-stage pipeline optimized for RL formats. Collectively, these corpora provide verifiable endpoints and stepwise rationales across domains, supporting scalable policy learning, reward modeling, and process-based reinforcement.

Agent-centric RL datasets concentrate on two complementary capabilities, search-as-action and tool use, while releasing verifiable process signals such as search/browse traces, evidence URLs, and tool-execution logs that enable process rewards and offline evaluation. Search-R1~\citep{jin2025search} builds on NQ/HotpotQA to train interleaved reasoning–search behavior. ToRL~\citep{li2025torl} scales tool-integrated RL from base models to learn when and how to invoke computational tools. ToolRL~\citep{qian2025toolrl} studies fine-grained reward design for learning tool selection and application. ZeroSearch~\citep{sun2025zerosearch} formulates offline information-seeking tasks that incentivize search without real web calls. WebShaper~\citep{tao2025webshaper} synthesizes information-seeking data via an ``Expander Agent'', covering diverse task forms and reasoning structures with URL evidence. MicroThinker~\citep{miromind2024opendata} contributes full rollout trajectories and rich tool-use logs for multi-step agents. ASearcher~\citep{gao2025beyond} releases Apache-2.0-licensed training splits for long-horizon search agents with question/answer fields and source annotations. Collectively, these corpora span planning, retrieval, tool orchestration, evidence verification, and answer generation, supporting scalable policy learning, reward modeling, and process-based reinforcement for web/search and tool-using agents.

\subsection{Dynamic Environment}
\label{sec:resource_dynamic_environment}

\begin{myboxi}[Takeaways]
    \begin{itemize}
        \item Static RL training datasets are increasingly insufficient for advanced and generalizable reasoning abilities.
        \item Scalable RL for LLMs needs to turn to synthesized or generated data and interactive environments, such as various gyms and world models.
    \end{itemize}
\end{myboxi}

\begin{table}[!t]
\centering
\caption{Dynamic RL Environments for RL Training of LLMs. Data source legend: \textbf{RD} = Read Data, \textbf{RS} = Rule-based Synthesis, \textbf{MS} = Model-based Synthesis. Scale legend: Training/Test set.}
\label{tab:Envs}
\resizebox{\textwidth}{!}{
\begin{tabular}{llcccccc}
\toprule
\textbf{Category} & \textbf{Date} & \textbf{Name} & \textbf{Data Source} & \textbf{Interactive} & \textbf{Scale} & \textbf{Multimodal} & \textbf{Link} \\
\midrule
\multirow{6}{*}{Rule-based} 
 & 2025.02 & AutoLogi & \textbf{RD} + \textbf{MS} & \xmark & $2458/6739$ puzzles & \xmark & \href{https://github.com/8188zq/AutoLogi}{\faGithub}\\
 & 2025.02 & Logic-RL & \textbf{RS} & \xmark & $5k$ samples & \xmark & \href{https://github.com/Unakar/Logic-RL}{\faGithub}\\
 & 2025.05 & Reasoning Gym & \textbf{RS} & \xmark & $104$ tasks & \xmark & \href{https://github.com/open-thought/reasoning-gym}{\faGithub}\\
 & 2025.05 & SynLogic & \textbf{RS} & \xmark & $35$ tasks & \xmark & \href{https://github.com/MiniMax-AI/SynLogic}{\faGithub} \href{https://huggingface.co/datasets/MiniMaxAI/SynLogic}{\faHuggingFace}\\
 & 2025.06 & ProtoReasoning & \textbf{RD} + \textbf{MS} & \xmark & $6620$ samples & \xmark & -\\
 & 2025.06 & Enigmata & \textbf{RD} + \textbf{RS} & \xmark & $36$ tasks & \xmark & \href{https://github.com/BytedTsinghua-SIA/Enigmata}{\faGithub} \href{https://huggingface.co/datasets/BytedTsinghua-SIA/Enigmata-Data}{\faHuggingFace}\\
\midrule
\multirow{9}{*}{Code-based} & 2024.07 & AppWorld & \textbf{RD} + \textbf{RS} & \cmark & $750$ tasks & \xmark & \href{https://github.com/StonyBrookNLP/appworld}{\faGithub}\\
& 2025.02 & AgentCPM-GUI & \textbf{RD} + \textbf{RS} & \cmark & $55k$ trajectories & \cmark & \href{https://github.com/OpenBMB/AgentCPM-GUI}{\faGithub} \href{https://huggingface.co/openbmb/AgentCPM-GUI}{\faHuggingFace}\\
& 2025.02 & MLGym & \textbf{RD} + \textbf{RS} & \cmark & $13$ tasks & \xmark & \href{https://github.com/facebookresearch/MLGym}{\faGithub}\\
& 2025.03 & ReCall & \textbf{RD} + \textbf{MS} & \cmark & $10010$ samples & \xmark & \href{https://github.com/Agent-RL/ReCall}{\faGithub} \href{https://huggingface.co/collections/agentrl/research-67e506a0311bea06dc54878b}{\faHuggingFace}\\
 & 2025.04 & R2E-Gym & \textbf{RD} + \textbf{MS} & \cmark & $8135$ cases & \xmark & \href{https://github.com/R2E-Gym/R2E-Gym}{\faGithub} \href{https://huggingface.co/R2E-Gym}{\faHuggingFace}\\
 & 2025.05 & MLE-Dojo & \textbf{RD} + \textbf{RS} & \cmark & $202$ tasks & \cmark & \href{https://github.com/MLE-Dojo/MLE-Dojo}{\faGithub} \href{https://huggingface.co/spaces/MLE-Dojo/Leaderboard}{\faHuggingFace}\\
 & 2025.05 & SWE-rebench & \textbf{RD} + \textbf{MS} & \cmark & $21336$ cases & \xmark & \href{https://swe-rebench.com/}{\faGithub} \href{https://huggingface.co/datasets/nebius/SWE-rebench-leaderboard}{\faHuggingFace}\\
 & 2025.05 & ZeroGUI & \textbf{MS} & \cmark & - & \cmark & \href{https://github.com/OpenGVLab/ZeroGUI}{\faGithub} \href{https://huggingface.co/collections/OpenGVLab/zerogui-68388cb7dbf608133c4b5fb2}{\faHuggingFace}\\
  & 2025.06 & MedAgentGym & \textbf{RD}& \cmark & $72,413$ cases & \xmark & \href{https://github.com/wshi83/MedAgentGym}{\faGithub} \href{https://huggingface.co/MedAgentGym}{\faHuggingFace}\\
\midrule
\multirow{10}{*}{Game-based} & 2020.10 & ALFWorld & \textbf{RS} & \cmark & $6$ tasks & \cmark & \href{https://github.com/alfworld/alfworld}{\faGithub}\\
& 2022.03 & ScienceWorld & \textbf{RS} & \cmark & $30$ tasks & \xmark & \href{https://github.com/allenai/ScienceWorld}{\faGithub} \href{https://huggingface.co/spaces/MarcCote/ScienceWorld}{\faHuggingFace}\\
& 2025.04 & Cross-env-coop & \textbf{RS} & \cmark & $1.16e17$ cases & \xmark & \href{https://github.com/KJha02/crossEnvCooperation}{\faGithub}\\
 & 2025.05 & lmgame-BENCH & \textbf{RD} + \textbf{RS} & \cmark & $6$ games & \cmark & \href{https://github.com/lmgame-org/GamingAgent/tree/main/lmgame-bench}{\faGithub} \href{https://huggingface.co/spaces/lmgame/lmgame_bench}{\faHuggingFace}\\
 & 2025.05 & G1(VLM-Gym) & \textbf{RD} + \textbf{RS} & \cmark & $4$ games & \cmark & \href{https://github.com/chenllliang/G1}{\faGithub}\\
 & 2025.06 & Code2Logic (GameQA) & \textbf{RD} + \textbf{MS} & \xmark & $140k$ QA & \cmark & \href{https://github.com/tongjingqi/code2logic}{\faGithub} \href{https://huggingface.co/Code2Logic}{\faHuggingFace}\\
 & 2025.06 & Play to Generalize & \textbf{RS} & \cmark & $36k$ samples $\times$ $2$ games & \cmark & \href{https://github.com/yunfeixie233/ViGaL}{\faGithub} \href{https://huggingface.co/yunfeixie/ViGaL-7B}{\faHuggingFace}\\
 & 2025.06 & KORGym & \textbf{RS} & \cmark & $51$ games & \cmark & \href{https://github.com/multimodal-art-projection/KORGym}{\faGithub} \href{https://huggingface.co/multimodal-art-projection/KORGym}{\faHuggingFace}\\
 & 2025.06 & Optimus-3 & \textbf{RS} & \cmark & $6$ tasks & \cmark & \href{https://github.com/JiuTian-VL/Optimus-3}{\faGithub} \href{https://huggingface.co/MinecraftOptimus}{\faHuggingFace}\\
 & 2025.08 & PuzzleJAX & \textbf{RS} & \cmark & $\sim900$ games & \cmark & \href{https://github.com/smearle/script-doctor}{\faGithub}\\
 
\midrule
\multirow{6}{*}{Model-based} & 2025.03 & Sweet-RL & \textbf{RD} + \textbf{MS} & \cmark & $10k/1k$ tasks & \xmark & \href{https://github.com/facebookresearch/sweet\_rl}{\faGithub} \href{https://huggingface.co/datasets/facebook/collaborative_agent_bench}{\faHuggingFace}\\
 & 2025.04 & TextArena & \textbf{RS} & \cmark & $99$ games & \xmark & \href{https://github.com/LeonGuertler/TextArena}{\faGithub}\\
 & 2025.05 & Absolute Zero & \textbf{MS} & \cmark & - & \xmark & \href{https://github.com/LeapLabTHU/Absolute-Zero-Reasoner}{\faGithub} \href{https://huggingface.co/collections/andrewzh/absolute-zero-reasoner-68139b2bca82afb00bc69e5b}{\faHuggingFace}\\
 & 2025.06 & SwS & \textbf{RD} + \textbf{MS} & \xmark & $40k$ samples & \xmark & \href{https://github.com/MasterVito/SwS}{\faGithub} \href{https://huggingface.co/datasets/MasterVito/SwS-Demo-Dataset}{\faHuggingFace}\\
 & 2025.07 & SPIRAL & \textbf{RS} & \cmark & $3$ games & \xmark & \href{https://github.com/spiral-rl/spiral}{\faGithub} \href{https://huggingface.co/collections/spiral-rl/spiral-68627f14c250c3cc1fdbf6fe}{\faHuggingFace}\\
 & 2025.08 & Genie 3 & \textbf{MS} & \cmark & - & \cmark & \href{https://deepmind.google/discover/blog/genie-3-a-new-frontier-for-world-models/}{\faGithub}\\
\midrule
\multirow{2}{*}{Ensemble-based} & 2025.06 & InternBootcamp & \textbf{RD} + \textbf{RS} & \cmark & $1060$ tasks & \xmark & \href{https://github.com/InternLM/InternBootcamp}{\faGithub}\\
& 2025.07 & Synthetic-2 & \textbf{RD} + \textbf{MS} & \cmark & $19$ tasks & \xmark & \href{https://huggingface.co/collections/PrimeIntellect/synthetic-2-68544d1853830a653bdc5712}{\faHuggingFace}\\

\bottomrule
\end{tabular}
}
\end{table}

Existing static RL corpora, whether manually annotated, semi-automatically labeled, or scraped from the Web, are increasingly insufficient for training models that require more advanced and generalizable reasoning abilities.
A growing number of works are now leveraging ``Dynamic Environments'' to jointly ensure both \textit{scalability} and \textit{verifiability}, two essential properties for effective model training \citep{wei2023asymmetry}.

Unlike traditional reasoning corpora, these dynamic environments represent a paradigm shift. They enable either the automated and limitless synthesis of data, or provide step-level, multi-turn feedback on a model's entire reasoning process. As shown in Table \ref{tab:Envs}, based on the methods used for synthesis and interaction, these environments can be categorized, serving as the interaction objects for the RL process. Given our focus on resources for training, this subsection's organization of datasets and environments will exclude benchmarks intended solely for evaluation.

\paragraph{Rule-based Environment.}
Relying solely on feedback like ``Exact Match'' can lead models to shortcut to memorization rather than actual reasoning. To counteract this, some environments offer complex and diverse tasks that require deterministic rule-based operations as a verifier.
AutoLogi \citep{zhu2025autologi} generates open-ended logic puzzles with controllable difficulty by building code that checks the correctness of logical constraints based on a fixed model output format.
Logic-RL \citep{xie2025logic} uses a scalable Knights and Knaves puzzle to create a rule-based RL environment, which generalized the reasoning capabilities of a 7B model to the mathematical domain.
Projects like SynLogic \citep{liu2025synlogic}, Reasoning Gym \citep{stojanovski2025reasoning}, and Enigmata \citep{chen2025enigmata} expand the task diversity further. They identify the key parameters that control the difficulty for each task, allowing for the unlimited generation of data across various logic-related reasoning challenges. In contrast, ProtoReasoning \citep{he2025protoreasoning} operates on the hypothesis that a model's generalization ability comes from shared abstract reasoning prototypes. It normalizes different task types into a consistent format, like Prolog questions or PDDL tasks, and then automatically verifies the model's output using an interpreter.

\paragraph{Code-based Environment.}
An important application area for LLM reasoning is software engineering and code development. A key characteristic of these environments is that models must interact with a compilable code environment during training. Therefore, how to scalably construct code-based task environments remains a significant research direction.
To teach agents to use tools, ReCall \citep{chen2025learning} leverages advanced LLMs to construct a Python-based tool interaction environment, autonomously synthesizing its own SynTool data for RL training. 
In the field of AutoML, MLGym \citep{nathani2025mlgym} was among the first to support an interactive environment for iterative experimentation and training. It isolates each task's execution environment using Docker containers. Though its tasks are largely fixed, it offers less scalability. MLE-Dojo \citep{qiang2025mle} offers more scalability as it is easier for users to integrate new tasks. In a similar vein, MedAgentGym \citep{xu2025medagentgym} is an efficient and scalable interactive training environment for the medical domain. In software engineering, R2E-Gym \citep{jain2025r2e} reduces the reliance on manually authored GitHub issues and test cases by programmatically generating environments directly from GitHub commit histories, integrating with OpenHands for interactive capabilities. Similarly, SWE-rebench \citep{badertdinov2025swe} extends the original static SWE-bench by proposing a scalable pipeline for constructing software engineering tasks. This pipeline includes complex, interactive tasks that simulate real-world software development scenarios, ensuring data freshness and avoiding data contamination. In the field of computer use, AgentCPM-GUI \citep{zhang2025agentcpm} constructs an interactive GUI environment during the RFT phase to provide feedback on the model's actions. Similarly, AppWorld \citep{trivedi2024appworld} uses an environment comprising various mobile application APIs. ZeroGUI \citep{yang2025zerogui} takes this a step further by using existing advanced VLMs to construct tasks for both Ubuntu and Android. During training, a GUI agent interacts with the environment, and the feedback is then provided to the VLM to give rewards, all without the need for manual data curation.

\paragraph{Game-based Environment.} 
Game environments are characterized by their clear and complex state spaces, where an AI's behavior is tightly coupled with the environment's state.This leads to a more multi-step and continuous interaction process compared to the environments mentioned previously, and such environments naturally support dense rewards in $\S$~\ref{sec:reward_dense}, making RL training more efficient and stable.
Early works on interactive environments for training agents, such as ALFWorld \citep{shridhar2020alfworld} and ScienceWorld \citep{wang2022scienceworld}, remain influential in the agent planning field. 
Code2Logic \citep{tong2025code2logic} utilized game code and Q\&A templates to automatically generate multimodal reasoning data, resulting in the GameQA dataset. This dataset is not only scalable but also tests a model's multimodal reasoning capabilities with graduated difficulty. lmgame-Bench \citep{hu2025lmgame}, in a different approach, directly selects classic games and interacts with an LLM via a unified API. The game environment updates its state and provides a reward based on the LLM's action, which the LLM then uses to adjust its strategy. Similarly, Play to Generalize \citep{xie2025play} used a simple, scalable game environment for RL to train a 7B-parameter MLLM. The research found that the reasoning skills acquired by the model could generalize to unseen games and multidisciplinary reasoning tasks. The work G1 \citep{chen2025g1} introduced the VLM-Gym, an RL environment that supports the parallel execution of multiple game states, facilitating large-scale training. KORGym \citep{shi2025korgym} further expands the number of supported simple games, offering interactive and difficulty-configurable RL environments. PuzzleJAX \citep{earle2025puzzlejaxbenchmarkreasoninglearning} takes a different approach by accelerating games generated from the PuzzleScript language using JAX. This not only speeds up the game environment to support RL-based training but also provides access to a community of game developers with a source of unlimited games. To learn general cooperative skills, Cross-environment Cooperation \citep{jha2025cross} leverages the game Overcooked and maximizes environmental diversity within a self-play framework. For more complex, high-degree-of-freedom games like Minecraft, the Optimus series of work \citep{li2025optimus} leverages knowledge graphs to interact with the game environment, constructing data to evaluate a model's long-term planning ability.

\paragraph{Model-based Environment.}
This paradigm facilitates the creation of highly flexible and diverse RL environments through model-to-model interaction or self-play. SwS \citep{liang2025sws} utilizes a model's failed training cases to abstract key concepts and generate new problems, thus enhancing its reasoning abilities in a targeted manner. SPIRAL \citep{liu2025spiral} uses three zero-sum games for self-play to prevent overfitting to a static policy. For model-to-model interaction, Sweet-RL \citep{zhou2025sweet} uses a prover-verifier-like training framework, where an agent interacts and collaborates with an LLM-based human simulator to solve front-end design and back-end programming tasks. TextArena \citep{guertler2025textarena} proposes using adversarial text games combined with a ranking system, which overcomes the bottleneck of human scoring by allowing models to interact directly to relatively measure their abilities. Absolute Zero \citep{zhao2025absolute} goes a step further by completely moving away from human-defined evaluation tasks, utilizing three reasoning modes for a model to autonomously generate its own tasks and improve its reasoning capabilities through self-evolution. In the visual domain, Genie 3 \citep{genie3} generates near-realistic and interactive 3D virtual environments, laying the foundation for future multimodal environment-interactive RL. While some existing world models have already enabled RL-based model training \citep{russell2025gaia,hafner2023mastering,dedieu2025improving}, and we have listed works that train LRMs using model-based environments above, there is still no sufficiently scalable solution to support RL training of LRMs based on world models. The ultimate form of such dynamic environments, we posit, would be an oracle world model capable of simulating a complete, self-contained world.

\paragraph{Ensemble-based Environment.}
There are also works that involve significant engineering effort that integrate various tasks and datasets to form interactive environments and training data for RL. InternBootcamp \citep{li2025internbootcamp} is a large-scale, extensible library of environments designed to train LRMs. It supports over 1000 general reasoning tasks across eight domains by providing difficulty-controllable generators and rule-based verifiers. A key contribution is its empirical demonstration of ``Task Scaling,'' showing that increasing the number of training tasks significantly boosts both reasoning performance and training efficiency. Synthetic-2 \citep{synthetic2} contributes to this approach by providing a massive, open dataset of four million verified reasoning traces. These traces were collaboratively generated via a ``planetary-scale, pipeline-parallel, decentralized inference run,'' showcasing a highly scalable method for creating verified training data for complex RL tasks.

\label{sec:infra}
\subsection{RL Infrastructure}
\label{sec:resource_rl_infra}

\begin{myboxi}[Takeaways]
\begin{itemize}
    \item Modern RL infrastructure centers on flexible pipelines and communication layers that allocate resources between agent rollout and policy training, typically implemented as wrappers over mature distributed training frameworks and inference engines.
    \item Specialized variants (agentic workflows, multi-agent, and multimodal) commonly support asynchronous rollouts/training and standardized environment interfaces.
\end{itemize}
\end{myboxi}

In this subsection, we introduce the open-source RL infrastructure that promotes the development not only in algorithmic research but also in downstream applications. We begin by presenting primary development frameworks, which mainly provide basic wrappers around LLM training and inference frameworks. Next, we introduce secondary development frameworks, which are built upon these primary frameworks and further adapted to various downstream applications, including agentic RL, coding RL, multi-agent RL, and multimodal RL, distributed RL, and others. We compare these open-source RL frameworks in Table~\ref{tab:rl_infra} and introduce the main frameworks below..

\begin{table}[!t]
\small
\setlength{\tabcolsep}{5pt}             %
\renewcommand{\arraystretch}{1.15}      %
\centering
\caption{Open-source RL infrastructure for LLM post-training. Status legend: \cellcolor{yes}\cmark\ = native, \cellcolor{no}\xmark\ = unsupported, \cellcolor{partial}\textbf{P} = partial.}
\label{tab:rl_infra}
\resizebox{\textwidth}{!}{%
\begin{tabular}{l*{10}{c}}               %
\toprule
\multirow{2}[2]{*}{\textbf{Date}} &
\multirow{2}[2]{*}{\textbf{Framework}} &
\multicolumn{4}{c}{\textbf{Runtime}} &
\multicolumn{2}{c}{\textbf{Serving}} &
\multicolumn{3}{c}{\textbf{Training}}\\
\cmidrule(lr){3-6}\cmidrule(lr){7-8}\cmidrule(lr){9-11}
&  & Async & Agents & Multi-Agents & Multimodal & vLLM & SGLang & DeepSpeed & Megatron & FSDP \\
\midrule
\rowcolor{hdr}  &  \textit{Primary development} &  &  &  &  &  &  &  &  &  \\
2020.03 & \href{https://github.com/huggingface/trl}{TRL}                        & \xmark  & \xmark  & \xmark  &  P & \cmark  & \xmark  & \cmark  & \xmark  & \cmark \\
2023.11 & \href{https://github.com/OpenRLHF/OpenRLHF}{OpenRLHF}                 & \cmark  & \cmark  & \xmark  &  \xmark  & \cmark  & \xmark  & \cmark  & \xmark & \xmark\\
2024.11 & \href{https://github.com/volcengine/verl}{veRL}                       & \cmark  & \cmark  & \xmark  &  P & \cmark  & \cmark  & \xmark  & \cmark  & \cmark \\
2025.03 & \href{https://github.com/inclusionAI/AReaL}{AReaL}                    & \cmark  & \cmark  & \xmark  & P  & \cmark  & \cmark  & \cmark  & \cmark  & \cmark \\
2025.05 & \href{https://github.com/NVIDIA-NeMo/RL}{NeMo-RL}                     & P  & P  & \xmark  & \cmark  & \cmark  & \xmark  & \xmark  & \cmark  & \cmark \\
2025.05 & \href{https://github.com/alibaba/ROLL}{ROLL}                          & \cmark  & \cmark  & \xmark  & \cmark  & \cmark  & \cmark  & \cmark  & \cmark  & \xmark \\
2025.07 & \href{https://github.com/THUDM/slime}{slime}                          & \cmark  & P  & \xmark  & \xmark  & \xmark  & \cmark  & \xmark  & \cmark  & \xmark \\
2025.09 & \href{https://github.com/RLinf/RLinf}{RLInf}                         &   \cmark &  \cmark & \xmark  & \cmark  &  \cmark  & \cmark  & \xmark  & \cmark  & \cmark \\
\rowcolor{hdr}  &  \textit{Secondary development} &  &  &  &  &  &  &  &  &   \\
2025.02 & \href{https://github.com/rllm-org/rllm}{rllm}                         & P  & \cmark  & \xmark  & \xmark  & \cmark  & \cmark  & \xmark  & \xmark  & \cmark \\
2025.02 & \href{https://github.com/om-ai-lab/VLM-R1}{VLM-R1}                         & \xmark  & \xmark  & \xmark  & \cmark  & \cmark  & \xmark  & \cmark  & \xmark  & \xmark \\
2025.03 & \href{https://github.com/hiyouga/EasyR1}{EasyR1}                         & \xmark  & \xmark  & \xmark  & \cmark  & \cmark  & \xmark  & \xmark  & \xmark  & \cmark \\
2025.03 & \href{https://github.com/willccbb/verifiers}{verifiers}               & \cmark  & \cmark  & \xmark  & \xmark  & \cmark  & \xmark  & \cmark  & \xmark  & \cmark \\
2025.05 & \href{https://github.com/PrimeIntellect-ai/prime-rl/}{prime-rl}       & \cmark  & \xmark  & \xmark  & \xmark  &  \cmark &  \xmark & \xmark  & \xmark  & \cmark \\
2025.05 & \href{https://github.com/TsinghuaC3I/MARTI}{MARTI}                    & P  & \cmark  & \cmark  & \xmark  & \cmark  & \xmark  & \cmark  & \xmark  & \xmark \\
2025.05 & \href{https://github.com/Simple-Efficient/RL-Factory/}{RL-Factory}    & \cmark  & \cmark  & \xmark  & \cmark  & \cmark  & \cmark  & \cmark  & \cmark  & \cmark \\
2025.06 & \href{https://github.com/langfengQ/verl-agent}{verl-agent}            & \cmark  & \cmark  & \xmark &  \cmark &  \cmark & \cmark  & \cmark  & \cmark  & \cmark \\
2025.08 & \href{https://github.com/microsoft/agent-lightning}{agent-lightning}  & \cmark  & \cmark  & P  & \xmark  & \cmark  & \xmark  & \xmark  & \cmark  & \cmark \\
\bottomrule
\end{tabular}%
}
\end{table}

\paragraph{Primary Development.}
Current RL infrastructure relies heavily on mature training frameworks and inference engines designed for LLMs. Frameworks such as DeepSpeed~\citep{rasley2020deepspeed}, Megatron~\citep{megatron-lm}, and Fully Sharded Data Parallel (FSDP)~\citep{zhao2023pytorch} are optimized for both pre-training and post-training of LLMs. In terms of inference, vLLM~\citep{kwon2023efficient} and SGLang\footnote{\url{https://github.com/sgl-project/sglang}} are tailored for efficient inference, incorporating advanced schedulers and flash attention mechanisms. These optimizations enable significantly faster inference compared to direct forward computation on PyTorch models.
Many open-source RL frameworks are built upon plug-and-play training and inference frameworks, most of which are implemented on distributed computing engines such as Ray\footnote{\url{https://github.com/ray-project/ray}}.
Here, we review RL frameworks that are directly developed based on the aforementioned backbone training and inference frameworks.

\begin{itemize}
    \item \textbf{TRL}~\citep{vonwerra2022trl}: TRL focuses on trainer-centric post-training with SFT, PPO/GRPO, DPO, and a dedicated RewardTrainer (plus recent online variants), rather than a bespoke distributed runtime. It integrates vLLM for online methods (server or colocated modes) but does not natively target SGLang or TensorRT-LLM. Scaling is delegated to accelerate, which natively supports DDP, DeepSpeed ZeRO, and FSDP; Megatron is not a backend. Reward modeling is supported out-of-the-box through the RewardTrainer, and the library provides clear APIs for GRPO/DPO/online rollouts.
    \item \textbf{OpenRLHF}~\citep{hu2024openrlhf}: OpenRLHF provides distributed implementations of PPO, GRPO, REINFORCE++ (and its baseline variant) and RLOO, and also includes preference-learning baselines such as DPO/IPO/cDPO and KTO. Its runtime supports both asynchronous pipeline RLHF and asynchronous agentic RL modes, exposing a class-based agent API for multi-turn settings. For serving, OpenRLHF integrates tightly with vLLM for high-throughput rollouts. Training is organized around DeepSpeed ZeRO-3 with Auto Tensor Parallelism (AutoTP), without requiring Megatron or FSDP. The framework ships recipes for RMs and PRMs training and integrates PRM signals into rollouts. 
    \item \textbf{Verl}~\citep{sheng2025hybridflow}: Verl offers one of the broadest algorithm menus (PPO, GRPO, GSPO, ReMax, REINFORCE++, RLOO, PRIME, DAPO/DrGRPO, and more) together with multi-turn training and tool use. Its runtime is centered on the HybridFlow controller and adds agentic RL rollout and prototypes for disaggregated asynchronous training (with ``Async and off-policy architecture'' on the public roadmap). Verl supports vLLM and SGLang for serving, and provides both FSDP and Megatron-LM training backends. Reward options include model-based and function/verifiable rewards (e.g., math/coding), with multi-GPU LoRA-RL support.
    \item \textbf{AReaL}~\citep{fu2025areal}: AReaL targets high-throughput RL for large reasoning models with a fully asynchronous design that decouples generation from training via interruptible rollout workers, a replay buffer, and a parallel reward service (e.g., unit-test–based code rewards), stabilized by a staleness-aware PPO objective. Empirically, the system reports up to $2.77\times$ training speedups at matched or better final accuracy on math/code benchmarks and scales near-linearly to 512 GPUs. The open-source stack emphasizes SGLang-based rollout serving and Ray launchers for single-node to $\sim$1K-GPU clusters, with PyTorch FSDP as the main training backend (Megatron also available); the newer ``AReaL-lite'' adds an algorithm-first API with GRPO examples and support for multi-turn agentic RL/RLVR workflows.
    \item \textbf{NeMo-RL}~\citep{nemo-rl}: NVIDIA's NeMo stack now exposes a dedicated ``NeMo RL'' library and the earlier NeMo-Aligner toolkit for alignment. Algorithmically, NeMo covers SFT and preference training (DPO/RPO/IPO/REINFORCE) as well as full RLHF with PPO and GRPO, including multi-turn variants. The runtime emphasizes scalable, production-oriented orchestration and extensive parallelism; training is built on Megatron Core (tensor/data/pipeline/expert parallelism) for 100B-scale models and multi-node clusters. For serving, the NeMo framework documents deployment with TensorRT-LLM and vLLM. Reward-model training is first-class in the RLHF tutorials, with end-to-end pipelines from RM fitting to PPO.
    \item \textbf{ROLL}~\citep{wang2025reinforcement-roll}: ROLL targets large-scale RL for LLMs with GRPO/PPO/REINFORCE++ and additional recipes (e.g., TOPR/RAFT++/GSPO), and explicitly supports asynchronous training and agentic RL pipelines. The runtime follows a Ray-based multi-role design and integrates SGLang and vLLM for rollout serving. Training is built primarily around Megatron-Core, with FSDP2 listed on the public roadmap; DeepSpeed is acknowledged as a dependency. Reward handling is modular via Reward Workers (e.g., verifiers, sandbox tools, LLM-as-judge) and pluggable environments. A technical report details the system and scaling considerations.
    \item \textbf{slime}~\citep{thudm-slime}: Slime is positioned as an SGLang-native post-training framework for RL scaling, connecting SGLang on the rollout side with Megatron on the training side. It emphasizes infrastructure over algorithm breadth, but ships examples for dense and MoE models, and includes multi-turn + tool-calling (``Search-R1 lite''). The runtime supports asynchronous training and agentic workflows; serving is first-class via SGLang. Training uses Megatron-LM with Ray for cluster launch; reward modeling per se is not the primary focus, although verifier/``reward'' signals can be produced on the rollout plane.
    \item \textbf{RLinf}~\citep{RLinf_repo}: RLinf is a framework for embodied intelligence that emphasizes modularity and adaptability. Motivated by the coexistence of ``large-brain'' and ``small-brain'' paradigms and the field's still-evolving trajectory, RLinf adopts a Macro-to-Micro Flow (M2Flow) paradigm, which separates macro-level logical workflows from micro-level physical execution, thereby enabling programmable composition with efficient scheduling. At runtime, RLinf allows reinforcement learning components (e.g., Actor, Critic, Reward, Simulator) to be flexibly placed on arbitrary GPUs and configured with collocated, disaggregated, or hybrid execution modes—ranging from shared-all placement to fine-grained pipelining. A representative case decouples the Generator and GPU-based Simulator into a pipeline, while Inference and Trainer share execution. For serving, RLinf supports vLLM/SGLang, and for training it integrates Megatron/FSDP.
\end{itemize}

\paragraph{Secondary Development.}
In this part, we introduce several representative frameworks that are built upon primary development frameworks and extend their features to support a broader range of downstream applications. We primarily focus on frameworks for agentic RL, multimodal RL, and multi-agent RL. Although some primary frameworks already offer partial support for these areas, we highlight specialized frameworks designed for specific domain studies.

\begin{itemize}
    \item \textbf{Agentic RL}: This area focuses on training LLMs to utilize external tools in a variety of scenarios, such as search engines~\citep{jin2025search}, Python interpreters~\citep{feng2025retool}, web browsers~\citep{li2025websailor}, and more. Primary frameworks like veRL~\citep{sheng2025hybridflow} and AReaL~\citep{fu2025areal} have been updated or specifically designed to support these capabilities. A core feature of agentic RL is asynchronous generation and training, which significantly reduces computational time during long-term interactions between LLMs and external environments. The secondary frameworks are mostly built upon veRL to integrate additional tools and environments, and their new features are gradually incorporated back into veRL. More details about Agentic RL will be discussed in $\S$~\ref{sec:application_coding} and~\ref{sec:application_agentic}.
    \item \textbf{Multimodal RL}: Although the primary development frameworks were originally designed for training language models, they are typically based on transformers, which support both inference and training of vision language models. The main challenges in this area involve data processing and loss function design. Notable frameworks such as VLM-R1~\citep{shen2025vlm} and EasyR1~\citep{zheng2025easyr1} have been developed for training vision-language models based on veRL. For multimodal generation, certain frameworks have been specifically developed for RL training of diffusion-based models, such as DanceGRPO~\citep{xue2025dancegrpo}. However, these approaches are beyond the scope of this paper, and readers may refer to recent RL surveys focused on vision models for further details~\citep{wu2025reinforcement}. More details about Multimodal RL will be discussed in $\S$~\ref{sec:application_multimodal}.
    \item \textbf{Multi-Agent RL}: Frameworks for agentic RL primarily focus on implementing dynamic workflows for asynchronous rollouts and training. While most of these frameworks are still limited to single-agent applications, LLM-based MARL remains an area under active exploration. \citet{marti2025} propose the first high-performance, open-source framework for LLM-based multi-agent reinforced training and inference, enabling centralized interactions and distributed policy training. In addition, recent frameworks such as Agent-Lightning~\citep{luo2025agent} have implemented disentanglement of training and inference, making it easier to support multi-agent training. More details about Multi-Agent RL will be discussed in $\S$~\ref{sec:application_mas}.
\end{itemize}

\label{sec:infras}

\section{Applications}
\label{sec:application}
\begin{figure}
\footnotesize
\begin{forest}
    for tree={
        forked edges,
        grow'=0,
        draw,
        rounded corners,
        node options={align=center,},
        text width=2.7cm,
        s sep=6pt,
        calign=child edge, calign child=(n_children()+1)/2,
    },
    [Applications~\S~\ref{sec:application}, fill=gray!45, parent
        [Coding Tasks~\S~\ref{sec:application_coding}, for tree={fill=blue!45, answer}
            [Code Generation, answer
                [{\tstyle{Competitive-Code}: e.g., Code-R1 \citep{code-r1}; Open-R1 \citep{faceopen}; DeepCoder \citep{luo2025deepcoder}; AceReason-Nemotron \citep{chen2025acereason}; SkyWork OR1 \citep{he2025skywork}; AReaL \citep{fu2025areal}}
                , answer_work]
                [{\tstyle{Domain-Specific-Code}: e.g., Reasoning-SQL \citep{pourreza2025reasoning}; ReEX-SQL \citep{dai2025reex}; CogniSQL-R1-Zero \citep{gajjar2025cognisql}; Kimina-Prover \citep{wang2504kimina}; DeepSeek-Prover-v2 \citep{ren2025deepseek}; StepFun-Prover \citep{shang2025stepfun}; Leanabell-Prover-V2 \citep{ji2025leanabell}; MedAgentGym \citep{xu2025medagentgym}; VeriReason \citep{wang2025verireason}; CodeV-R1 \citep{zhu2025codev}}
                , answer_work]
            ]
            [Software Engineering, answer
                [{\tstyle{Code-Quality-Improvement}: e.g., RePaCA \citep{fuster2025repaca}; Repair-R1 \citep{hu2025repair}; CURE \citep{wang2025co}; Afterburner \citep{du2025afterburner}; REAL \citep{yao2025training}}
                , answer_work]
                [{\tstyle{Repository-Level-Code-Generation}: e.g., RLCoder \citep{wang2024rlcoder}; RepoGenReflex \citep{wang2024repogenreflex}; SWE-RL \citep{wei2025swe}; Satori-SWE \citep{zeng2025satori}}
                , answer_work]
            ]
        ]
        [Agentic Tasks~\S~\ref{sec:application_agentic}, for tree={fill=blue!45, multiple}
            [Agentic Coding , multiple
                [{e.g., SWE-RL \citep{wei2025swe}; Satori-SWE \citep{zeng2025satori}; Kimi K2 \citep{team2025kimi}; Qwen3 Coder \citep{yang2025qwen3}; GLM4.5 \citep{zeng2025glm}; ARPO~\citep{dong2025agentic}; AutoTIR~\citep{wei2025autotir}; CoRT~\citep{li2025cort}; ToRL~\citep{li2025torl}; FormaRL~\citep{huang2025formarl}; MLE-bench \citep{chan2024mle}; MLE-STAR \citep{nam2025mle}; ML-Agent \citep{liu2025ml}}
                , multiple_work]
            ]
            [Search \& Deep Research, multiple
                [{e.g., Search-R1 \citep{jin2025search}; R1-Searcher \citep{song2025r1}; DeepResearcher \citep{zheng2025deepresearcher}; ZeroSearch \citep{sun2025zerosearch}; SSRL \citep{fan2025ssrl}; R1-Searcher++ \citep{song2025r1++}; $O^2$-searcher \citep{mei20252}; ReZero \citep{dao2025rezero}; S3 \citep{jiang2025s3}; WebSailor \citep{li2025websailor}; WebShaper \citep{tao2025webshaper}; WebThinker \citep{li2025webthinker}; WebGPT \citep{nakano2021webgpt}; Web-RL \citep{qi2024webrl}; WebAgent-R1 \citep{wei2025webagent}; WebDancer \citep{wu2025webdancer}; Kimi-Searcher \citep{kimi-researcher2025}; Jan-nano \citep{dao2025jan}; MicroThinker \citep{2025mirothinker}; Webwatcher \citep{geng2025webwatcher}; Atom-Searcher \citep{deng2025atom}; WebExplorer \citep{liu2025webexplorerexploreevolvetraining}; SFR-DeepResearch \citep{nguyen2025sfrdeepresearcheffectivereinforcementlearning}; MedResearcher-R1 \citep{yu2025medreseacher}}
                , multiple_work]
            ]
            [GUI \& Compute-use, multiple
                [{e.g., UI-R1 \citep{lu2025ui}; GUI-R1 \citep{luo2025gui}; GUI-Critic-R1~\citep{wanyan2025look}; GUI-G1 \citep{zhou2025gui}; InfiGUI-R1 \citep{liu2025infigui}; GUI-Reflection~\citep{wu2025gui}; UIShift~\citep{gao2025uishift}; ZeroGUI~\citep{yang2025zerogui}; WEBAGENT-R1~\citep{wei2025webagent}; ARPO~\citep{lu2025arpo}; Computer-RL~\citep{lai2025computerrlscalingendtoendonline}}
                , multiple_work]
            ]
            [Other-tasks, multiple
                [{e.g., AdLlama \citep{jiang2025improving}; Shop-R1 \citep{zhang2025shop}; LaviPlan \citep{oh2025laviplan}; Drive-R1 \citep{li2025drive}; OpenTab-R1 \citep{qiu2025opentable}; TooRL~\citep{qian2025toolrl}; K2 \citep{team2025kimi}}
                , multiple_work]
            ]
        ]
        [Multimodal Tasks~\S~\ref{sec:application_multimodal}, for tree={ pretrain}
            [Multimodal Understanding,  pretrain
                [{\tstyle{Image}: e.g., Vision-R1~\citep{huang2025vision}; VLM-R1~\citep{shen2025vlm}; Taco~\citep{kan2025taco}; Visionary-r1~\citep{xia2025visionary}; Visual-RFT~\citep{liu2025visual}; Deepeyes~\citep{zheng2025deepeyes}; CoF~\citep{zhang2025chain}; Ground-r1~\citep{cao2025ground}; Grit~\citep{fan2025grit}}
                , pretrain_work]
                [{\tstyle{Video}: e.g., Video-R1~\citep{feng2025video-r1}; Focused Thinking~\citep{dang2025focused-thinking}; VQ-Insight~\citep{zhang2025vq-insight}; Ego-R1~\citep{tian2025ego}; Long-RL~\citep{chen2025long-rl-video}; Video-RFT~\citep{wang2025videorft}; VideoChat-R1~\citep{li2025videochat-r1}}
                , pretrain_work]
                [{\tstyle{3D}: e.g., MetaSpatial~\citep{pan2025metaspatial}; Spatial-MLLM~\citep{wu2025spatial-mllm}; SpaceR~\citep{ouyang2025spaceR}; 3D-R1~\citep{huang20253d-r1}; R1-Zero-VSI~\citep{liao2025improved-visual-spatial}}
                , pretrain_work]
            ]
            [Multimodal Generation,  pretrain
                [{\tstyle{Image}: e.g., DanceGRPO~\citep{xue2025dancegrpo}; Flow-GRPO~\citep{liu2025flow}; Qwen-Image~\citep{wu2025qwen}; MixGRPO~\citep{li2025mixgrpo}; TempFlow-GRPO~\citep{he2025tempflow}; SimpleAR~\citep{wang2025simplear}; FocusDif~\citep{pan2025focusdiff}; RePrompt~\citep{wu2025reprompt}; T2I-R1~\citep{jiang2025t2i}; GoT-R1~\citep{duan2025got}; ReasonGen-R1~\citep{zhang2025reasongen}; CoRL~\citep{jiang2025co}; DSR~\citep{hong2025reinforcing}}
                , pretrain_work]
                [{\tstyle{Video}: e.g., DanceGRPO~\citep{xue2025dancegrpo}; InfLVG~\citep{fang2025inflvg}; Phys-AR~\citep{lin2025reasoning}}
                , pretrain_work]
            ]
        ]
        [Multi-Agent Systems~\S~\ref{sec:application_mas}, for tree={fill=red!45,template}
            [LLM-based MAS,  template
                [{e.g., LLaMAC \citep{zhang2023controlling}; CTRL \citep{xie2025teaching}; MAPoRL \citep{park2025maporl}; MAGRPO \citep{liu2025llm}; ReMA \citep{wan2025rema}; JoyAgents-R1 \citep{han2025joyagents}}
                , template_work]
            ]
        ]
        [Robotics Tasks~\S~\ref{sec:application_robotics}, for tree={fill=red!45,template}
            [Dual/Single-Arm Manipulation,  template
                [{e.g., SimpleVLA-RL~\citep{li2025simplevla}; VLA-RL~\citep{lu2025vla}; VLA RL Generalization~\citep{liu2025can}; RIPT-VLA~\citep{tan2025interactive}; ConRFT~\citep{chen2025conrft}; RLinf~\citep{RLinf_repo}}
                , template_work]
            ]
        ]
        [Medical Tasks~\S~\ref{sec:application_medical}, for tree={fill=blue!45, tuning}
            [Medical Understanding, tuning
                [{e.g., Med-U1 \citep{zhang2025med-u1}; MED-RLVR \citep{zhang2025med}; 
                 Open-Medical-R1 \citep{qiu2025openmedicalr1choosedatarlvr}; 
                 Gazal-R1 \citep{arora2025gazalr1}; ProMed \citep{ding2025promed};
                 MedVLM-R1 \citep{pan2025medvlmr1}; 
                 MedGround-R1 \citep{xu2025medgroundr1}; ARMed \citep{liu2025armed}; 
                 MMedAgent-RL \citep{xia2025mmedagent};
                 MedGR$^2$ \citep{zhi2025medgr2};
                 }
                , tuning_work]
            ]
            [Medical Generation, tuning
                [{e.g., FineMedLM-o1 \citep{yu2025finemedlmo1enhancingmedicalknowledge}; 
                 Med-REFL \citep{yang2025medreflmedicalreasoningenhancement}; 
                 DOLA \citep{nusrat2025dola}; 
                 LA-CDM \citep{baniharouni2025lacdm}; PPME \citep{sun2025ppme}; 
                 Baichuan-M1 \citep{2025baichuan};
                 Baichuan-M2 \citep{2025baichuan2};
                 MORE-CLEAR \citep{lim2025moreclear}}
                , tuning_work]
            ]
        ]
    ]
\end{forest}
\caption{Taxonomy of applications, including research directions and representative works.}
\label{fig:taxonomy_applications}
\end{figure}
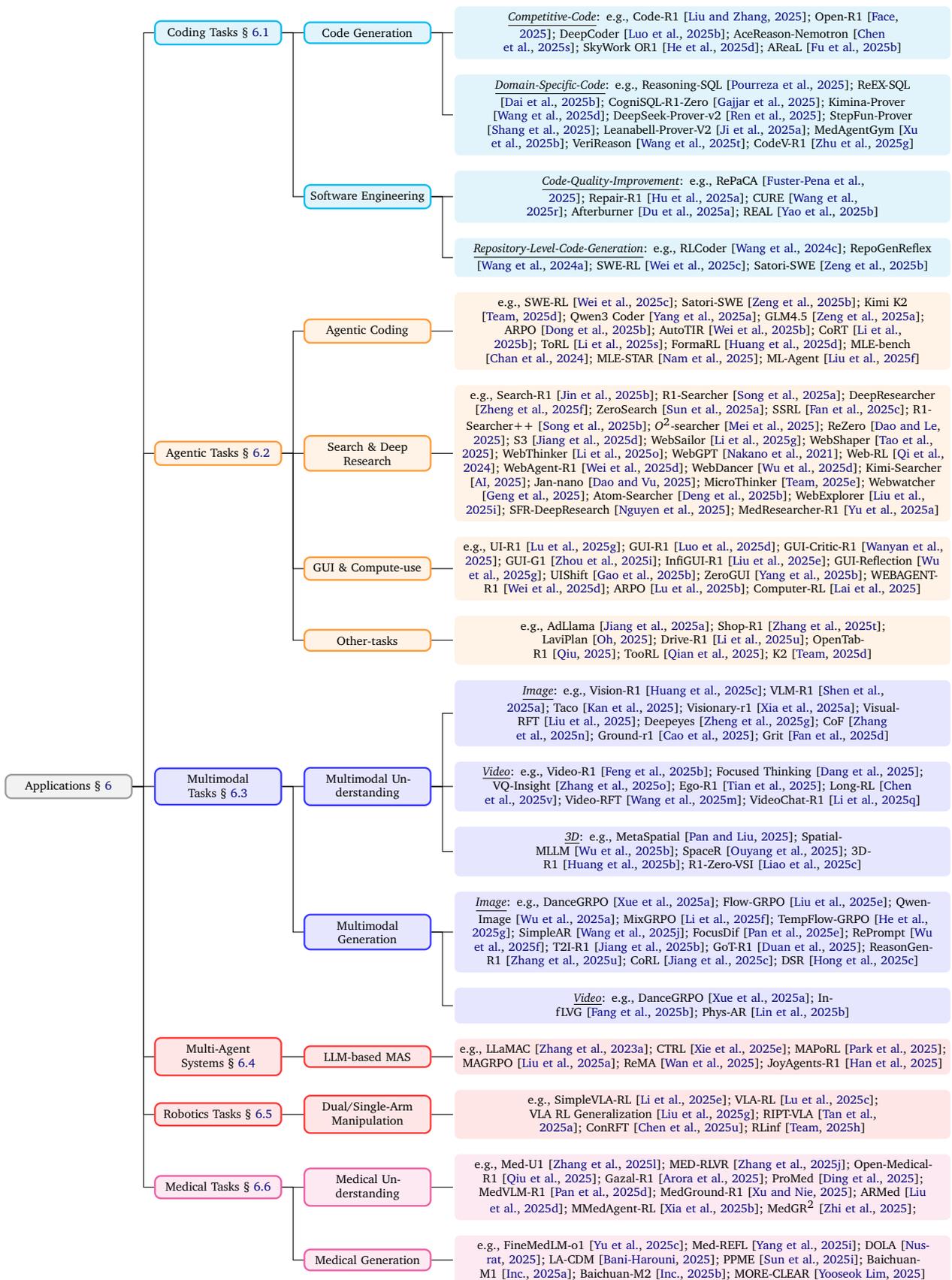

Advancements in RL for LLMs are best understood through their practical impact across a variety of domains. In this section, we review recent progress and challenges associated with applying RL-trained language models to real-world tasks. We highlight how RL-driven methods have improved capabilities in coding tasks ($\S$~\ref{sec:application_coding}), enabled more autonomous and adaptive agentic behaviors ($\S$~\ref{sec:application_agentic}), and extended LLMs to multimodal reasoning across text, vision, and beyond ($\S$~\ref{sec:application_multimodal}). Further, we discuss applications in multi-agent systems ($\S$~\ref{sec:application_mas}), robotics ($\S$~\ref{sec:application_robotics}), and medicine ($\S$~\ref{sec:application_medical}), illustrating both the broad potential and unique requirements of each area.
We provide the overall taxonomy of applications along with corresponding related works in Figure~\ref{fig:taxonomy_applications}.

\subsection{Coding Tasks}
\label{sec:application_coding}
\begin{myboxi}[Takeaways]
\begin{itemize}
    \item RL has advanced LLMs' reasoning and code generation in competitive programming and domain-specific tasks, driving progress toward agentic, closed-loop coding.
	\item However, scalability, cross-task generalization, and robust automation in large-scale software settings remain open challenges.
\end{itemize}
\end{myboxi}

Recently, numerous studies have demonstrated that RL offers significant advantages in verifiable tasks.
Given the inherent verifiability and practical importance of coding tasks, RL has become a core approach for improving code reasoning and continues to attract substantial attention.
To systematically review the field, we categorize existing research into three directions: code generation, software engineering assistance, and agentic coding, based on task complexity and developmental trend, from simpler verifiable tasks toward more complex, autonomous agentic coding.

\paragraph{Code Generation.}
The primary objective of this direction is to generate correct and executable code. Research focuses on using RL to adjust LLM generation distributions to meet the requirements of diverse coding tasks. Following the demonstration of RL's potential for complex reasoning in DeepSeek-R1, an increasing number of studies have applied RL to code generation.
\begin{itemize}
    \item \textbf{Competitive Programming}: Competitive programming, one of the earliest benchmarks, has inspired studies including Code-R1~\citep{code-r1}, Open-R1~\citep{faceopen}, DeepCoder~\citep{luo2025deepcoder}, AceReason-Nemotron~\citep{chen2025acereason}, SkyWork-OR1~\citep{he2025skywork}, and AReaL~\citep{fu2025areal}, which replicate DeepSeek-R1 results in code tasks. To address RL training instabilities and slow inference, DeepCoder~\citep{luo2025deepcoder} and SkyWork OR1~\citep{he2025skywork} adopted staged RL training, progressively increasing the context length to stabilize the learning process; DeepCoder~\citep{luo2025deepcoder} and AReaL~\citep{fu2025areal} further employed asynchronous rollouts to decouple training from inference and accelerate learning. To address the lack of explicit abstract reasoning capabilities in code generation, AR$^2$~\citep{yeh2025ar} framework (Adversarial Reinforcement Learning for Abstract Reasoning) iteratively trains teacher and student models by RLVR. In addition to attempts at code generation using autoregressive models, Dream-Coder~\citep{xie2025dream} incorporates the RLVR training paradigm into diffusion models, achieving faster generation speeds. Regarding cross-task generalization, AceReason-Nemotron~\citep{chen2025acereason} observed a positive transfer effect from mathematical reasoning tasks to competitive programming.
    \item \textbf{Domain-Specific Code}: Due to domain-specific differences in code requirements, RL is increasingly applied to specialized tasks. In data retrieval, Reasoning-SQL~\citep{pourreza2025reasoning}, ReEX-SQL~\citep{dai2025reex}, and CogniSQL-R1-Zero~\citep{gajjar2025cognisql} applied the GRPO algorithm to Text-to-SQL tasks, achieving notable performance on corresponding benchmarks. In formal proofs, Kimina-Prover~\citep{wang2504kimina} and DeepSeek-Prover-v2~\citep{ren2025deepseek} unified informal and formal proofs by combining natural language with Lean, while StepFun-Prover~\citep{shang2025stepfun} developed an end-to-end tool-integrated training pipeline, and Leanabell-Prover-V2~\citep{ji2025leanabell} directly optimized reasoning trajectories via multi-round verifier feedback, further advancing RL's capabilities in this field. In other domains, MedAgentGym~\citep{xu2025medagentgym} provided an executable coding environment for large-scale trajectory generation to improve LLM-based medical reasoning; VeriReason~\citep{wang2025verireason} , Proof2Silicon~\citep{jha2025proof2silicon} and CodeV-R1~\citep{zhu2025codev} extended RLVR to the field of electronic design automation (EDA), accelerating LLM-driven hardware design. Additionally, chart-to-code generation enables agents to process structured or visual inputs and translate them into executable code, exemplifying cross-modal domain-specific code generation~\citep{chen2025breaking}.
\end{itemize}

\paragraph{Software Engineering.}
Despite progress in competitive programming and domain-specific tasks, these studies often fall short of real-world software development environments. Consequently, RL research also focuses on real-world software engineering, including code repair, quality optimization, and repository-level generation.
\begin{itemize}
    \item \textbf{Code Quality Improvement}: Automated code repair and quality improvement enhance software reliability while preserving functionality. RL significantly improves repair effectiveness and generalization, enabling models to handle unseen defects. RePaCA~\citep{fuster2025repaca} mitigates APR patch overfitting by guiding LLMs with chain-of-thought reasoning and GRPO-based fine-tuning, while Repair-R1~\citep{hu2025repair} jointly optimizes test-case generation and repair, reducing reliance on post-hoc validation. Beyond bug fixing, RL enhances code efficiency, maintainability, readability, and security. CURE~\citep{wang2025co} and UTRL~\citep{lee2025learning} evolve code and unit tests via encoder-tester interactions without ground-truth supervision, and Afterburner~\citep{du2025afterburner} leverages execution feedback, raising \passo~from 47\% to 62\% and surpassing human-level efficiency. REAL~\citep{yao2025training} integrates program analysis and unit testing as hybrid rewards to improve scalability and quality, achieving high-quality code generation without human intervention.
    \item \textbf{Repository-Level Code Generation}: Beyond function- and snippet-level tasks, recent work explores repository-level code generation and maintenance, emphasizing consistency and maintainability across complex cross-file and cross-module dependencies. RLCoder~\citep{wang2024rlcoder} combines Retrieval-Augmented Generation (RAG) with RL to train a retriever and improve code completion accuracy. RepoGenReflex~\citep{wang2024repogenreflex} further introduces a reflection mechanism to evaluate generated results and provide feedback, continuously optimizing generation strategies and improving generalization. By integrating RL with automated testing and continuous integration, this approach aligns LLM optimization with real-world development processes, advancing software engineering automation.
\end{itemize}

\subsection{Agentic Tasks}
\label{sec:application_agentic}
\begin{myboxi}[Takeaways]
\begin{itemize}
    \item Agentic RL enables advanced behaviors but faces scalability issues from high computational costs and long rollout times within environments.
    \item Asynchronous rollouts and memory agents help reduce latency and manage context, but further progress relies on better training data.
\end{itemize}
\end{myboxi}

Tool use is considered a fundamental ability of language models \citep{schick2023toolformer}. Recent works leverage RL to help LLMs master tools and complete more complex problems~\citep{team2025kimi,dong2025tool}. We group them into \textbf{Coding Agent}, \textbf{Simple Search Agent}, \textbf{Browser-use Agent}, \textbf{DeepResearch}, \textbf{GUI \& Computer-use Agent}, and \textbf{Other Tasks}.

\paragraph{Coding Agent.}
The integration of RL and agent paradigms has advanced code generation from single-step outputs to multi-round interactions and autonomous iteration, endowing LLMs with execution and verification capabilities for closed-loop optimization.
\begin{itemize}
    \item \textbf{Code Agents}: A common practice is to integrate RL into code agents equipped with execution and verification capabilities, and evaluate them on realistic benchmarks such as SWE-Bench. SWE-RL~\citep{wei2025swe} applies GRPO to the patch generation–execution–correction loop, enabling continuous policy optimization and improving mathematical reasoning, general code generation, and cross-domain tasks. EvoScale (Satori-SWE)~\citep{zeng2025satori} allows agents to autonomously enhance patch quality without external verifiers. RL-enhanced models such as Kimi-K2~\citep{team2025kimi}, Qwen3-Coder, and GLM-4.5 demonstrate stronger agentic behavior, promoting greater autonomy and scalability. \cite{sinha2025illusion} investigate long-horizon execution in LLMs and demonstrates that improvements in single-step accuracy do not necessarily translate to successful multi-step task performance due to error accumulation.
    \citet{du2025generalizable} introduce CodeGym, a scalable RL framework that synthesizes multi-turn tool-use environments from coding problems to enable generalizable LLM agent training.
    These developments suggest that combining RL with agentic coding is driving a shift from ``single-step generation'' toward ``autonomous iteration.''
    \item \textbf{Tool-Integrated Reasoning}: Another emerging application of RL lies in Tool-Integrated Reasoning (TIR), which enhances LLMs' code reasoning capabilities by tightly coupling natural language reasoning with external tool execution environments. This approach enables models to generate, execute, and verify intermediate code or program outputs, reducing errors and improving verifiability. Representative works such as ARPO~\citep{dong2025agentic}, AutoTIR~\citep{wei2025autotir}, CoRT~\citep{li2025cort}, and ToRL~\citep{li2025torl} adopt similar strategies: models are post-trained with SFT or RL (mainly GRPO or variants), and outputs are structured (e.g., <code>...</code>) to trigger tool execution, feeding results back into the reasoning loop. \cite{paprunia2025advancing,xue2025simpletir,li2025encouraging} extend RL-based Tool-Integrated Reasoning by improving small LLMs’ tool-use capability, stabilizing multi-turn reasoning, and rewarding tool-use sequences independent of final answers. This tight integration provides explicit RL reward signals, guiding models to produce logically consistent outputs and iteratively refine them through verifiable computation. Additionally, autoformalization approaches such as FormaRL~\citep{huang2025formarl} extend TIR to Lean-based formal proof generation by integrating compiler-based syntax checks and LLM consistency evaluation with minimal labeled data, further improving reliability and correctness.
    \cite{tan2025process} introduce Turn-level Adjudicated Reinforcement Learning (TARL) to train multimodal agents for tool-integrated reasoning, leveraging a sandbox environment and mixed-task training curriculum.
    Tool-R1~\citep{zhang2025tool} introduces an RL framework to enable LLMs to perform multi-step tool use with outcome-based reward functions and dynamic sampling strategies, achieving improved accuracy and robustness on complex tasks.
    \item \textbf{Automated ML Programming}: RL shows promise in automated machine learning (AutoML), expanding code agents into ML engineering agents (MLE agents) capable of autonomous data processing, model building, and optimization. MLE-bench~\citep{chan2024mle} evaluates ML agent capabilities; MLE-STAR~\citep{nam2025mle} proposes a search- and optimization-based ML engineering agent; ML-Agent~\citep{liu2025ml} shows RL-driven autonomous ML engineering. \cite{yang2025reinforcement} show that agents powered by relatively smaller models trained with RL can outperform agents using much larger but static models, especially when enhanced with duration-aware updates and environment instrumentation to provide finer-grained reward signals.
\end{itemize}

\paragraph{Simple Search Agent.} 
LLMs can be trained to function as search agents through structured prompting, multi-turn generation, and integration with either online search engines (e.g., Google) or static local corpora such as Wikipedia \citep{jin2025empirical, jin2025search, song2025r1}. However, training with online search engines often incurs substantial API costs, making this approach prohibitively expensive. To address this challenge, \citet{sun2025zerosearch} propose simulating a search engine during the training of search-capable LLMs, significantly reducing costs while maintaining or even improving performance. Other works such as R1-Search++ \citep{song2025r1++} and SEM \citep{sha2025sem} leverage the internal knowledge of LLMs to reduce training budgets while yielding better performance. Specifically, SSRL~\citep{fan2025ssrl} proposes training models in fully-simulated environments that can be seamlessly adapted to real scenarios through Sim2Real Generalization. Meanwhile, diverse reward signals can be developed for specific applications. \citet{mei20252, dao2025rezero} employ diversity rewards to encourage comprehensive yet accurate information gathering. \citet{wang2025stepsearch} leverage step-level rewards to further enhance the performance of search agents. S3 \citep{jiang2025s3} utilizes gains beyond RAG to achieve better performance with fewer data. To enhance LLMs' capabilities on more challenging queries, such as those in benchmarks like GAIA \citep{mialon2023gaia} and BrowseComp \citep{wei2025browsecomp}, WebSailor \citep{li2025websailor} constructs training data from knowledge graphs, enabling models to search and browse open web environments to solve obscure problems. WebShaper \citep{tao2025webshaper} introduces a formalized data construction framework aimed at improving general AI assistants' problem-solving abilities.

\paragraph{Browser-use Agent.}
Besides using search engines, other browser-user agents leverage web-browsing as well. WebGPT \citep{nakano2021webgpt} uses textual web description to train a model to possess the ability to browse websites. Web-RL \citep{qi2024webrl} employs a curriculum strategy along with ORM to convert LLMs into web agents. DeepResearcher \citep{zheng2025deepresearcher} leverages another LLM to serve as a summarizer when browsing to help the search process. \citet{vattikonda2025train} bootstrap to train a student model using a variety of hyperparameters for stable training and better performance. WebAgent-R1 \citep{wei2025webagent} proposes a multi-turn asynchronous GRPO to train an end-to-end web browse agent, achieving strong performance. WebDancer \citep{wu2025webdancer} conducts SFT and RL to enable in-depth information
seeking and multi-step reasoning by web searching and browsing. Besides, other tasks are calling for a web agent, e.g., Academic Browse \citep{zhou2025academicbrowse}.

\paragraph{DeepResearch Agent.}
DeepResearch is introduced for gathering information from various sources online to help complete real-world problems, e.g., report generation.
WebThinker \citep{li2025webthinker}, trained with iterative DPO, leverages the long-cot abilities of LRMs, using deep web explorer along with an LLM writer to finish challenging tasks. Kimi-Searcher \citep{kimi-researcher2025} identifies the dilemma of multi-agent, and automatically constructs intensive tool-use data to end-to-end train a single agent model, achieving great performance on HLE \citep{prabhudesai2025maximizing}. Jan-nano \citep{dao2025jan} eliminates the need for cold-start or SFT by taking multi-stage RLVR, focusing on tool calling, answering quality, and extending response length, respectively. MicroThinker \citep{2025mirothinker} uses SFT and DPO to train Qwen3 \citep{wu2025qwen}, enhancing its performance in real-world applications. Recently, WebWatcher is proposed \citep{geng2025webwatcher} which is a multi-modal deepresearch-model capable of using external tools and visual information to solve extremely complex problems. Atom-Searhcer \citep{deng2025atom} leverages an LRM as a PRM to provide fine-grained reward signals during training, achieving better performance. ASearcher \citep{gao2025beyond} scales the interaction turns to more than 10 turns to elicit the reasoning capability of the deep research agent. WebExplorer \citep{liu2025webexplorerexploreevolvetraining} employs a model-based data synthesization method to construct high-quality data, achieving even better performance. SFR-DeepResearch \citep{nguyen2025sfrdeepresearcheffectivereinforcementlearning} empowers a single agent with minimal turns of tool usage, and results in comparable performance with longer trajectories. Besides general QA tasks, MedResearcher-R1 \citep{yu2025medreseacher} is proposed to solve clinical questions.

\paragraph{GUI \& Computer-use Agent.}
UI-R1~\citep{lu2025ui} is the first work to apply rule-based RL to graphical user interface (GUI) tasks. It introduces a novel rule-based action reward and is optimized using a small, human-curated training set.
Building on this practice, GUI-R1~\citep{luo2025gui}, GUI-Critic-R1~\citep{wanyan2025look}, and so on~\citep{du2025test,lin2025r1,ai2025inquiremobile}, carefully design fine-grained rule-based rewards tailored to specific objectives of GUI tasks, such as action accuracy, argument correctness, and step-level status.
GUI-G1~\citep{zhou2025gui} presents an empirical analysis of prior methods, identifying issues such as length bias, difficulty bias, and susceptibility to reward hacking, and reformulates the reward normalization scheme to mitigate these limitations.
Furthermore, recent studies~\citep{gu2025mobile,shi2025mobilegui} have attempted to obtain feedback from online GUI environments to better simulate real-world operating conditions.
GUI-Reflection~\citep{wu2025gui} and UIShift~\citep{gao2025uishift} derive binary rewards based on changes of UI elements to indicate action success or failure.
\citet{liu2025infigui} propose a two-stage training paradigm that explicitly enhances planning and reflective reasoning capabilities.
ZeroGUI~\citep{yang2025zerogui} introduces an automated pipeline for generating challenging tasks and estimates rewards solely based on online environmental feedback, eliminating the need for human annotation.
Different from the above step-level methods, there is a growing trend towards applying end-to-end asynchronous RL frameworks to train agents for mobile~\citep{lu2025arpo,ye2025mobile,lu2025swirl}, and computer~\citep{lai2025computerrlscalingendtoendonline} use, which optimize the model using only rule-based task-level completion rewards without requiring step-wise reward signals. UI-TARS \citep{wang2025ui} learns from mistakes and adapts to unforeseen situations through iterative training and reflection tuning. To step forward, UI-TARS 2~\citep{qin2025ui} features enhanced capabilities in GUI, Game, Code and Tool Use with end-to-end RL.
GUI-Cursor~\citep{zhao2025learninggui} reframes GUI grounding as an interactive search task using multi-step online RL and dense rewards to improve accuracy in identifying on-screen coordinates.

\paragraph{Other Tasks.}
Beyond search and GUI agents, RL has also been successfully applied to a variety of other agentic tasks. For example,~\citet{jiang2025improving} improve ad copy generation by leveraging historical performance metrics, such as click-through rates, as reward signals to guide RL-based optimization.
In the e-commerce domain, Shop-R1 \citep{zhang2025shop} introduces a composite reward function that combines internal model logits with external hierarchical feedback to better simulate human-like decision-making in shopping environments. For autonomous driving, LaviPlan \citep{oh2025laviplan} aligns perceptual vision capabilities with context-aware decision-making, enabling more robust navigation under dynamic conditions. Similarly, Drive-R1 \citep{li2025drive} is designed to balance reasoning and planning abilities for complex driving scenarios, improving both strategic and reactive behavior. In structured data interaction, OpenTab-R1 \citep{qiu2025opentable} employs a two-stage training framework to enhance LLMs' proficiency in table-based question answering. Furthermore, general-purpose agentic models such as those in \citet{qian2025toolrl} and \citet{team2025kimi} demonstrate the ability to master multiple commonly used tools (e.g., calculators, APIs, and databases) to solve diverse real-world tasks, showcasing the scalability of RL in building versatile, tool-augmented agents.

\subsection{Multimodal Tasks}
\label{sec:application_multimodal}

\begin{myboxi}[Takeaways]
\begin{itemize}
    \item RL strengthens multimodal models to address challenges such as limited-data settings, long-video reasoning, and numerically or attribute-sensitive cross-modal generation.
    \item Exploring unified RL frameworks for understanding and generation is an urgent task.
\end{itemize}
\end{myboxi}

The success of RL is evident not only in language models, but also in fostering notable progress in multimodal tasks.
Specific optimization has been developed to enhance capabilities such as spatial perception~\citep{su2025thinking, chen2025sifthinker} and cross-modal controllability~\citep{wu2025reinforcement, chen2025visrl}.
In the following, we discuss RL applications in multimodal tasks in terms of understanding and generation.

\paragraph{Multimodal Understanding.}
Compared to the language scenario, multimodal understanding demands powerful spatial perception and semantic alignment cross-modalities.
Recently, a surge of research has employed RL to enhance reasoning ability across images, videos, and 3D spaces, demonstrating significant improvements in understanding capability.

\begin{itemize}
    \item \textbf{RL in Image Understanding}:
Vision-R1~\citep{huang2025vision}, VLM-R1~\citep{shen2025vlm}, and Visual-RFT~\citep{liu2025visual} represent the first attempt to extend the DeepSeek-R1 styled RFT from math and code domains to multimodal perception tasks.
These methods mark a shift in training paradigm: moving from data scaling in SFT toward the strategic design of verifiable reward functions tailored to task-specific objectives.
They achieve strong performance on several detection and grounding benchmarks, demonstrating the advanced generalization ability of Reinforced Fine-Tuning (RFT) even with limited training data.
Subsequently, several visual reasoning models~\citep{xia2025visionary,kan2025taco} adopt a similar thinking-answer format in an attempt to learn through trial and error. 
These methods enhance reasoning abilities via outcome-reward-driven optimization, eliminating the need for costly step-wise supervision or CoT training data.
Recently, Deepeyes~\citep{zheng2025deepeyes}, CoF~\citep{zhang2025chain}, and others~\citep{su2025pixel,cao2025ground,fan2025grit} have extended beyond pure text-based CoT to explicit multimodal-interleaved reasoning chains.
These methods attempt to iteratively identify regions of interest in images using off-the-shelf tools~\citep{su2025openthinkimg} or image generation models~\citep{xu2025visual}, achieving more interpretable reasoning processes.
Other methods~\citep{chung2025don, chu2025qwen} implement implicit multimodal-interleaved COT by copying and routing visual tokens during the reasoning stage, which mitigates hallucinations in long text-based CoT.
\citet{xing2025caprl} introduce CapRL, a framework using RLVR to enhance image captioning quality by deriving rewards from the accuracy of language models answering questions based on captions.
\citet{ling2025table2latex} propose a reinforced multimodal language model framework using a dual-reward strategy to improve the fidelity and quality of LaTeX code generation from table images.
\citet{chen2025perception} introduce a two-stage RL framework to jointly enhance the visual perception and reasoning capabilities of vision-language models (VLMs).
Despite the remarkable success, several challenges remain to be addressed: 1) Inconsistent reasoning and answering: The thinking generated by the model fails to map to the final answer. 
2) Long-chain exploration collapse: As the response length increases, the model becomes fragile and prone to generating hallucinations. 
3) Sensitivity to data quality: RL sample selection is crucial, as low-quality training data may lead to suboptimal performance or even negative optimization.

\item \textbf{RL in Video Understanding}:
Extending video understanding capacity to interpret and reason over dynamic visual content is essential for multimodal understanding. 
To achieve this goal, Video-R1~\citep{feng2025video-r1} introduces a systematic RL framework for video Multimodal Large Language Models (MLLMs), using a temporal-aware GPRO algorithm (T-GRPO) to improve spatial-temporal reasoning. 
ReAd-R~\citep{long2025adsqa} proposes a framework optimized via rule-based RL to simulate human heuristic thinking for ad video understanding.
Focused Thinking~\citep{dang2025focused-thinking} employs a token-weighted reward scheme that trims verbose, generic chains-of-thought and uses graded (partial-credit) rewards to enhance video reasoning. VQ-Insight~\citep{zhang2025vq-insight} designs hierarchical rewards with general task-specific temporal learning tailored QA process over long videos. To understand human daily lives from a first-person perspective, Ego-R1~\citep{tian2025ego} trains a chain-of-tool-thought agent via RL to tackle ultra-long egocentric videos (days or weeks in length) by dynamically invoking retrieval and vision tools for stepwise reasoning. Likewise, LongVILA~\citep{chen2025long-rl-video}'s Long-RL framework builds a large LongVideo-Reason dataset and a specialized two-stage CoT-SFT and RL pipeline with sequence parallelism, enabling MLLMs to process ultra-long videos. To automate more video CoT data creation, VideoRFT~\citep{wang2025videorft} uses an LLM to generate initial rationales from rich video descriptors with a VLM refinement and introduces a semantic consistency reward to align textual reasoning with visual evidence. Meanwhile, VideoChat-R1~\citep{li2025videochat-r1} demonstrates that targeted multi-task RL fine-tuning can markedly enhance specific spatio-temporal skills without degrading general chat performance. Collectively, these studies pave the way for the development of robust and generalizable video reasoning through RL.

\item \textbf{RL in 3D Understanding}:
While MLLMs have made significant progress in 2D visual understanding through RL, extending their ability to visual-spatial understanding in 3D space remains a challenging frontier~\citep{yang2025thinking-in-space, wu2025spatial-mllm}. MetaSpatial~\citep{pan2025metaspatial} employs a multi-turn RL-based optimization mechanism that integrates physics-aware constraints to enhance spatial reasoning in MLLMs. Building upon GRPO~\citep{shao2024deepseekmath}, Spatial-MLLM~\citep{wu2025spatial-mllm} and SpaceR~\citep{ouyang2025spaceR} demonstrate that even small-scale models can close the performance gap with much larger counterparts through R1-Zero-like training~\citep{liao2025improved-visual-spatial}. Further, RoboRefer~\citep{zhou2025roborefer} expand RL-based spatial reasoning to embodied settings to ground reasoning in real-world dynamics.

\end{itemize}

\paragraph{Multimodal Generation.}
The exploration of RL in LLMs has also been extended to multimodal generation.
Pioneering researches on test-time scaling~\citep{liu2025video, ma2025inference, singhalgeneral} and DPO~\citep{wallace2024diffusion, liang2025aesthetic, tong2025delving, liu2025videodpo, black2024training} have driven significant progress in aesthetic and text fidelity in image and video generation.
Recently, increasing attention has been devoted to enhance reasoning capabilities in image and video generation~\citep{guo2025can, jiang2025t2i}.

\begin{itemize}

\item \textbf{RL in Image Generation}:
Diffusion models have substantially advanced visual generation~\citep{rombach2022high, liuflow, esser2024scaling}, and a growing body of research incorporates RL to implicitly perform reasoning by treating the denoising steps as the CoT trajectory~\citep{pan2025self, liu2025flow, xue2025dancegrpo}.
However, GRPO exhibits an inherent conflict between ordinary differential equation (ODE) sampling in diffusion models.
Specifically, GRPO relies on stochastic sampling to estimate advantage, whereas ODE sampling follows a deterministic denoising trajectory, which limits the diversity of rollout samples.
To address this issue, an ODE-to-SDE conversion is employed~\citep{liu2025flow, xue2025dancegrpo, wu2025qwen} to encourage the stochastic term in the sampling process.
Considering the inefficiency of SDE, MixGRPO~\citep{li2025mixgrpo} designs mixed sampling strategies through the integration of SDE and ODE.
In addition, TempFlow-GRPO~\citep{he2025tempflow} explicitly exploits the temporal structure in the flow-based model, enabling more precise credit assignment and policy optimization.
Recently, GPT-4o has demonstrated powerful text fidelity and editing consistency~\citep{openai2024gpt4oimage}, sparking interest in the controllability of autoregressive models.
Building on large-scale image–text training data, SimpleAR~\citep{wang2025simplear} directly applies GRPO for post-training and achieves remarkable performance in high-resolution image generation.
To strengthen adherence to fine-grained attributes such as spatial relations and numerical consistency, FocusDiff~\citep{pan2025focusdiff} constructs paired datasets that differ only in subtle attribute variations and uses them to train the generation model. 
Furthermore, RePrompt~\citep{wu2025reprompt} incorporates an additional multimodal understanding model into the image generation framework and trains it with GRPO to refine prompts. 
Meanwhile, T2I-R1~\citep{jiang2025t2i}, GoT-R1~\citep{duan2025got}, and ReasonGen-R1~\citep{zhang2025reasongen} unify prompt refinement and image generation within a single model, leveraging GRPO for joint optimization.

\item \textbf{RL in Video Generation}:
Compared to image generation, extending RL to video generation poses greater challenges in terms of temporal coherence and physical realism.
DanceGRPO~\citep{xue2025dancegrpo} conducts post-training on HunyuanVideo~\citep{kong2024hunyuanvideo}, and uses VideoAlign~\citep{liu2025improving} to provide rewards based on video aesthetics, motion quality, and text-video consistency. 
Furthermore, InfLVG~\citep{fang2025inflvg} employs GRPO to guide token selection according to contextual relevance, thereby enabling semantically consistent and temporally coherent long video generation.
In addition, Phys-AR~\citep{lin2025reasoning} introduces velocity and mass as verifiable rewards for ball motion scenario, substantially enhancing the physical realism of video generation.

\end{itemize}

Currently, several ULM models employ a unified framework to optimize multimodal understanding and generation simultaneously.
To this end, bidirectional~\citep{jiang2025co} and dual~\citep{hong2025reinforcing} rewards from text to image and from image to text are proposed to enhance both the generation and understanding capabilities.
For multimodal understanding, Deepeyes and CoF have attempted to employ generative models or external tools to realize multimodal CoT. For multimodal generation, using refined text as the CoT also relies on the multimodal understanding capability. Therefore, exploring unified post-training methods for multimodal understanding and generation is an urgent task for future research.
From the perspective of specific-domain, code generation can serve as a bridge between text and image generation.
The application of RL to facilitate models to reason over complex charts and produce structured code for domain-specific image generation~\citep{chen2025chart, chen2025breaking, tan2025chartmaster} is a promising application.

\subsection{Multi-Agent Systems}
\label{sec:application_mas}
\begin{myboxi}[Takeaways]
\begin{itemize}
    \item It is important to improve collaboration, reasoning, and credit assignment in Multi-Agent Systems (MAS), enabling more stable and effective teamwork on complex tasks.
	\item Key challenges remain in developing efficient collaboration and interaction mechanisms to fully unlock collective capabilities and further raise agent performance.
\end{itemize}
\end{myboxi}

Currently, most of the research on RL for LLM-based reasoning predominantly centers on single models, whereas applying RL to MAS has emerged as a prominent and frontier research direction. This section begins with an overview of the fundamental concepts of traditional RL and Multi-Agent RL (MARL), highlighting their primary challenges. Furthermore, the section discusses innovative applications of LLMs in MARL, emphasizing their advantages in information sharing and credit assignment. Finally, recent advances in MAS integrating RL with LLMs are examined, with a focus on how RL can be exploited to enhance collaboration and policy optimization among agents, thereby promoting the development of multi-agent reasoning capabilities.

\paragraph{Traditional MARL.}
In recent years, as a complex distributed intelligent system, MAS have attracted widespread attention in the field of RL~\citep{dorri2018multi}. Traditional MARL~\citep{busoniu2008comprehensive} primarily focuses on the interactions and joint learning of multiple agents within a shared environment to achieve global objectives. The main challenges in conventional MARL include the complexity of credit assignment, the nonstationarity of the environment, and the efficiency of communication and cooperation among agents~\citep{canese2021multi}. To address these issues, researchers propose a centralized training with decentralized execution (CTDE) paradigm~\citep{lowe2017multi}, in which agents share global information for policy optimization during the training phase, while decision-making during execution relies solely on local observations. Based on the CTDE paradigm, researchers introduce value-based methods (such as VDN~\citep{sunehag2017value} and QMIX~\citep{rashid2020monotonic}), policy gradient-based methods (such as MADDPG~\citep{lowe2017multi}), and actor-critic methods (such as COMA~\citep{foerster2018counterfactual}). 
Moreover, as PPO is considered to be SOTA in traditional RL, MAPPO has also been shown to have surprising effects in some simple collaborative tasks~\citep{yu2022surprising}.
However, as the number of agents increases and the task complexity rises, traditional MARL methods face significant challenges in terms of sample efficiency and scalability.
To address this issue, scholars have considered replacing current agent with neighboring agents in the interaction with all agents (such as MF-MARL~\citep{yang2018mean}), which effectively alleviates the dimensionality curse caused by the increase in the number of agents in MARL. However, it still cannot be efficiently applied to complex task scenarios that require multiple agents to collaborate simultaneously.

\paragraph{LLM for MARL.}
The rapid development of LLMs has demonstrated tremendous potential in addressing challenges within MARL. Leveraging their powerful natural language understanding and generation capabilities, LLMs can provide effective information-sharing mechanisms in MAS. For instance, in credit assignment problems of MARL, researchers utilize LLMs to design intuitive reward allocation mechanisms, thereby enhancing the accuracy and interpretability of credit assignment. \citet{zhang2023proagent} significantly improve multi-agent collaboration efficiency in sparse reward scenarios by enabling the LLMs to infer each agent’s intention in real time and generate the next cooperative plan. \citet{ding2023entity} leverage LLMs to parse natural language task descriptions into executable entity-level sub-goals, thereby achieving reward shaping and policy sharing, which effectively alleviates the credit assignment problem in MARL. \citet{li2023theory} utilize the LLM's ``theory of mind'' capability, allowing agents to generate linguistic beliefs about teammates’ potential strategies, thus enabling more accurate decision-making in multi-agent coordination.

\paragraph{RL for LLM-based MAS.}
In the context of integrating RL with LLMs, research on MAS based on LLMs has gradually become a hotspot. Related studies primarily focus on how to fully leverage the language understanding and generation capabilities of LLMs, while utilizing RL to achieve efficient collaboration and policy optimization among multiple agents. Frameworks such as LLaMAC and CTRL integrate LLMs with the actor-critic architecture. LLaMAC~\citep{zhang2023controlling} employs a centralized LLM-Critic to provide natural language-based value feedback to multiple LLM-Actors, thereby facilitating collaborative learning among multiple agents. CTRL~\citep{xie2025teaching} trains LLMs to ``self-criticize'' by using synthetic data, and iteratively refines model outputs through RL (such as GRPO), which can improve test-time performance without the need for human annotation. %

In large-scale multi-agent collaboration scenarios, MAPoRL~\citep{park2025maporl} promotes efficient and transferable collaboration in multi-turn tasks by jointly training multiple LLMs and introducing reasoning-aware rewards. MAGRPO~\citep{liu2025llm} models LLM collaboration as a cooperative multi-agent RL problem, which proposes a group-level relative policy optimization mechanism that significantly enhances the quality of multi-turn joint outputs in tasks such as writing and code generation. ReMA~\citep{wan2025rema} introduces dual LLM structure of high-level agent and low-level agent, which achieves synergistic enhancement of meta-thinking and reasoning abilities through alternating freezing and updating of policies. JoyAgents-R1~\citep{han2025joyagents} designs a joint evolutionary training process, facilitating both diversity and consistency within heterogeneous LLM teams in open-domain question answering tasks through alternating global experience replay and individual PPO updates. AlphaEvolve~\citep{novikov2025alphaevolve} designs an evolutionary optimization mechanism to coordinate multi-LLM collaboration. By directly modifying code and continuously receiving evaluation feedback, the MAS enhances the capability to handle complex coding tasks. AutoAgents~\citep{chen2023autoagents} significantly enhance the adaptability and problem-solving capabilities of MAS in complex tasks by dynamically generating specialized agents tailored to task requirements and incorporating an observer role for reflection and improvement.

\subsection{Robotics Tasks}
\label{sec:application_robotics}
\begin{myboxi}[Takeaways]
\begin{itemize}
	\item RL addresses data scarcity and generalization challenges in robotics by adapting LLM-style approaches to Vision-Language-Action (VLA) models.
	\item Allowing VLAs to learn from environment interaction and simple rewards, recent RL methods (e.g., GRPO, RLOO, PPO) achieve superior performance and novel behaviors with minimal supervision.
\end{itemize}
\end{myboxi}

\paragraph{RL in Robotics Tasks.}
RL has been extensively applied in robotics, primarily focusing on three domains: robot control, Vision-and-Language Navigation (VLN), and robotic manipulation tasks. Traditional RL research in robot control has reached maturity with widespread applications, like action generation with human-like robots~\citep{peng2018deepmimic}, robust quadruped locomotion execution~\citep{hwangbo2019learning} and dexterous hand manipulation~\citep{chen2023bi}.
Similarly, VLN tasks have seen significant progress~\citep{anderson2018vision,wang2018look,wang2019reinforced}. However, these domains differ substantially from LLM-based RL in terms of model architecture, scale, task types, reward function design, optimization objectives, and algorithmic approaches, and thus fall outside the scope of this survey.

Robotic manipulation tasks, enabling robots to solve diverse manipulation problems in real-world environments, represent the most challenging and fundamental aspect of embodied intelligence~\citep{firoozi2025foundation}. These tasks demand not only a comprehensive understanding of visual and textual information and fine-grained motor control, but also physical reasoning, long-horizon planning, and logical inference capabilities. Leveraging the remarkable text and vision processing capabilities of LLMs and VLMs, several studies have explored using these models as core components combined with action modules for manipulation tasks, such as RobotBrain~\citep{ji2025robobrain} and RT-2~\citep{zitkovich2023rt}.

\paragraph{Vision-Language-Action Models.}
Recently, Vision-Language-Action (VLA) models, which integrate VLM backbones with action modules through unified end-to-end training, have emerged as the most promising solution and become the mainstream approach for robotic manipulation~\citep{zhong2025survey}. Current VLA models follow a two-stage paradigm~\citep{sapkota2025vision}: pretraining on multimodal data (e.g., Open X-Embodiment~\citep{o2024open}) followed by supervised fine-tuning on teleoperated robot trajectories. However, this imitation learning paradigm suffers from critical limitations: its performance heavily depends on high-quality trajectory data that is expensive and inefficient to collect, and the resulting models exhibit poor generalization to unseen scenarios. Given the architectural, scale, and methodological similarities between VLAs and LLMs~\citep{zhong2025survey}, adapting LLM-style RL approaches to VLA training presents a promising direction for addressing data scarcity and generalization challenges.

Applying DeepSeek-R1's RL methodology to VLAs requires addressing several challenges: 1) Unlike LLMs that complete tasks in a single round, VLAs require multi-round environment interactions to generate complete trajectories; 2) VLAs operate in continuous action spaces; 3) Traditional RL methods rely on hand-crafted process rewards, limiting scalability. Recent works including SimpleVLA-RL~\citep{li2025simplevla}, VLA-RL~\citep{lu2025vla}, VLA RL Generalization~\citep{liu2025can}, RIPT-VLA~\citep{tan2025interactive}, and ConRFT~\citep{chen2025conrft} have pioneered the application of DeepSeek-R1's methodology to VLA training.

SimpleVLA-RL~\citep{li2025simplevla} enables VLA models to interact with environments to rollout diverse complete trajectories, employing binary success/failure rewards as supervision signals and training OpenVLA-OFT~\citep{kim2025fine} using the GRPO algorithm. With just a single demonstration trajectory, this RL approach surpasses state-of-the-art VLA models like $\pi_0$~\citep{black2024pi_0} on LIBERO and RobotWin2.0 benchmarks, achieving SOTA performance and outperforming advanced RDT models in real-robot experiments.
In addition, as an upgraded version of $\pi_0,\pi_{0.5}$~\citep{intelligence2025pi_} uses multimodal robot data from different scenarios and sources for heterogeneous training, allowing VLA to provide a new milestone in generalizable real-world robot operation tasks.
Similar to DeepSeek-R1's ``aha moments'', RL-trained VLAs also discover novel behavioral patterns. 
VLA RL Generalization~\citep{liu2025can} investigates RL's impact on VLA generalization capabilities, demonstrating significant improvements over SFT in unseen environments, objects, and textures, while comparing GRPO and PPO effectiveness. RIPT-VLA~\citep{tan2025interactive} employs RLOO~\citep{ahmadian2024back} for VLA RL training. RLinf~\citep{RLinf_repo} designed a flexible, scalable RL framework for VLA RL that unifies rendering, inference, and training, improving both VLA training efficiency and performance. ConRFT~\citep{chen2025conrft} iteratively trains VLAs through alternating RL and SFT rounds, progressively enhancing performance through multiple iterations.
\citet{ghasemipour2025self} introduce a two-stage approach combining supervised fine-tuning and RL to enhance robotics foundation models for autonomous skill acquisition and improved policy success rates.

The data efficiency, improved generalization, and minimal supervision requirements of RL effectively address VLA's current challenges of data scarcity and poor generalization. By allowing VLAs to autonomously explore and learn from trial-and-error with only outcome supervision, this approach dramatically reduces implementation costs compared to complex and expensive teleoperation data collection. Moreover, RL's data efficiency eliminates the need for large-scale expensive trajectory datasets, enabling scalable VLA post-training capabilities.

However, current VLA RL research remains primarily simulation-based. While SimpleVLA-RL~\citep{li2025simplevla} achieved real-world deployment through Sim2Real transfer~\citep{chen2025robotwin}, few works have yet deployed physical robots to collect real-world trajectories for RL.
In addition, research on VLA RL is also limited by the current development of RL in robotics, including but not limited to sample efficiency, reward sparsity, and sim2real.
Key challenges include autonomous sampling on physical robots requiring multiple devices for efficiency, continuous manual resetting and annotation. 

\subsection{Medical Tasks}
\label{sec:application_medical}
\begin{myboxi}[Takeaways]
\begin{itemize}
	\item RL for medical LLMs faces distinct challenges: verifiable tasks allow stable reward design, while non-verifiable tasks make reward definition difficult.
	\item Verifiable tasks use SFT+RL with rule-based rewards; non-verifiable tasks leverage DPO, rubrics, curriculum RL, or offline RL, though scalability and stability remain open issues.
\end{itemize}
\end{myboxi}

RL optimizations in medical LLMs typically aim to enhance reasoning and generalization ability, often adopting a two-stage pipeline of SFT followed by RL. Existing works can be broadly categorized into \textbf{verifiable problems with rule-based rewards}, and \textbf{non-verifiable problems with generative or rubric-based rewards}.

\paragraph{Medical Understanding.}  
These tasks, such as multiple-choice QA, structured prediction, clinical coding, or visual grounding, allow the use of deterministic rewards, making them the most mature field for RL  in medical LLMs. The typical paradigm is a two-stage pipeline of SFT followed by RL, where algorithms such as GRPO optimize models directly against correctness-based signals. For example, HuatuoGPT-o1~\citep{chen2024huatuogpto1medicalcomplexreasoning}  enhances reasoning ability by synthesizing reliable reasoning trajectory data with a medical verifier and training the model with SFT and RL. Med-U1~\citep{zhang2025med-u1} employs mixed binary correctness rewards with length penalties to ensure both accuracy and format compliance, while MED-RLVR~\citep{zhang2025med} applies verifiable rewards to MCQA, improving OOD generalization. Open-Medical-R1~\citep{qiu2025openmedicalr1choosedatarlvr} demonstrates that careful data filtering improves the efficiency of RL. Gazal-R1~\citep{arora2025gazalr1} designs a multi-component reward system that refines accuracy, format adherence, and reasoning quality through GRPO for enhanced medical reasoning. ProMed~\citep{ding2025promed} shifts medical LLMs from reactive to proactive paradigms, where LLMs can ask clinically valuable questions before decision-making, using Shapley Information Gain rewards during MCTS-guided trajectory exploration and RL. MedGR$^2$~\citep{zhi2025medgr2} introduces a generative reward learning framework that creates a self-improving virtuous cycle, co-developing a data generator and reward model to enable automated creation of high-quality multimodal medical data for both SFT and RL training.

Beyond textual QA, recent models extend rule-based rewards to vision and multi-modal tasks. MedVLM-R1~\citep{pan2025medvlmr1} employs an RL  framework that incentivizes the model to discover human-interpretable reasoning paths without using any reasoning references through format and accuracy rewards. MedGround-R1~\citep{xu2025medgroundr1} introduces spatial-semantic rewards, which combine spatial accuracy reward and semantic consistency reward, for the medical imaging grounding task. ARMed~\citep{liu2025armed} addresses reward collapse in open-ended medical VQA through adaptive semantic rewards that dynamically adjust the semantic reward during training based on historical reward distributions. \citet{liu2025efficientvie} leverage rule-based format and matching rewards to guide structured JSON generation for medical visual information extraction with only 100 annotated samples. MMedAgent-RL~\citep{xia2025mmedagent} is an RL-based multi-agent framework that enables dynamic and optimized collaboration among medical agents.
MedGemma~\citep{sellergren2025medgemma} was post-trained with RL and is further evaluated on MedXpertQA~\citep{zuo2025medxpertqa}, which is an expert-level medical multi-choice benchmark and includes a subset for assessing reasoning models.

For other clinical applications, DRG-Sapphire~\citep{wang2025drg} applies GRPO with rule-based rewards to diagnosis-related grouping.
EHRMIND~\citep{lin2025ehrmind} combines SFT warmup and RL VR for complex clinical reasoning tasks using electronic health records (EHR) data, including medical calculations, patient trial matching, and disease diagnosis. ChestX-Reasoner~\citep{fan2025chestx} incorporates process rewards from clinical reports to train the model to emulate radiologists' step-by-step reasoning. 
CX-Mind~\citep{li2025cxmind} employs SFT and RL  with format, result, and process rewards to train interleaved reasoning for chest X-ray diagnostics.
To enable benchmarking of code-based medical reasoning, MedAgentGym~\citep{xu2025medagentgym} presents a benchmark for code generation of medical agents, and demonstrates that RL  can improve this reasoning ability.
Fleming-R1~\citep{liu2025fleming} leverages a reasoning-oriented data strategy, structured initialization, and RLVR to achieve expert-level clinical reasoning.

\paragraph{Medical Generation.}  
These tasks include radiology report generation~\citep{jing2025reason}, multi-turn clinical dialogue~\citep{baniharouni2025lacdm}, treatment planning~\citep{nusrat2025dola}, and diagnostic narratives~\citep{lim2025moreclear}, which lack unique ground-truth answers. 
As such, rule-based rewards are not directly applicable. While DPO has been applied to improve medical LLMs on preference-aligned generation tasks~\citep{yu2025finemedlmo1enhancingmedicalknowledge,yang2025medreflmedicalreasoningenhancement}, large-scale RL  on non-verifiable tasks is emerging but remains relatively underexplored. 
For example, DOLA~\citep{nusrat2025dola} integrates LLM agents with a commercial treatment planning system, incorporating a reward function that guides the trade-offs between target coverage and organ at risk sparing for optimized treatment plan generation. 
LA-CDM~\citep{baniharouni2025lacdm} proposes a two-agent structure trained via a hybrid training paradigm which combined supervised fine-tuning with RL  to balance diagnostic accuracy, uncertainty calibration, and decision efficiency. In diagnostic dialogue, PPME~\citep{sun2025ppme} develops a plug-and-play framework using large-scale EMRs and hybrid training to enhance LLM interactive diagnostic capabilities through specialized inquiry and diagnosis models. In clinical decision support, MORE-CLEAR~\citep{lim2025moreclear} applies multi-modal offline RL  to sepsis treatment policies, improving survival-predictive decision-making in MIMIC-III/IV. For radiology report generation, BoxMed-RL~\citep{jing2025reason} leverages RL in its pretraining phase, using a format reward and an Intersection-over-Union (IoU) reward to ensure that the generated reports correspond to pixel-level anatomical evidence. Baichuan-M1~\citep{2025baichuan} employs a three-stage RL approach: ELO (Exploratory Log-likelihood Optimization) to enhance chain-of-thought reasoning diversity, TDPO (Token-Level Direct Preference Optimization) to address length-dependent constraints, and finally PPO with reward model feedback for policy refinement. Baichuan-M2~\citep{2025baichuan2} introduces a novel dynamic verification framework that moves beyond static answer verifiers, establishing a large-scale, high-fidelity interactive reinforcement learning system with a Patient Simulator and Clinical Rubrics Generator for realistic clinical environments.

Overall, RL  in medical LLMs is well established for verifiable problems, where deterministic correctness allows for rule-based rewards and stable GRPO training. In contrast, generation-oriented tasks remain challenging: current solutions adopt rubric-based rewards, curriculum transfer, or offline RL  to approximate quality signals. The scarcity of scalable RL  on non-verifiable tasks highlights a critical future direction for building trustworthy, reasoning-capable medical foundation models.

\section{Future Directions}
\label{sec:future}

While RL for LLMs has made remarkable strides, many fundamental challenges and opportunities lie ahead. This section outlines several promising directions that are poised to shape the next wave of advances in the field. We highlight the importance of continual RL for adapting to evolving data and tasks ($\S$~\ref{sec:future_continual}), memory-based and model-based RL for enhancing reasoning capabilities ($\S$~\ref{sec:future_memory} and $\S$~\ref{sec:future_model_based}), and emerging approaches for teaching LLMs both efficient and latent-space reasoning ($\S$~\ref{sec:future_efficient} and $\S$~\ref{sec:future_latent}). We also discuss frontiers in leveraging RL during pre-training ($\S$~\ref{sec:future_pre_training}), applying RL to diffusion-based architectures ($\S$~\ref{sec:future_diffusion}), and driving scientific discovery ($\S$~\ref{sec:future_scientific}). Finally, we consider the challenges and prospects of architecture-algorithms co-design to meet the demands of ever-larger and high-efficiency intelligent models ($\S$~\ref{sec:future_codesign}). By surveying these directions, we aim to provide both a roadmap and inspiration for future research in RL for LLMs.

\subsection{Continual RL for LLMs}
\label{sec:future_continual}

To enhance the multi-domain performance of LLMs during RL-based post-training, the mainstream approach is to mix data from different tasks and train in a unified manner \citep{guo2025deepseek,yang2025qwen3}. On synthetic data \citep{chen2025enigmata,liu2025synlogic}, Multi-stage RL has been shown to perform worse than training with mixed data, and even curriculum learning with increasing difficulty may not be necessary in RL \citep{xie2025logic}. However, \citet{chen2025enigmata} suggest that multi-stage RL across different tasks has advantages in generalizing to difficult or unseen problems. Despite these ongoing debates of multi-stage RL's effectiveness, as the field advances toward building AI systems that must adapt to evolving data and tasks in dynamic environments, it becomes necessary to explore Continual Reinforcement Learning (CRL) for LLMs.

Similar to traditional CRL, LLMs face the fundamental challenge of balancing stability and plasticity during multi-stage RL training \citep{pan2025survey}. Plasticity may be particularly concerning for LLMs, as widely used deep learning techniques can cause large models to perform no better than shallow networks in continual learning settings \citep{dohare2024loss}. Another challenge of CRL for LLMs lies in the entangled nature of knowledge and reasoning in LLMs, which distinguishes from traditional RL settings where tasks can be discretely defined and policies can be modularly organized, such as in game-like environments \citep{chevalier2023minigrid,towers2024gymnasium} or embodied scenarios \citep{todorov2012mujoco,wolczyk2021continual}.

Existing methodological frameworks from traditional CRL research provide a promising foundation for addressing LLM-specific requirements. Core methodological insights from traditional CRL research, including Experience Replay \citep{rolnick2019experience,berseth2021comps,li2021sler}, Policy Reuse \citep{garcia2019meta,gaya2022building}, and Reward Shaping \citep{jiang2021temporal,zheng2022lifelong}. It remains a valuable research direction for developing CRL frameworks tailored to LRMs. The development of specialized CRL techniques for LLMs or LRMs will be crucial for creating more adaptive and efficient AI systems capable of lifelong learning and operating in dynamic and ever-changing environments.

\subsection{Memory-based RL for LLMs}
\label{sec:future_memory}

Although many works in agentic RL have explored memory mechanisms, ranging from external long-term storage and insertion \citep{zhong2024memorybank,chhikara2025mem0,xu2025mem} to internal memory processing and working-memory control \citep{zhou2025mem1,yu2025memagent}, most designs remain tailored to the current task with limited generalization beyond it. As \citet{silver2025welcome} emphasize, the next generation of intelligent agents will learn primarily from experience, acquiring skills through continual interaction. In this spirit, a key direction is to transform agent memory from task-specific buffers into experience repositories that are structured, reusable, and transferable across diverse tasks, allowing memory to evolve into a foundation for broader adaptability and lifelong learning. Such an experience-centric view also aligns naturally with RL, since the data generated from the interactions between an agent and its environment provides rich experiential traces that can be utilized effectively. Moreover, although recent works have explored maintaining a shared pool of experiences to retrieve relevant strategies from past histories and adapt other agents’ experiences to new task scenarios \citep{tang2025agent}, this direction remains underexplored. A core challenge here is enabling agents, through RL, to automatically learn how to operate and manage memory, composing and generalizing experiential knowledge across tasks. Addressing this challenge is essential for moving toward an ``experience era'' where collective interaction traces become a foundation for broader agent intelligence. 

\subsection{Model-based RL for LLMs}
\label{sec:future_model_based}

A core challenge in RL lies in obtaining scalable and robust reward signals as well as meaningful state representations from the environment. Prior work has investigated the construction of world models~\citep{moerland2023model,luo2024survey} to supply informative states for RL agents, and more recently, LLMs have been adopted as world models in various RL contexts~\citep{hu2023language,benechehab2024zero,gu2024your}. In the case of RL with LLMs, especially for language agents, the ability to construct world models that accurately capture environmental states and generate reliable rewards is critical. Recent advances show that generative world models, including those enhanced by video pre-training~\citep{bruce2024genie,genie3,assran2025v}, are both practical and effective. Nevertheless, seamlessly integrating world models with RL for LLM-based agents remains an open research problem. As such, model-based RL with LLMs is emerging as a particularly promising and scalable direction for future research.

\subsection{Teaching LRMs Efficient Reasoning}
\label{sec:future_efficient}

Inference-time scaling has improved the accuracy of LRMs on difficult tasks, but it also introduces systematic \emph{over-thinking} (needlessly long reasoning chains for easy instances)~\citep{qu2025survey,sui2025stop,chen2024not,yan2025drqa} and, under aggressive truncation, \emph{under-thinking} (premature halting and reliance on brittle shortcuts)~\citep{wang2025thoughts,su2025between}. A central challenge for RL-for-LLMs is to develop \emph{compute-allocation policies} that adapt the depth and halting of reasoning to instance difficulty and epistemic uncertainty.
Current research has explored hard-coded reasoning levels in prompts~\citep{openai2025gpt-oss-120b,wen2025budgetthinker,zhu2025think}, adaptive length-based reward shaping~\citep{yuan2025efficient,liu2025learn}, and the use of length penalties in the loss function~\citep{aggarwal2025l1,xiang2025just}.

However, generalizing these approaches into a principled cost-performance trade-off remains an open question~\citep{gan2025cotspacetheoreticalframeworkinternal}.
Teaching LRMs to be \emph{resource-rational}, to reason longer only when the marginal utility justifies it, remains a central, unsolved problem for RL in language reasoning.

\subsection{Teaching LLMs Latent Space Reasoning}
\label{sec:future_latent}

CoT \citep{wei2022chain} encourages step-by-step reasoning by prompting models to articulate intermediate steps, improving both interpretability and accuracy.
Recent research has combined CoT and RL to further improve reasoning quality, which samples long-form thought before answering for modeling training \citep{guo2025deepseek}. However, current implementations often rely on token-level sampling \citep{ouyang2022training, rafailov2023direct, cui2025process} in a discrete scalar space, which can act as a bottleneck as the lost of meaningful semantic information in continuous space \citep{hua2024intuitive}.
A recently proposed method, named Latent Space Reasoning (LSR) \citep{hao2024training, arriola2025block, geiping2025scaling}, may be more friendly for RL optimization. LSR operates reasoning in the continuous latent space of LLMs, facilitating more nuanced and fluid semantic reasoning. This characteristic contributes to smoother learning dynamics and a better integration with RL techniques. The combination of RL and LSR holds significant potential for the development of more powerful and adaptable reasoning models in the future. However, assessing the quality of continuous latent thought is more challenging than evaluating token-based thought. This will complicate the provision of accurate supervisory signals, such as rewards and advantages, which will become an open challenge against the combination of LSR and RL.

\subsection{RL for LLMs Pre-training}
\label{sec:future_pre_training}

Traditional pre-training relies on large text corpora and next-token prediction, and scaling this paradigm has already been shown to be central to the development of foundation models~\citep{brown2020language,kaplan2020scaling}.
Emerging research now explores shifting RL earlier in the pipeline, applying it not only in post-training but also during pre-training itself.
For instance, Reinforcement Pre-Training~\citep{dong2025reinforcement} reconceptualizes next-token prediction as an RL problem with verifiable rewards derived from the corpus, reporting consistent gains that increase with available compute, thereby positioning RL as a promising scaling strategy for pre-training.

In parallel, open initiatives such as avataRL~\citep{avatarl2025} demonstrate training language models from random initialization purely with RL, bootstrapping token-level rewards and employing iterative ``referee'' scoring, thus illustrating a concrete path toward RL-from-scratch training.
It is consistent with the reincarnated RL paradigm~\citep{agarwal2022reincarnating}, in which previously acquired computational knowledge (the pre-trained critic) is leveraged rather than training from the ground up.
These developments sharpen a practical question: \textit{how can RL-style pre-training be made cost-effective at scale?}
Addressing this challenge will likely require reducing both the verifier burden and the costs associated with reward engineering, which appear to be critical for scaling RL-based pre-training.
Moreover, this line of research is closely related to unsupervised reward design introduced in $\S$~\ref{sec:reward_unsupervised}, raising important questions about \textit{how to obtain rewards that are both scalable and reliable}.

\subsection{RL for Diffusion-based LLMs}
\label{sec:future_diffusion}

Diffusion Large Language Models (DLLMs)~\citep{nie2025largelanguagediffusionmodels,ye2025dream,xie2025dream,JetAstra2025,tae2025tess,labs2025mercury} represent an emerging paradigm in language generation. 
Compared to autoregressive (AR) models, DLLMs offer advantages including superior decoding efficiency and a greater potential for self-correction through multiple rounds of diffusion. Initial efforts have begun to explore RL for DLLMs \citep{gong2025diffucoder, yang2025mmada, borso2025preference}, yet several key issues remain unresolved.

A central challenge in applying RL to DLLMs lies in accurately and efficiently estimating log probabilities of sampled responses.
This is due to a fundamental difference in how autoregressive models and diffusion language models inherently model the likelihood of samples.
AR models generate sequences through next-token prediction and factorize joint probabilities via the chain rule, enabling straightforward left-to-right sampling. However, DLLMs approximate likelihood optimization by maximizing the Evidence Lower Bound (ELBO).
ELBO involves a double expectation over diffusion timesteps and masked data, and typically demands extensive sampling to achieve accurate estimates; otherwise, it introduces high variance during preference optimization. 
Although methods like the one-step estimator in \citep{zhao2025d1scalingreasoningdiffusion} and the sampling allocation strategy in \citep{zhu2025llada} have been proposed to mitigate variance, efficient and accurate ELBO estimation remains an open problem for on-policy learning.

Furthermore, the existence of multiple feasible decoding trajectories in DLLMs introduces an additional research dimension: leveraging RL to guide the model toward optimal sampling traces.
This requires designing effective reward functions for intermediate denoising steps. For example, \citet{he2025mdpo} formulate denoising as a multi-step decision problem and applies reward models to intermediate states, \citep{wang2025trado} proposed a diffusion-based value model that computes prefix-conditioned, token-wise advantages to enable trajectory-level rewards, while \citet{song2025seed} utilize edit-distance-based rewards to maximize decoding efficiency. 
Future work may also draw inspiration from RL techniques developed for continuous diffusion models in computer vision~\citep{black2024training, xue2025dancegrpo, yang2024using}, potentially paving the way toward a unified multimodal framework.

\subsection{RL for LLMs in Scientific Discovery}
\label{sec:future_scientific}

Recent research has shown that involving RL can improve the performance of LLMs on reasoning-heavy scientific tasks, in some cases even allowing them to surpass specialized methods~\citep{fang_cell-o1_2025, rizvi_scaling_2025, narayanan_training_2025, fallahpour_bioreason_2025}. In domains such as biology and chemistry, a core challenge for RL is performing result verification at scale, a process conventionally dependent on wet lab experimentation. Several existing methods have focused on replacing or supplementing experimental verification: Pro-1~\citep{hla_pro-1_2025} uses a Rosetta energy function as the reward function for optimizing protein stability, and rbio1~\citep{istrate_rbio1-training_2025} verifies gene perturbation result predictions using biological models and external knowledge sources. 

Much room for exploration remains for both reward formulation and improving the oracle models themselves. Related to this is the broader problem of constructing suitable RL environments that support rapid experimentation-feedback loops. Agentic systems such as Coscientist~\citep{boiko2023autonomous} and Robin~\citep{ghareeb_robin_2025} have gained success through lab-in-the-loop verification, but such sparse, delayed, and costly feedback signals are impractical for directly training the underlying LLM. \textit{In silico} simulations of experimental environments, for instance perturbation response prediction at the cellular level~\citep{bunne2024build, noutahi_virtual_2025}, represent a potential path forward. However, many of these systems are far from sufficient for replacing realistic lab environments due to their limited scope and critical lack of accuracy and generalizability~\citep{ahlmann-eltze_deep-learning-based_2025, kedzierska_assessing_2023}. Other lines of research have explored incorporating domain-specific models into LLM training to handle scientific data~\citep{fallahpour_bioreason_2025} and developing generalist models capable of a suite of well-defined tasks~\citep{narayanan_training_2025, bigaud_owkinzero_2025}. These directions, coupled with advances in general RL methodology, will continue expanding the use cases of LLMs from narrowly defined tasks to complex interactions with open-ended objectives, enabling them to more substantially contribute to novel discoveries.

\subsection{RL for Architecture-Algorithm Co-Design}
\label{sec:future_codesign}
Most current RL pipelines for LLMs assume a dense Transformer~\citep{vaswani2017attention} or Mixture-of-Experts (MoE)~\citep{shazeer2017outrageously,jiang2024mixtral,dai2024deepseekmoe} backbone, optimizing rewards that are almost exclusively tied to task accuracy. As a result, architectural degrees of freedom, and their hardware implications are left outside the learning loop. In parallel, a new wave of hardware, architecture co-design has emerged (e.g., hardware-aligned sparse attention as in DeepSeek's NSA~\citep{yuan2025native} and model–system co-design in Step-3~\citep{wang2025step}), indicating that greater efficiency and capability can be achieved by aligning model structure with computational substrates.

We argue that making architecture a first-class action space in RL represents an open and high-impact challenge for next-generation LLMs. For instance, reinforced MoE approaches could enable models to learn routing policies, expert activation, capacity allocation, or sparsity patterns during RL, optimizing not only for task reward, but also for hardware-aware objectives such as latency, memory traffic, energy consumption, and activation budgets. In this framing, RL is tasked with learning to ``reason'' not only over tokens~\citep{guo2025deepseek}, but also across parameters and modules, dynamically adapting the model’s topology to each prompt’s difficulty and to real-time compute constraints. This perspective goes beyond classic RL-based neural architecture search (NAS)~\citep{zoph2016neural}, which typically finds a fixed architecture for a given task or dataset. In contrast, reinforced MoE focuses on optimizing routing and modular adaptation per input during inference~\citep{han2021dynamic}, potentially yielding both greater efficiency and flexibility. Key open questions include designing robust multi-objective reward functions that avoid trivial solutions (e.g., all-expert sparsity), achieving stable credit assignment when architectural actions modify network topology, and amortizing architecture policy learning across prompts, tasks, and deployment scales. Addressing these challenges will be crucial for enabling truly integrated architecture–algorithm co-optimization in future LLMs.

\section{Conclusion}
We survey recent advances in RL for LRMs with a particular emphasis on reasoning, effectively transforming LLMs into LRMs. In contrast to prior approaches such as RLHF or DPO, which are primarily designed for human alignment, our focus is on RLVR for LLMs. RLVR enhances the reasoning abilities of LLMs by providing direct outcome-level rewards. Firstly, we present the core components of RLVR, including reward design, policy optimization, and sampling strategies. We summarize multiple research directions and existing work for each section. 
And then we discuss several of the most hotly debated issues in RL training for LLMs.
In addition, we introduce training resources for RL of LLMs, covering static datasets, dynamic environments, and RL infrastructure. Finally, we review downstream applications of RL in LLMs across various scenarios and highlight several promising research directions aimed at achieving super-intelligence through RL-based LLMs.

\newpage

\section*{Author Contributions}
\addcontentsline{toc}{section}{Author Contributions}

We present below the contributions of all participating authors, specifying the primary responsible individual as well as the contributing participants for each section. The specific contributions of each author are detailed as follows:

\begin{itemize}
    \item \textbf{Corresponding Author:} Biqing Qi, Ning Ding, Bowen Zhou
    \item \textbf{Project Lead:} Kaiyan Zhang, Yuxin Zuo
    \item \textbf{Introduction:} Kaiyan Zhang
    \item \textbf{Preliminaries:}
        \begin{itemize}
            \item Background: Kaiyan Zhang
            \item Frontier Models: Shang Qu, Yuru Wang (Survey)
            \item Related Surveys: Yuxin Zuo
        \end{itemize}
    \item \textbf{Foundational Components:}
        \begin{itemize}
            \item Verifiable Rewards: Yuxin Zuo
            \item Generative Rewards: Bingxiang He, Sihang Zeng (Draft)
            \item Dense Rewards: Runze Liu, Yu Fu (Turn-level)
            \item Unsupervised Rewards: Bingxiang He, Yuxin Zuo (Survey)
            \item Reward Shaping: Kaiyan Zhang
            \item Policy Gradient Objective: Youbang Sun, Kaiyan Zhang (Formula)
            \item Critic-based Algorithms: Youbang Sun, Kaiyan Zhang (Formula)
            \item Critic-Free Algorithms: Youbang Sun, Kaiyan Zhang (Formula)
            \item Off-policy Optimization: Xingtai Lv, Yu Fu (Replay Buffer)
            \item Regularization Objectives: Yuchen Zhang, Bingxiang He (Entropy), Yuxin Zuo (Survey)
            \item Dynamic and Structured Sampling: Yuchen Fan, Xuekai Zhu (Efficiency-oriented Sampling)
            \item Sampling Hyper-parameters: Yuxin Zuo, Bingxiang He (Survey)
        \end{itemize}
    \item \textbf{Foundational Problems:} Kaiyan Zhang, Yuxin Zuo
    \item \textbf{Training Resource:}
        \begin{itemize}
            \item Static Corpus: Kai Tian, Zhenzhao Yuan (Survey)
            \item Dynamic Environment: Che Jiang
            \item RL Infrastructure: Kaiyan Zhang
        \end{itemize}
    \item \textbf{Applications:}
        \begin{itemize}
            \item Coding Tasks: Pengfei Li, Xiang Xu (Tool-Integrated Reasoning)
            \item Agentic Tasks: Yuchen Fan, Xinwei Long (GUI/Computer-use)
            \item Multimodal Tasks: Guoli Jia, Fangfu Liu (Video/3D), Xinwei Long (Image)
            \item Multi-Agent Systems: Shijie Wang
            \item Robotics Tasks: Haozhan Li
            \item Medical Tasks: Sihang Zeng, Jiaze Ma (Survey)
        \end{itemize}
    \item \textbf{Future Directions:}
        \begin{itemize}
            \item Che Jiang (CRL), Yu Fu (MemRL), Ermo Hua (Latent), Yuxin Zuo (Pre-Training), Yihao Liu (Diffusion), Shang Qu (Scientific), Kaiyan Zhang (Rest of All)
        \end{itemize}
    \item \textbf{Other Contributions:}
        \begin{itemize}
            \item Figures: Yuru Wang, Kaiyan Zhang, Yuxin Zuo
            \item Review and Editing: Zhiyuan Ma, Ganqu Cui, Huayu Chen, Weize Chen, Yafu Li, Xiaoye Qu, Junqi Gao, Dong Li, Zonglin Li, and all above authors
        \end{itemize}
\end{itemize}

We also thank the broader community for their valuable suggestions and feedback. In particular, we are grateful to Mingjie Wei, Wei Shen, Thomas Schmied, Muhammad Khalifa, Zhangquan Chen, Xinyu Zhu, Jacob Dineen, and Michal Kozak, Zhenpeng Su, Peter Chen for providing constructive feedback, including identifying typographical errors, inaccuracies in descriptions, and missing citations in the paper. We welcome further feedback to help make this work a more valuable resource for the field.

\newpage
\bibliography{srl}

\end{document}